\documentclass[nohyperref]{article}

\usepackage{hyperref}

\usepackage[accepted]{icml2023}

\usepackage{amsmath}
\usepackage{amssymb}
\usepackage{mathtools}
\usepackage{amsthm}

\usepackage[capitalize,noabbrev]{cleveref}

\theoremstyle{plain}
\newtheorem{theorem}{Theorem}[section]
\newtheorem{proposition}[theorem]{Proposition}
\newtheorem{lemma}[theorem]{Lemma}

\theoremstyle{definition}
\newtheorem{definition}[theorem]{Definition}

\theoremstyle{remark}

\usepackage[utf8]{inputenc}
\usepackage{url}            
\usepackage{booktabs}       
\usepackage{amsfonts}      
\usepackage{nicefrac}      
\usepackage{microtype}     
\usepackage{xcolor}       
\usepackage{multirow}

\usepackage{bbm}
\usepackage{dsfont}
\usepackage{bm}
\usepackage{subfig}
\usepackage[colorinlistoftodos, textwidth=26mm, shadow,color=blue!30!white]{todonotes}
\usepackage{cancel,booktabs,tikz} 
\tikzset{
    double color fill/.code 2 args={
        \pgfdeclareverticalshading[%
            tikz@axis@top,tikz@axis@middle,tikz@axis@bottom%
        ]{diagonalfill}{100bp}{%
            color(0bp)=(tikz@axis@bottom);
            color(50bp)=(tikz@axis@bottom);
            color(50bp)=(tikz@axis@middle);
            color(55bp)=(tikz@axis@top);
            color(100bp)=(tikz@axis@top)
        }
        \tikzset{shade, left color=#1, right color=#2, shading=diagonalfill}
    }
}

\usepackage{lipsum,graphicx,multicol}
\usepackage{vwcol}

\renewenvironment{proof}{{\bfseries \textit{Proof.}}}{\qed}

\newcommand{\thistheoremname}{}
\newtheorem*{genericthm*}{\thistheoremname}
\newenvironment{namedprop*}[1]
  {\renewcommand{\thistheoremname}{\normalfont{\textbf{#1}}}%
   \begin{genericthm*}}
  {\end{genericthm*}}
  
\newenvironment{nameddef*}[1]
  {\renewcommand{\thistheoremname}{\normalfont{\textbf{#1}}}%
   \begin{genericthm*}}
  {\end{genericthm*}}

\usepackage[algo2e]{algorithm2e} 
\usepackage{algorithm,algorithmic}
\usepackage{enumitem}
\usepackage{tikz-network}
\tikzstyle{dot}=[circle,fill,inner sep=2.5pt]  
\tikzstyle{dgraph}=[->]
\tikzstyle{dgraph}=[<->]
\newcommand{\arr}{-{Triangle[length=2mm, width=2mm]}}

\newcommand{\acro}[1]{\textsc{#1}\xspace}
\newcommand{\acronospace}[1]{\textsc{#1}}

\newcommand{\expectation}[2]{\mathbb{E}_{#1}[#2]}
\newcommand{\indep}{\rotatebox[origin=c]{90}{$\models$}}
\DeclareMathOperator*{\argmin}{arg\,min}

\newcommand{\datao}{\mathcal{D}^O}
\newcommand{\datai}{\mathcal{D}^I}
\newcommand{\graph}{\mathcal{G}}

\newcommand{\powerI}{\mathcal{P}_{\I}}
\newcommand{\CX}{\C_{\X}}
\newcommand{\lC}{\lambda^{C}}
\newcommand{\lCX}{\lambda^{\CX}}
\newcommand{\domX}{D(\X)}
\newcommand{\Nobs}{N_{O}}
\newcommand{\Nint}{N_{I}}
\newcommand{\ntasks}{M}
\newcommand{\muX}{\mu_{\text{do}(\X=\x)}}
\newcommand{\muXprime}{\mu_{\text{do}(\X^\prime=\x^\prime)}}

\newcommand{\calG}{\mathcal G}

\newcommand{\pa}{\text{pa}}
\newcommand{\an}{\text{an}}

\renewcommand{\eqref}[1]{Eq. (\ref{#1})}
\newcommand{\figref}[1]{Fig. \ref{#1}}
\newcommand{\secref}[1]{Section \ref{#1}}
\newcommand{\ie}{i.e.\xspace}
\newcommand{\eg}{e.g.\xspace}
\newcommand{\runexample}{\textit{\textbf{Example: }}}

\newcommand{\cgo}{\acro{cgo}}
\newcommand{\Cgo}{c\acro{go}}
\newcommand{\ccgo}{c\acro{cgo}}
\newcommand{\bo}{\acro{bo}}
\newcommand{\cbo}{\acro{cbo}}
\newcommand{\Cbo}{c\acro{bo}}

\newcommand{\ccbo}{c\cbo}
\newcommand{\stgpcausal}{\acro{stgp$^+$}}
\newcommand{\mtgpcausal}{\acro{mtgp$^+$}}

\newcommand{\cmis}{c\acro{mis}}

\newcommand{\erk}{\acronospace{e}rk}
\newcommand{\akt}{\acronospace{a}kt}
\newcommand{\mek}{\acronospace{m}ek}
\newcommand{\pka}{\acro{pka}}
\newcommand{\pkc}{\acro{pkc}}

\newcommand{\psa}{\acro{psa}}
\newcommand{\bmi}{\acro{bmi}}

\DeclareBoldMathCommand{\B}{B}
\DeclareBoldMathCommand{\C}{C}
\DeclareBoldMathCommand{\E}{E}
\DeclareBoldMathCommand{\O}{O}
\DeclareBoldMathCommand{\Q}{Q}
\DeclareBoldMathCommand{\U}{U}
\DeclareBoldMathCommand{\W}{W}
\DeclareBoldMathCommand{\V}{V}

\newcommand{\X}{\mathbf{X}}
\newcommand{\x}{\mathbf{x}}
\DeclareBoldMathCommand{\x}{x}
\DeclareBoldMathCommand{\X}{X}
\DeclareBoldMathCommand{\q}{q}
\DeclareBoldMathCommand{\y}{y}
\DeclareBoldMathCommand{\I}{I}
\DeclareBoldMathCommand{\i}{i}
\DeclareBoldMathCommand{\F}{F}

\newcommand{\cond}{\,|\,}

\newcommand{\ei}{\acro{ei}}
\newcommand{\cei}{c\acro{ei}}

\newcommand{\cdf}{\acro{cdf}}

\newcommand{\gp}{\mathcal{GP}}
\newcommand{\gptext}{\acro{gp}}
\newcommand{\gpstext}{\acronospace{gp}s}
\newcommand{\stgp}{\acro{stgp}}
\newcommand{\mtgp}{\acro{mtgp}}
\newcommand{\Gmtgp}{$\graph$-\acro{mtgp}}
\newcommand{\doi}{\text{do}}
\newcommand{\scm}{\acro{scm}}
\newcommand{\scms}{\acronospace{scm}s}

\newcommand{\synone}{\acro{synthetic}-1}
\newcommand{\syntwo}{\acro{synthetic}-2}
\newcommand{\health}{\acro{health}}
\newcommand{\protein}{\acro{protein-signaling}}

\newcommand{\allmis}[1]{\mathbb{M}_{#1,\graph}}
\newcommand{\allmisnull}[1]{n\mathbb{M}_{#1,\graph}}
\newcommand{\cmisreduce}{\texttt{cMISReduce}}

\icmltitlerunning{Constrained Causal Bayesian Optimization}

\begin{document}

\twocolumn[
\icmltitle{Constrained Causal Bayesian Optimization}

\begin{icmlauthorlist}
\icmlauthor{Virginia Aglietti}{1}
\icmlauthor{Alan Malek}{1}
\icmlauthor{Ira Ktena}{1}
\icmlauthor{Silvia Chiappa}{1}
\end{icmlauthorlist}
\icmlaffiliation{1}{DeepMind, London, UK}
\icmlcorrespondingauthor{Virginia Aglietti}{aglietti@deepmind.com}

\vskip 0.3in
]

\printAffiliationsAndNotice{}

\begin{abstract}
We propose constrained causal Bayesian optimization (\ccbo), an approach for finding interventions in a known causal graph that optimize a target variable \emph{under some constraints}. \ccbo first reduces the search space by exploiting the graph structure and, if available, an observational dataset; and then solves the restricted optimization problem by modelling target and constraint quantities using Gaussian processes and by sequentially selecting interventions via a constrained expected improvement acquisition function. We propose different surrogate models that enable to integrate observational and interventional data while capturing correlation among effects with increasing levels of sophistication. We evaluate \ccbo on artificial and real-world causal graphs showing successful trade off between fast convergence and percentage of feasible interventions.
\end{abstract}

\section{Introduction}
\begin{figure}
\centering
\scalebox{0.89}{
\begin{tikzpicture}[dgraph]
\node[] (a) [label= north:(a)] at (-1.8,1.9) {};
\node[dot,double color fill={gray!70}{pink},shading angle=135] (PKC) [label=north:\acro{pkc}] at (0,2) {};
\node[dot] (Raf)
[fill=brown!70,label={[xshift=0.05cm, yshift=0.09cm]:\acronospace{R}af}] at (-1.1,0.7) {};
\node[dot] (Jnk) [fill=brown!70,label=south:\acronospace{j}nk] at (1.3, 1.3) {};
\node[dot] (P38) [fill=brown!70,label=north:\acronospace{p}38] at (1.3, 2.) {};
\node[dot,double color fill={gray!70}{pink},shading angle=135] (PKA) [fill=pink,label={[xshift=0.5cm, yshift=-0.6cm]:\acro{pka}}] at (0,1.3) {};
\node[dot] (Mek) [fill=gray!70,label=south:\acronospace{m}ek] at (-1.1,0) {};
\node[dot] (Erk) [fill={rgb:red,2;white,1},label=south:\acronospace{e}rk] at (0,0) {};
\node[dot] (Akt) [fill=gray!70, label=south:\acronospace{a}kt] at (0.7, 0.) {};
\draw[line width=0.6pt, \arr](PKC)--(PKA);
\draw[line width=0.6pt, \arr](PKC)--(Raf);
\draw[line width=0.6pt, \arr](PKC)--(Jnk);
\draw[line width=0.6pt, \arr](PKC)--(P38);
\draw[line width=0.6pt, \arr](PKC)to [bend right=+100] (Mek);
\draw[line width=0.6pt, \arr](PKA)--(Raf);
\draw[line width=0.6pt, \arr](PKA)--(Mek);
\draw[line width=0.6pt, \arr](PKA)--(Erk);
\draw[line width=0.6pt, \arr](PKA)--(Akt);
\draw[line width=0.6pt, \arr](PKA)--(Jnk);
\draw[line width=0.6pt, \arr](PKA)--(P38);
\draw[line width=0.6pt, \arr](Raf)--(Mek);
\draw[line width=0.6pt, \arr](Mek)--(Erk);
\end{tikzpicture}}
\hskip0.4cm
\scalebox{0.89}{
\begin{tikzpicture}[dgraph]
\node[] (b) [label= north:(b)] at (-1.8, 1.9) {};
\node[dot] (cal) [fill=gray!70,label=north:\acro{ci}] at (0, 2) {};
\node[dot] (bmr) [fill=brown!70,label=north:\small{\acro{bmr}}] at (-0.8, 2) {};
\node[dot] (age) [fill=brown!70,label=left:\small{Age}] at (-0.6,0.8) {};
\node[dot] (weight) [fill=brown!70,label=left:\small{Weight}] at (0, 1.4) {};
\node[dot] (height) [fill=brown!70,label={[xshift=-0.05cm, yshift=-0.08cm]\small{Height}}] at (0.8, 2) {};
\node[dot, double color fill={brown!70}{pink},shading angle=135] (BMI)[label={[xshift=0.5cm, yshift=-0.4cm]:$\bmi$}] at (1,1.4) {};
\node[dot] (Aspirin) [fill=gray!70,label={[xshift=-0.1cm, yshift=-0.8cm]\small{Aspirin}}] at (0.3,0.6) {};
\node[dot] (Statin) [fill=gray!70,label=south:\small{Statin}] at (-0.6, 0) {};
\node[dot] (PSA) [fill={rgb:red,2;white,1},label=south:\psa] at (1.6,0) {};
\draw[line width=0.6pt, \arr](age)--(weight);
\draw[line width=0.6pt, \arr](bmr)--(weight);
\draw[line width=0.6pt, \arr](cal)--(weight);
\draw[line width=0.6pt, \arr](height)--(weight);
\draw[line width=0.6pt, \arr](height)--(BMI);
\draw[line width=0.6pt, \arr](weight)--(BMI);
\draw[line width=0.6pt, \arr](BMI)--(PSA);
\draw[line width=0.6pt, \arr](BMI)--(Aspirin);
\draw[line width=0.6pt, \arr](age)--(Aspirin);
\draw[line width=0.6pt, \arr](age)--(Statin);
\draw[line width=0.6pt, \arr](Aspirin)--(PSA);
\draw[line width=0.6pt, \arr](Statin)--(PSA);
\draw[line width=0.6pt, \arr](BMI)to [bend right=+20](Statin);
\draw[line width=0.6pt, \arr](age)to [bend left=+40](PSA);
\end{tikzpicture}}
\caption{(a) Protein-signaling network and (b) causal \psa graph. Red, grey and pink nodes indicate target, intervenable, and constrained variables respectively.}
\label{fig:causalgraphs1}
\end{figure}
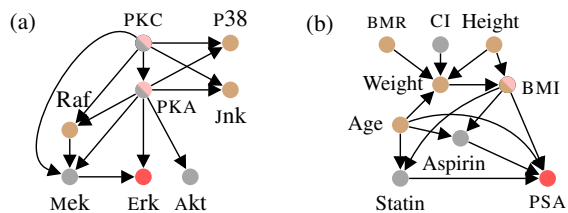

The problem of understanding which interventions in a system optimize a target variable is of relevance in many scientific disciplines, including biology, medicine, and social sciences. Often, the investigator might want to solve this problem while also ensuring that interventions satisfy some constraints. As an example, consider the protein-signalling network of \figref{fig:causalgraphs1}(a), which describes the causal pathways linking several phosphorylated proteins and phospholipids in human immune system cells, including the mitogen-activated protein kinases \erk~studied in cancer therapy \citep{fremin2010basic,sachs2005causal}. 
An investigator might wish to find which variables among $\text{\mek}, \text{\pkc}, \text{\pka}$ and $\text{\akt}$ to perturb and to which levels in order to minimize \erk, under the constraints of not inhibiting \pka and \pkc which play an important role in the functioning of healthy cells. As another example, consider the graph of \figref{fig:causalgraphs1}(b) describing the causal relationships between prostate specific antigen (\psa) and its risk factors, some of which might be due to an existing policy at the time of study \citep{ferro2015use}. An investigator might wish to understand which variables among Aspirin, Statin, and Calories Intake (\acro{ci}) to fix and to which values in order to minimize \psa, under the constraint of keeping \bmi below a certain value.

In this paper, we propose \emph{constrained causal Bayesian optimization} (\ccbo), a sequential approach for efficiently solving such constrained optimization problems in the setting of known causal graph that builds on the literature on causal Bayesian optimization \citep{aglietti2020causal}, constrained Bayesian optimization \citep{gardner2014bayesian}, and multi-task learning with Gaussian processes (\gpstext) \citep{alvarez2011kernels}. Our contributions are summarized as follows:
\begin{itemize}[leftmargin=*]
\setlength{\itemsep}{-6pt}  
\setlength{\parskip}{-4pt}
\setlength{\parsep}{-4pt}
\item We provide a formalization of the constrained optimization problem using the framework of structural causal models (\scms).\\ 
\item We introduce a novel theory for reducing the search space of interventions that eliminates intervention sets of higher cardinality.\\
\item We propose different \gptext surrogate models for the unknown target and constraint quantities that leverage observational data, interventional data and the \scm structure, capturing correlations with increasing level of sophistication. This enables uncertainty quantification thereby speeding up the identification of an optimal intervention while limiting the number of \emph{infeasible} interventions, \ie not satisfy the constraints.\\
\item We propose a constrained expected improvement acquisition function for sequentially selecting interventions based on both output optimization and constraint satisfaction. This enables finding the optimal solution with a smaller number of interventions.\\
\item We evaluate \ccbo on synthetic and real-world graphs with different \scm characteristics showing how it successfully trades off achieving a fast convergence and collecting a high percentage of feasible interventions.
\end{itemize}

\section{Constrained Causal Global Optimization}
We consider a system of observable random variables $\V$ with \emph{target variable} $Y\in \V$ and \emph{intervenable variables} $\I\subseteq \V\backslash Y$\footnote{To simplify the notation we write $Y$ to denote the set $\{Y\}$, and similarly throughout the paper.}, and the problem of finding an intervention set $\X^*\subseteq\I$ and intervention values $\x^*$ that optimize the expectation of $Y$ while ensuring that the expectations of \emph{constrained variables} $\C \subseteq \V \backslash Y$ are above/below certain values. 

We assume that the system can be described by a \emph{structural causal model} (\scm) \citep{pearl2000causality,pearl2016causal} ${\cal M}=\langle \V, \U, \F, p(\U) \rangle$, where $\U$ is a set of exogenous unobserved random variables with distribution $p(\U)=\prod_{U\in\U} p(U)$, and $\F = \{f_V\}_{V\in \V}$ is a set of deterministic functions such that $V=f_V(\pa(V), \U_V)$ with $\pa(V)\subseteq \V\backslash V$ and $\U_V\subseteq\U$, $\forall V\in \V$. ${\cal M}$ has associated a \emph{causal graph} $\graph$ with nodes $\V$, and with a directed edge from $V$ to $W$ if $V\in \pa(W)$, in which case $V$ is called a \emph{parent} of $W$, and a bi-directed edge between $V$ and $W$ if $\U_V \cap \U_W \neq \emptyset$ ($\U_V \cap \U_W$ is an unobserved \emph{confounder} between $V$ and $W$). We assume that $\graph$ is \emph{acyclic}, namely that there are no \emph{directed paths}\footnote{A directed path is a sequence of linked nodes whose edges are directed and point from preceding towards following nodes in the sequence.} starting and ending at the same node. We refer to the joint distribution of $\V$ determined by $p(\U)$, which we denote by $p(\V)$, as \emph{observational distribution}, and to an observation from $p(\V)$ as \emph{observational data sample}. 

An \emph{intervention} on $V\in \I$ setting its value to $v$, denoted by $\doi(V=v)$, corresponds to modifying ${\cal M}$ by replacing $f_V(\pa(V), \U_V)$ with $v$. We refer to the joint distribution of $\V$ in the modified \scm under such an intervention, indicated by $p_{\doi(V=v)}(\V)$, as \emph{interventional distribution}, and to an observation from it as \emph{interventional data sample}. Notice that, in the case of no unobserved confounders, $p_{\text{do}(V=v)}(\V) = \prod_{W \in \V\backslash V} p(W \cond \pa(W))\delta_{V=v}$ with $\delta_{V=v}$ denoting a delta function centered at $v$. An intervention on a set of variables is defined similarly. 

Let $\mu_{\doi(\X=\x)}^Y:= \expectation{p_{\doi(\X=\x)}}{Y}$ denote the expectation of $Y$ w.r.t. the interventional distribution\footnote{This expectation is commonly indicated with $\expectation{}{Y\,|\,\doi(\X=\x)}$ \cite{pearl2000causality}.}, 
which we refer to as the \emph{target effect}. The goal of the investigator is to find a set $\X^*\subseteq \I$ and values $\x^*$ that optimize the target effect while ensuring that $\mu_{\doi(\X^*=\x^*)}^C$, which we refer to as \emph{constraint effect}, is greater/smaller than a threshold $\lambda^C$, $\forall C \in \C \subseteq \V \backslash Y$. Target and constraint effects are unknown.

The constraint for a variable $X\in\C\cap \X$ can directly be enforced by restricting the range of intervention values, $D(X)$, to be in accordance with the threshold. Instead, the constraints for $\CX:=\C \backslash (\C \cap \X)$ need to be included in the optimization problem. The constrained optimization problem can thus be formalized as follows.
\begin{definition}[\ccgo Problem]\label{def:ccgo} 
The \emph{constrained causal global optimization} (\ccgo) problem is the problem of finding a tuple $(\X^*,\x^*)$ such that\footnote{Depending on the application, the investigator might be interested in maximizing $\muX^Y$ and/or ensuring that some or all constraints effects are smaller than the thresholds. In such cases \eqref{alg:ccbo} would need to be adjusted accordingly.
}
\begin{align}
\X^*,\x^* &= \argmin_{\substack{\X \in \powerI, \\ \x \in \domX}} \muX^Y \,\,\textrm{ s.t. } \muX^{\CX} \geq \lCX,
\label{eq:ccgo}
\end{align}
where $\powerI$ indicates the power set\footnote{Excluding $\emptyset$.} of $\I$, $D(\X):=\times_{X \in \X}D(X)$, and $\mu_{\doi(\X=\x)}^{\CX}$ and $\lCX$ denote the constraint effects and corresponding thresholds on all variables in $\CX$. 
\end{definition}
The \ccgo problem extends the \emph{causal global optimization} (\cgo) problem defined in \citet{aglietti2020causal} to incorporate constraints.

\section{Constrained Causal Bayesian Optimization}
\begin{algorithm}[t]
\textbf{Inputs:} $\graph$, $\I, \C, Y$, $\datai:= \{\datai_{\X}\}_{\X\subseteq \I}$, $\datao$, $\lambda^{\C}$, $S$, $T$ \\
$\allmisnull{\C \cup Y} \leftarrow \texttt{cMISReduce}(\graph, \I,\C,Y,\datao)$ \\
Initialise \gpstext~ $g_{\X}(\x)$ $\forall\X \in \allmisnull{\C \cup Y}$ with $\datai_{\X}$ and $\datao$\\
\For{$t=1,\ldots, T$}{
    1. Select intervention $(\X_t,\x_t)$ with \cei \\
    2. Obtain $S$ samples $\{\mathbf{c}_{\X_t}^{(s)},y^{(s)}\}_{s=1}^{S}$ from the distribution $p_{\doi(\X_t=\x_t)}(\C_{\X_t},Y)$, and use them to compute the sample mean estimate $\hat{\mu}_{\doi(\X_t=\x_t)}$\\
    3. Update dataset $\datai_{\X_t}\leftarrow \datai_{\X_t}\cup (\x_t,\hat{\mu}_{\doi(\X_t=\x_t)})$ \\
    4. Update \gptext{s} for $g_{\X_t}(\x) \,|\, \datai_{\X_t}$
    }
\textbf{Output:} 
 $(\X^\star, \x^\star)$ with min feasible $\hat{\mu}^Y_{\doi(\X^\star = \x^\star)}$ over $\datai$
\caption{\ccbo}
\label{alg:ccbo}
\end{algorithm} 
We propose to solve the \ccgo problem using the \emph{constrained causal Bayesian optimization} (\ccbo) method summarized in Algorithm \ref{alg:ccbo}, which assumes known casual graph $\graph$, continuous variables $\V$, and full observation of the system after an intervention.
First, the search space is reduced to a subset $\allmisnull{\C \cup Y}$ of $\powerI$ via the \cmisreduce~procedure described in \secref{sec:rss}. 
Then, the restricted \ccgo problem is solved by modelling, $\forall \X\in \allmisnull{\C \cup Y}$, the unknown target and constraint effects $\muX := (\muX^Y, \muX^{\C_{\X}})$  using Gaussian processes (\gptext{s}) \citep{rasmussen2006gaussian} $g_{\X}(\x):=\big(g^V_{\X}(\x)\big)_{V \in \CX \cup Y}$, as described in \secref{sec:sm}, and via the following sequential strategy. At each trial $t=1,\ldots,T$: (1) a set of intervenable variables $\X_t$ and intervention values $\x_t$ are selected via a constrained expected improvement (\cei) acquisition function that accounts for both the target and constraint effects, as described in \secref{sec:af}; (2) a set of $S$ interventional data samples is obtained and used to get a sample mean estimate $\hat\mu_{\doi(\X_t=\x_t)}$ of $\mu_{\doi(\X_t=\x_t)}$; (3) 
$(\x_t,\hat{\mu}_{\doi(\X_t=\x_t)})$ is added to the interventional dataset $\datai_{\X_t}$ of $\X_t$; (4) the \gpstext~for $g_{\X_t}(\x)$ are updated using $\datai_{\X_t}$. Once the maximum number of trials $T$ is reached, a tuple $(\X^\star, \x^\star)$ giving the minimum feasible target effect value in $\datai: = \{\datai_{\X}\}_{\X \subseteq \I}$ is returned. 

\subsection{Reducing the Search Space}
\label{sec:rss}
The cardinality of the power set, $|\powerI|$, grows exponentially with the cardinality of $\I$, thus solving the \ccgo problem by exploring the entire set could be prohibitively expensive. Even when $|\powerI|$ is small,
reducing the search space would simplify the problem as a smaller number of comparisons between constraint and target effects would be required. In this section, we propose a procedure, which we refer to as \cmisreduce, that leverages invariances of the target and constraint effects w.r.t. different intervention sets to eliminate intervention sets of higher cardinality. This is achieved by extending the theory of minimal intervention sets of \citet{lee2018structural} to account for the presence of constraints (a summary of the procedure and all proofs are given Appendix \ref{secapp:rss}).

The search space is first reduced from $\powerI$ to the set of \emph{constrained minimal intervention sets} (\cmis{s}) relative to $({\C \cup Y},\graph)$, denoted by $\allmis{\C \cup Y}$.  
\begin{definition}[\cmis] \label{def:mis_set} 
A set $\X \subseteq \I$ is said to be a \cmis relative to $(\C \cup Y,\graph)$ if there is no $\X^\prime \subset \X$ with $\C_{\X}=\C_{\X^\prime}$ such that $\mu_{\doi(\X=\x)}^W = \muXprime^W$, where $\x'$ indicates the subset of $\x$ corresponding to variables $\X^{\prime}$, $\forall \x\in D(\X)$, $\forall W \in \C_{\X}\cup Y$, and $\forall$ \scm  with causal graph $\graph$. 
\end{definition}
The following theorem justifies this reduction.
\begin{theorem}[Sufficiency of $\allmis{\C \cup Y}$]
\label{theorem:sufficiency_cmis}
$\allmis{\C \cup Y}$ contains a solution of the \ccgo problem (if a solution exists), $\forall$ \scm with causal graph $\graph$. 
\end{theorem}
\noindent Let $\an(W,\calG)$ be the set of \emph{ancestors} of $W$ in $\calG$, \ie the nodes with a directed path to $W$;  $\an(\W,\graph):=\cup_{W \in \W} \an(W,\graph)$; and $\graph_{\overline{\X}}$ the graph with all incoming links onto all elements of $\X$ removed. The following proposition gives a graphical criterion for identifying $\allmis{\C \cup Y}$. 
\begin{proposition}[Characterization of \cmis]
\label{prop:compute_mis}
$\X\subseteq\I$ is a \cmis relative to $({\C \cup Y},\graph)$  $\iff$ $\X \subseteq \an(\C \cup Y,\graph_{\overline{\X}}) \cup (\C \cap \X)$.
\end{proposition}
If an observational dataset $\datao$ is available, the search space is further reduced from $\allmis{\C \cup Y}$ to $\allmisnull{\C \cup Y}$ by checking, $\forall\X\in \allmis{\C \cup Y}$, if all reducible variables in $\CX$ are null-feasible. 
\begin{definition}[Reducibility] \label{def:reducible}
$C \in \C_{\X}$ is reducible if $\muX^C=\mu^C:=\expectation{p}{C}$, $\forall \x \in D(\X)$.
\end{definition}
\begin{definition}[Null-feasibility] \label{def:null_feas} 
$C \in \C_{\X}$ is null-feasible if $\mu^C\geq\lC$.
\end{definition}
If $\X \cap \an(C,\graph_{\overline{\X}})=\emptyset$, $C\in \C_X$ is reducible. Indeed, if $\X \cap \an(C,\graph_{\overline{\X}})=\emptyset$, by Lemma \ref{lemma:prop_an} in Appendix \ref{secapp:rss} (with $\W_1=\X,\W_2=C,\W_3=\emptyset$) we have $C \indep_{\graph_{\overline{\X}}} \X$, which implies $\mu_{\doi(\X=\x)}^C=\mu^C$ by rule 3 of do-calculus \cite{pearl2000causality}.

If a reducible $C\in\C_X$ is not null-feasible, $\X$ is removed from the search space as it is not a solution of the \ccgo problem. If instead $C$ is null-feasible, all $\X^{\prime}\supset \X$ in $\allmis{\C \cup Y}$ that are of the form $\X^\prime=\X\cup\X_1$ where $\X_1 \cap \an(\CX \backslash C \cup Y,\graph_{\overline{\X^{\prime}}})=\emptyset$, and for which (i) $\C_{\X^\prime}=\CX$ or (ii) all variables in $\CX \backslash \C_{\X^\prime}$ are reducible and null-feasible, are eliminated from the search space. Indeed, in such cases, if $\X^\prime$ were a solution of the \ccgo problem, then $\X$ would also be a solution as: (a) $\forall W \in (\CX\backslash C)\backslash (\C\cap \X_1) \cup Y$, $\mu_{\doi(\X^\prime=\x^\prime)}^W = \mu_{\doi(\X=\x)}^W$ (Lemma \ref{lemma:prop_an} in Appendix \ref{secapp:rss} (with $\W_1=\X_1,\W_2=\CX \backslash C \cup Y,\W_3=\X$) gives $\CX \backslash C \cup Y\indep_{\graph_{\overline{\X^{\prime}}}} \X_1 \,|\,\X$, which implies $\mu_{\doi(\X^\prime=\x^\prime)}^W = \mu_{\doi(\X=\x)}^W$ by rule 3 of do-calculus); (b) $\forall W\in \C \cap \X_1$ the constraint effects are satisfied for $\X$ as $\C \cap \X_1 = \CX \backslash \C_{\X^\prime}$.

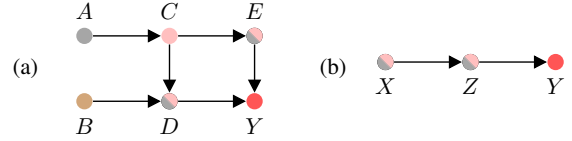
\begin{figure}[t]
\centering
\scalebox{0.87}{
\begin{tikzpicture}[dgraph]
\node[] (a) [label= north:(a)] at (-2.2, 0.1) {};
\node[dot] (A) [fill=gray!70,label=north:$A$] at (-1.3,1) {};
\node[dot] (C) [fill=pink,label=north:$C$] at (0,1) {};
\node[dot,double color fill={gray!70}{pink},shading angle=135] (E) [label=north:$E$] at (1.3,1) {};
\node[dot] (B) [fill=brown!70,label=south:$B$] at (-1.3,0) {};
\node[dot,double color fill={gray!70}{pink},shading angle=135] (D) [label=south:$D$] at (0,0) {};
\node[dot] (Y) [fill={rgb:red,2;white,1},label=south:$Y$] at (1.3,0) {};
\draw[line width=0.6pt, \arr](A)--(C);
\draw[line width=0.6pt, \arr](C)--(E);
\draw[line width=0.6pt, \arr](E)--(Y);
\draw[line width=0.6pt, \arr](B)--(D);
\draw[line width=0.6pt, \arr](D)--(Y);
\draw[line width=0.6pt, \arr](C)--(D);
\hskip0.6cm
\node[] (b) [label= north:(b)] at (-2.1+4, 0.1) {};
\node[dot,double color fill={gray!70}{pink},shading angle=135] (X) [label=south:$X$] at (-1.3+4,0.6) {};
\node[dot,double color fill={gray!70}{pink},shading angle=135] (Z) [label=south:$Z$] at (0+4,0.6) {};
\node[dot] (Y) [fill={rgb:red,2;white,1},label=south:$Y$] at (1.3+4,0.6) {};
\draw[line width=0.6pt, \arr](X)--(Z);
\draw[line width=0.6pt, \arr](Z)--(Y);
\end{tikzpicture}}
\caption{(a) Causal graph with $\I=\{A,D,E\}$ and $\C=\{C,D,E\}$. (b) Causal graph with $\I=\C=\{X,Z\}$.}
\label{fig:causalgraphs2-3}
\end{figure}

\noindent \runexample We describe the search space reduction for the casual graph of \figref{fig:causalgraphs2-3}(a). In this graph $\allmis{\C \cup Y}=\powerI=\{\{A\},\{D\},\{E\},\{A,D\},\{A,E\},\{D,E\},\{A,D,E\}\}$, as $\X \subseteq \an(\C\cup Y,\graph_{\overline{\X}}) \cup (\C \cap\X)$, $\forall \X\in\powerI$ (\eg $A\in \text{an}\big(\C\cup Y,\graph_{\overline{A}}\big) \cup (\C \cap A)$). Consider $\X=\{D,E\}\in\allmis{\C \cup Y}$ and $C$. $\X \cap \an\left(C,\graph_{\overline{\X}}\right)=\emptyset$, and therefore $C$ is reducible. If $C$ is not null-feasible, $\X$ is removed from the search space. If instead $C$ is null-feasible, the set $\X^{\prime}=\{A,D,E\}\supset\X$ in $\allmis{\C \cup Y}$ is of the form
$\X^{\prime}=\X \cup A$ where $A\not\in\text{an}(\CX \backslash C \cup Y,\graph_{\overline{\X^{\prime}}})$. 
Therefore $\X^{\prime}$ is removed from the search space. All other sets in $\allmis{\C \cup Y}$ have constrained variables that are non reducible and thus need to be included in $\allmisnull{\C \cup Y}$.

\subsection{Gaussian Processes Surrogate Models}
\label{sec:sm}
For each $\X \in \allmisnull{\C \cup Y}$ and $V\in\C_{\X} \cup Y$, we model $\mu^V_{\doi(\X=\x)}$  with a \gptext $g^V_{\X}(\x) \sim \gp (m^V_{\X}(\x), S^V_{\X}(\x, \x'))$, as \gptext{s} allow constructing flexible surrogate models while enabling uncertainty quantification and closed form updates. We propose one single-task \gptext (\stgp) and two multi-task \gptext{s} (\mtgp and \Gmtgp) which capture the correlation across effects with increasing level of sophistication. 

For $V\neq W$, the single-task \gptext treats $g^V_{\X}(\x)$ and $g^W_{\X}(\x')$ as independent, while the multi-task \gpstext~model their correlation via a covariance matrix $S^{V,W}_{\X}(\x_1, \x_2)$, with $(i,j)$-th element given by $\expectation{}{g^V_{\X}(\x_i)g^W_{\X}(\x_j)}-\expectation{}{g^V_{\X}(\x_i)}\expectation{}{g^W_{\X}(\x_j)}$, either by assuming a common latent structure among the \gpstext~(\mtgp) or, for the setting of no unobserved confounders and under the assumption $V=f_{V}(\pa(V)) + U_V$ with $p(U_V)={\cal N}(0,\sigma^2_V)$, by explicitly exploiting the \scm structure (\Gmtgp). We propose different prior parameters constructions for \stgp and \mtgp, including one that leverages the availability of an observational dataset  $\datao$ (\stgpcausal and \mtgpcausal). The different \gptext constructions allow the investigator to model the target and constraint effects in both settings where no information about the system is available and black-box models are preferred (\stgp and \mtgp) and settings in which one can leverage different sources of information and integrate them in a structured prior formulation that quantifies uncertainty in a principled way (\stgpcausal, \mtgpcausal, and \Gmtgp). 

For all surrogate models, after an intervention $(\X,\x)$ is selected by the acquisition function, as described in \secref{sec:af}, a set of $S$ interventional data samples $\{\mathbf{c}^{(s)}_{\X},y^{(s)}\}_{s=1}^{S}$ from $p_{\doi(\X=\x)}(\C_{\X},Y)$ is obtained. For each $V\in\C_{\X} \cup Y$, this set is used to form a sample mean estimate $\hat{\mu}_{\doi(\X=\x)}^V$ of $\mu_{\doi(\X=\x)}^V$, which is treated as a noisy realization of $g^V_{\X}(\x)$ with additive Gaussian noise. The tuple $(\x,\hat{\mu}_{\doi(\X=\x)})$ is then added to $\datai_{\X}$ and the posterior distribution of $g_{\X}(\x)$, denoted by $\tau(g_{\X}\,|\,\datai_{\X})$, is computed via standard \gptext updates \cite{rasmussen2006gaussian}. Full details are given in Appendix \ref{secapp:surrogatemodels}.

\textbf{\stgp.} 
For the single-task \gptext, we either assume $m^V_{\X}(\x)=0$ and radial basis function (\acro{rbf}) kernel $S^V_{\X}(\x, \x') = \sigma^2_f\exp(-\frac{||\x - \x'||^2}{2l^2})$ with $(\sigma^2_f, l)$ hyper-parameters, or the prior construction proposed in \cbo \cite{aglietti2020causal}. In the latter case, $\datao$ is used to obtain estimates of the target and constraint effects which are then used as prior mean functions (we refer to this variant as \stgpcausal).

\textbf{\mtgp.} 
Our first multi-task \gptext, inspired by the linear coregionalization model of \citet{alvarez2011kernels}, assumes that the target and constraint \gpstext~are linear combinations of shared independent \gptext{s}, \ie $g^V_{\X}(\x) = \sum_{q=1}^Q a^V_{\X,q} u_{\X,q}(\x)$ with $u_{\X,q}(\x) \sim \gp(m_{\X,q}(\x), S_{\X,q}(\x,\x^\prime))$. 
In this case, the variance and covariance terms across functions and intervention values are given by $S^V_{\X}(\x, \x')=\sum_{q=1}^Q \left(a^V_{\X,q}\right)^2S_{\X,q}(\x, \x')$ and $S^{V,W}_{\X}(\x, \x')=\sum_{q=1}^Q  a^V_{\X,q} a^W_{\X,q} S_{\X,q}(\x, \x')$ respectively. The scalar parameters $a^V_{\X,q}$ 
are learned with a standard type-2 \acro{ml} approach together with the kernel hyper-parameters.
We either consider $m_{\X,q}(\x)=0$ and \acro{rbf} kernel for each $S_{\X, q}(\x, \x')$, or the prior construction using $\datao$ as in \stgpcausal (we refer to this variant as \mtgpcausal).

\textbf{\Gmtgp.}
For the setting of no unobserved confounders and under the assumption $V=f_{V}(\pa(V)) + U_V$ with $p(U_V)={\cal N}(0,\sigma^2_V)$, we propose to model each $f_V$ as an independent \gptext with an \acro{rbf} kernel $f_V(\pa(v)) \sim \gp (0, S^V_{\acro{rbf}}(\pa(v),\pa(v)'))$, where $\pa(v)$ denotes a value taken by $\pa(V)$, and to fit it using $\datao$. 
Consider the intervention $\doi(\X=\x)$ and let $\U^{\text{an}(V)}_{\X}$ denote the subset of $\U$ corresponding to the ancestors of $V$ in $\graph$ that are not d-separated from $V$ by $\X$, and similarly for $f^{\text{an}(V)}_{\X}$. We can write $V$ as an explicit function of $\U^{\text{an}(V)}_{\X}$ and $f^{\text{an}(V)}_{\X}$ by recursively replacing parents with their functional form in the modified \scm under $\doi(\X=\x)$.
Taking the expectation w.r.t. $\U^{\text{an}(V)}_{\X}$ therefore gives $g^{V}_{\X}(\x)$ as a function of $f^{\text{an}(V)}_{\X}$.

For example, for the causal graph in \figref{fig:causalgraphs2-3}(b) with $X=U_X$, $Z=f_Z(X) + U_Z$ and $Y=f_Y(Z)+U_Y$, we can write $Y =f_Y(f_Z(X) + U_Z)+U_Y$, which gives $g^Y_{X}(x) = \expectation{p(U_Z)}{f_Y(f_Z(x) + U_Z)}$ and $g^Z_{X}(x)=f_Z(x)$. We can then obtain realizations $\{g^{V,(s)}_{\X}\}_{s=1}^{S'}$ of $g^V_{\X}$ by using samples of $f^{\text{an}(V)}_{\X}$ and $\U^{\text{an}(V)}_{\X}$ to form an approximation of $S^{V,W}_{\X}(\x_1, \x_2)$ as $
\frac{1}{S'} \sum_{s=1}^{S'} g^{V,(s)}_{\X}(\x_1)g^{W,(s)}_{\X}(\x_2)-\Big(\frac{1}{S'}\sum_{s=1}^{S'} g^{V,(s)}_{\X}(\x_1)\Big)
\Big(\frac{1}{S'}\sum_{s=1}^{S'} g^{W,(s)}_{\X}(\x_2)\Big)
$, and similarly for $S^V_{\X}(\x, \x')$ and $m_{\X}^V(\x)$.

\subsection{Acquisition Functions}
\label{sec:af}
To select interventions accounting for both the target and constraint effects we propose acquisition functions based on those used for constrained \bo \citep{gardner2014bayesian} and noisy \bo \cite{letham2019constrained}. We define the constrained expected improvement (\cei) per unit of intervention cost at an input point $\x$ as $\cei_{\X}(\x):=\mathbb{E}_{\tau(g_{\X}\,|\,\datai_{\X})}\left[\frac{\max(0, g^Y - g^Y_{\X}(\x))}{|\X|}\mathbb{I}_{g^{\C_{\X}}_{\X} \geq \lambda^{\CX}}\right]$, where $g^Y$ is the minimum feasible value attained by $g^Y_{\X}$ across interventional dataset $\mathcal{D}^I$ and $\mathbb{I}_{g^{\C_{\X}}_{\X} \geq \lambda^{\CX}}$ is an indicator variable equal to one if $g^{\C_{\X}}_{\X} \geq \lambda^{\CX}$ and to zero otherwise. The division by $|\X|$ is due to assuming an intervention cost for $\X$ equal to its cardinality. Alternative costs can be considered as long as they are greater than zero.

\textbf{\stgp.}
When using a single-task \gptext as surrogate model, we can exploit the factorization $\tau(g_{\X}\,|\,\datai_{\X}) = \tau(g^Y_{\X}\cond\datai_{\X})\prod_{C \in \CX}\tau(g^{C}_{\X}\cond\datai_{\X})$ to obtain $\cei_{\X}(\x)=\mathbb{E}_{\tau(g^Y_{\X}\cond \datai_{\X})}\left[\max(0, g^Y - g^Y_{\X}(\x))\right]\prod_{C\in\CX}\mathbb{P}(g^C_{\X} \geq \lambda_C)$ where $\mathbb{P}(g^C_{\X} \geq \lambda_C) = \Big[1 - \Phi\left(\frac{\lambda_C - m_{\X}^C(\x)}{S_{\X}^C(\x, \x)}\right)\Big]$ with $\Phi(\cdot)$ denoting the \cdf of a standard Gaussian random variable. In the case of noiseless observations of $g^{V_k}_{\X}(\x)$, $g^Y$ is known thus we can compute the first term in closed form as $\left(g^Y - m^Y_{\X}(\x)\right)\Phi\left(\frac{g^Y - m^Y_{\X}(\x)}{S^Y_{\X}(\x, \x)}\right) + S^Y_{\X}(\x, \x) \phi\left(\frac{g^Y - m^Y_{\X}(\x)}{S^Y_{\X}(\x, \x)}\right)$ with $\phi(\cdot)$ denoting the \acro{pdf} of a standard Gaussian random variable. In the case of noisy observations of $g^{V_k}_{\X}(\x)$, $g^Y$ is unknown as we observe noisy values of the target effects. We can get an estimate of $g^Y$ using samples from $\tau(g_{\X}\,|\,\datai_{\X})$ for all $\X \in \allmisnull{\C \cup Y}$, and use the estimate to compute $\cei_{\X}(\x)$ in closed form using the terms derived above. For every $\X \in \allmisnull{\C \cup Y}$ the values of $\cei_{\X}(\x)$ obtained with different $g^Y$ are then averaged to integrate out the uncertainty on the optimal feasible value observed across all intervention sets.

\textbf{\mtgp.}
When using a multi-task model, $\cei_{\X}(\x)$ cannot be computed in closed form as $\tau(g_{\X} \cond \datai_{\X})$ does not factorize. We thus compute it via Monte Carlo integration with a similar procedure as for the noisy \stgp setting.

\subsection{Computational Aspects}
The computational cost of \ccbo is dominated by the algebraic operations needed to compute the posterior parameters for the \gptext models of the target and constraints effects. Let $\Nint$ denote the largest among the cardinalities of the interventional datasets collected for the sets in $\allmisnull{\C \cup Y}$, \ie $\Nint = \max_{\X \in \allmisnull{\C \cup Y}}{|\mathcal{D}^I_{\X}|}$, 
and let $\ntasks$ denote the highest among the number of target and constraint effects for the sets in $\allmisnull{\C \cup Y}$, \ie $\ntasks = \max_{\X \in \allmisnull{\C \cup Y}}{1 + |C_{\mathbf{X}}|}$. As the \gptext{s} corresponding to different intervention sets are updated independently, the computational cost scales as $O(\Nint^3)$ when using a single-task model and as $O(\ntasks^3 \Nint^3)$ when using a multi-task model. Independent \gptext{s} updates also imply linear scaling of the computational complexity with respect to the cardinality of $\allmisnull{\C \cup Y}$. Therefore, a larger $\graph$ might induce a higher number of sets to explore, but does not induce higher computational complexity. 

In terms of convergence to the true global optimum, \ccbo inherits the properties of \bo algorithms. While any alternative acquisition function for constrained \bo can be used within \ccbo, in this work we resort to a constrained expected improvement function due to its computationally tractability. The expected improvement acquisition function was shown to have strong theoretical guarantees \cite{vazquez2010convergence, bull2011convergence} while performing well in practice \cite{snoek2012practical}. However, performance guarantees have yet to be established for the constrained version. 

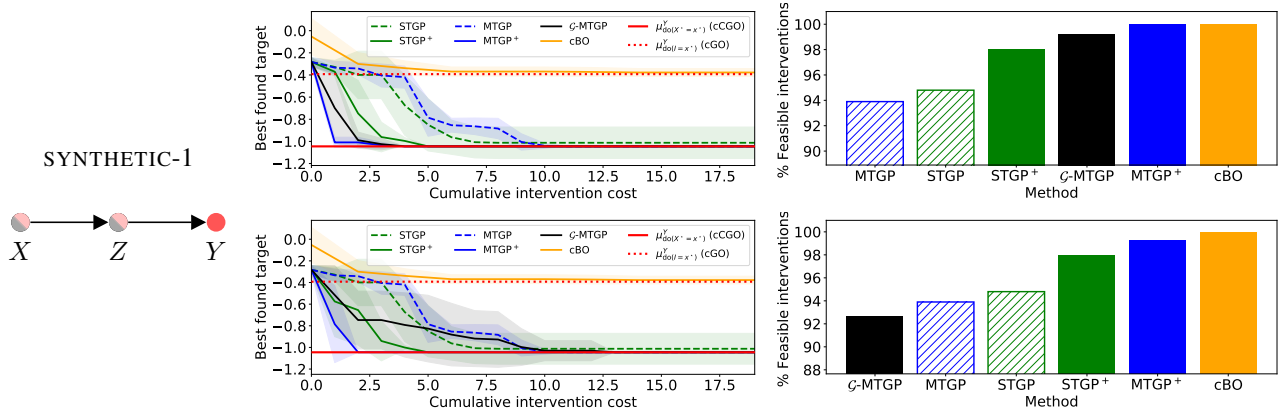
\begin{figure*}
    \centering
\begin{minipage}{.19\textwidth}
    \centering
    \begin{tikzpicture}[dgraph]
    \node[] (a) [label= north:\synone] at (0, 0.5) {};
    \node[dot,double color fill={gray!70}{pink},shading angle=135] (X) [label=south:$X$] at (-1.3,0) {};
    \node[dot,double color fill={gray!70}{pink},shading angle=135] (Z) [label=south:$Z$] at (0,0) {};
    \node[dot] (Y) [fill={rgb:red,2;white,1},label=south:$Y$] at (1.3,0) {};
    \node[] (a) at (0, -0.7) {};
    \draw[line width=0.6pt, \arr](X)--(Z);
    \draw[line width=0.6pt, \arr](Z)--(Y);
    \end{tikzpicture}
\end{minipage}
\begin{minipage}{0.4\textwidth}
    \centering
    \includegraphics[width=1.\textwidth]{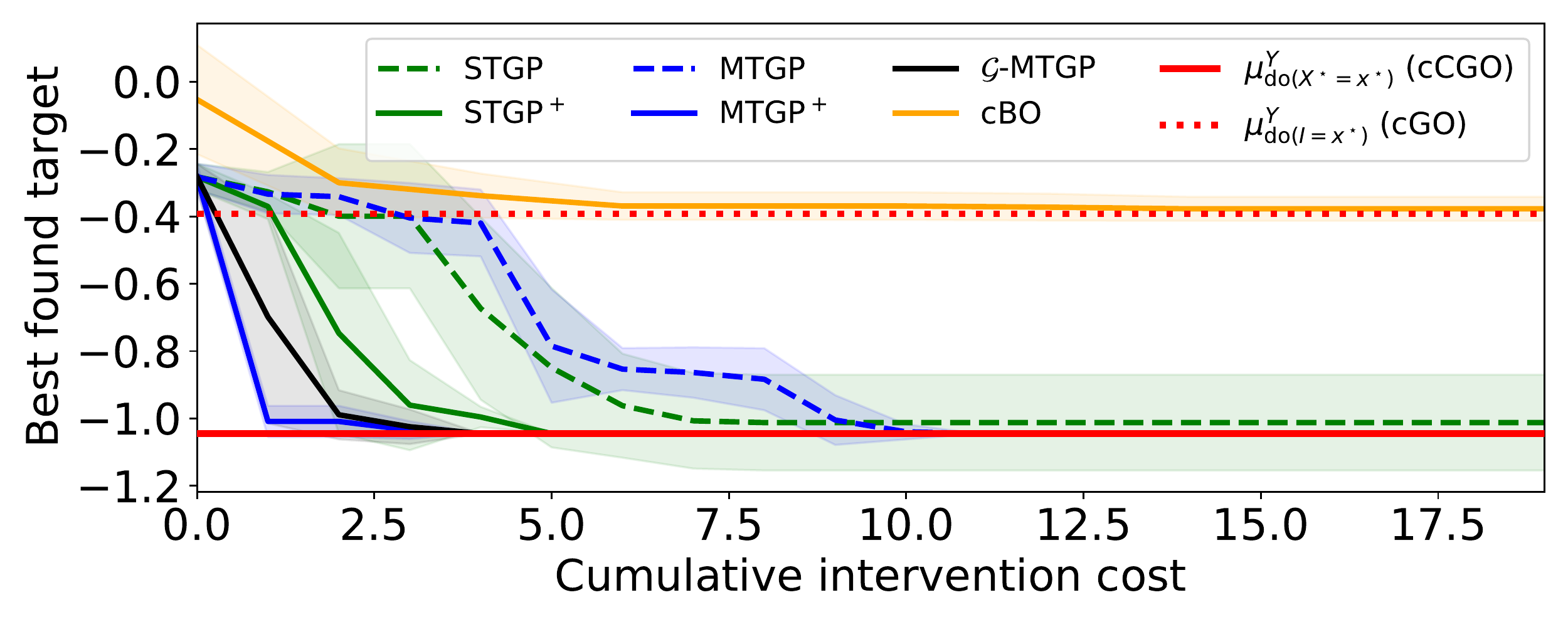}
    \includegraphics[width=1.\textwidth]{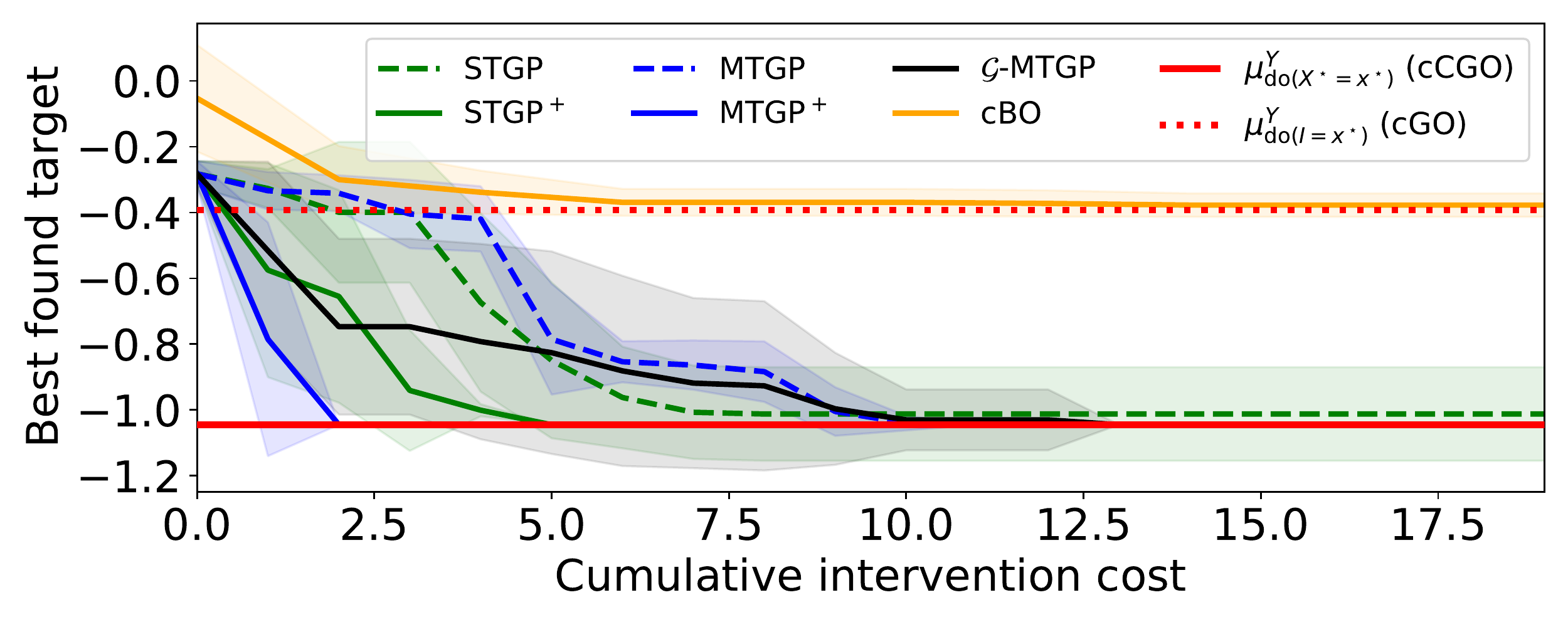}
\end{minipage}
\begin{minipage}{0.4\textwidth}
    \centering
    \includegraphics[width=1.\textwidth]{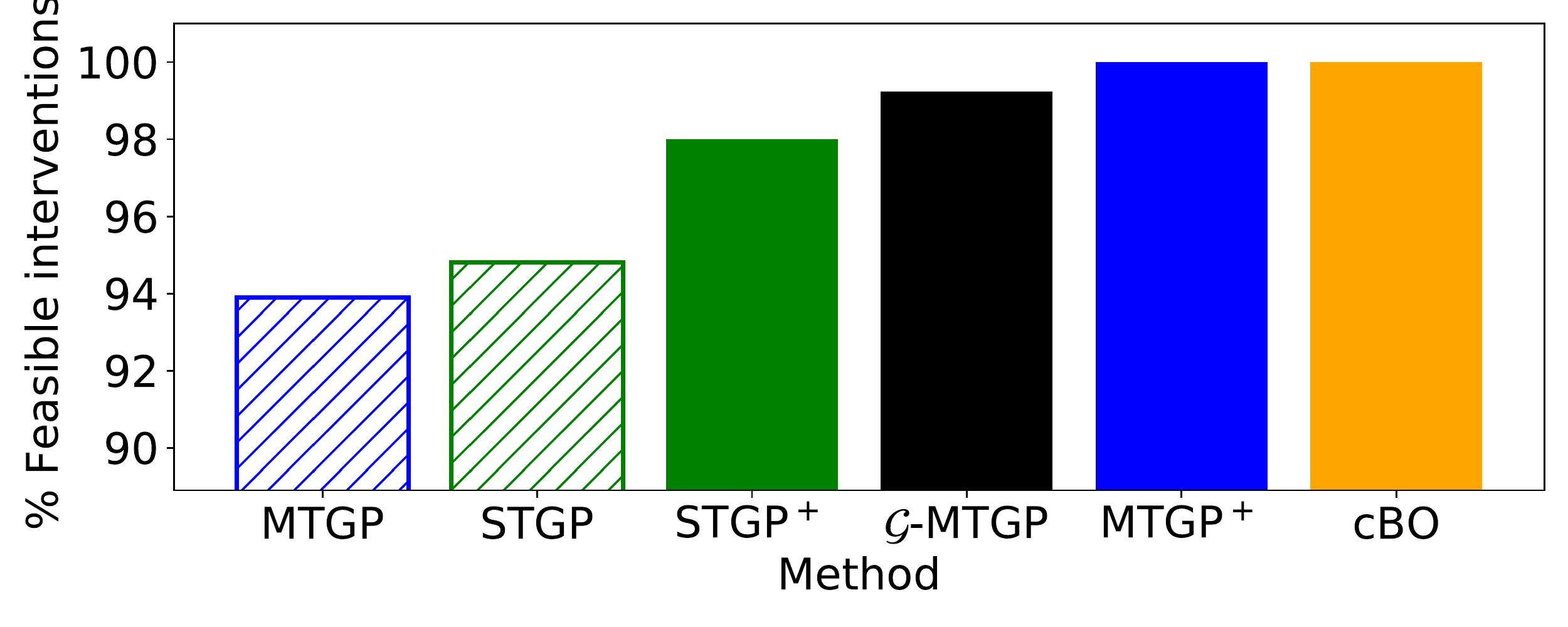}
    \includegraphics[width=1.\textwidth]{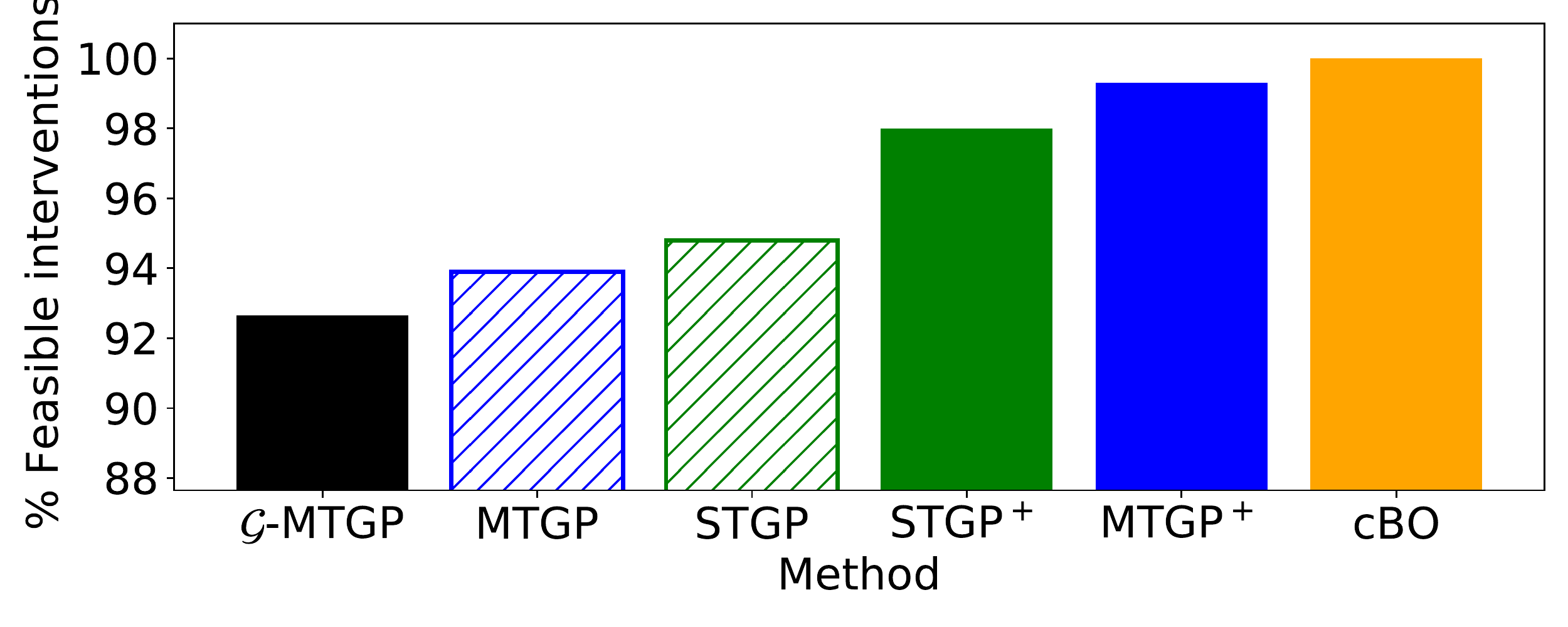}
\end{minipage}
\caption{\synone~with $\Nobs=500$ (top row) and $\Nobs=100$ (bottom row) and $\lambda^Z = 2$. \textit{Left}: Causal graphs. \textit{Center}: Convergence to the \ccgo (solid red line) and \Cgo (dotted red line) optima.
Lines give average results across different initialization of $\datai$. Shaded areas represent $\pm$ standard deviation. 
\textit{Right}: Average percentage of feasible interventions collected over trials.}
\label{fig:synone_N500_N100_lambda2}
\end{figure*}

\subsection{Related Work}
\ccbo is related to the vast literature on constrained \bo methods, which find feasible solutions either by using an acquisition function that accounts for the probability of feasibility \cite{gardner2014bayesian, gelbart2014bayesian, griffiths2017constrained, hernandez2014predictive, keane2008engineering, schonlau1998global, sobester2014engineering, tran2019pbo}, or by transforming the constrained problem into an unconstrained one \cite{ariafar2019admmbo, picheny2016bayesian}, or by exploiting trust regions \citep{eriksson2021scalable}. While these works disregard the causal aspect of the optimization and model the unknown functions independently, multi-task surrogate models have been considered by multi-objective \bo methods \citep{dai2020multi, feliot2017bayesian, hakhamaneshi2021jumbo, mathern2021multi, swersky2013multi} or, more recenlty, in the context of safe \bo \citep{bergmann2020safe, berkenkamp2016safe, berkenkamp2021bayesian, kirschner2019adaptive, sui2015safe, sui2018stagewise}. \ccbo is also related to the works combining causality with decision-making frameworks which have mainly focused on finding optimal interventions using observational data \citep{atan2018deep, haakansson2020learning, zhang2012robust} or on designing interventions for causal discovery \citep{tigas2022interventions}. The idea of identifying optimal interventions through sequential experimentation has been explored in causal bandits \cite{lattimore2016causal}, causal reinforcement learning \cite{zhang2020designing} and, more recently, in \cbo \cite{aglietti2020causal} and model-based causal \bo (\acro{mcbo}, \citet{sussex2023model}). All these works tackle unconstrained settings and disregard the effects that interventions optimizing a target variable might have on the constrained variables.

\section{Experiments}
\label{sec:exp_section}
We evaluate \ccbo with surrogate models \stgp, \stgpcausal, \mtgp, \mtgpcausal, and \Gmtgp on the causal graphs of \figref{fig:causalgraphs2-3}(b) (\synone), \figref{fig:causalgraphs2-3}(a) (\syntwo), \figref{fig:causalgraphs1}(b) (\health), and \figref{fig:causalgraphs1}(a) (\protein). We assume an initial $\datai$ that includes one point per intervention set, and consider different settings with respect to the \scm characteristics, null-feasibility,  $|\allmisnull{\C \cup Y}|$, and  $\Nobs=|\datao|$. See Appendix \ref{secapp:Exp} for full experimental details\footnote{Code for reproducing the experiments is available at \texttt{https://github.com/deepmind/ccbo}.}. 

To the best of our knowledge, there are no other constrained methods in the literature that exploit causal structure. Therefore, we compare to the closest method aiming at solving the non-causal \emph{constrained global optimization} (\Cgo) problem, namely the constrained \bo algorithm (\Cbo) proposed by \citet{gardner2014bayesian}. \Cbo intervenes on all variables in $\I$ simultaneously, models the target and constraints effects via independent \gptext{s}, and selects interventions by maximizing the constrained \ei acquisition function.

\begin{figure*}
    \centering
\begin{minipage}{.19\textwidth}
    \centering
    \begin{tikzpicture}[dgraph]
    \node[] (a) [label= north:\syntwo] at (0, 1.5) {};
    \node[dot] (A) [fill=gray!70,label=north:$A$] at (-1.3,1) {};
    \node[dot] (C) [fill=pink,label=north:$C$] at (0,1) {};
    \node[dot,double color fill={gray!70}{pink},shading angle=135] (E) [label=north:$E$] at (1.3,1) {};
    \node[dot] (B) [fill=brown!70,label=south:$B$] at (-1.3,0) {};
    \node[dot,double color fill={gray!70}{pink},shading angle=135] (D) [label=south:$D$] at (0,0) {};
    \node[dot] (Y) [fill={rgb:red,2;white,1},label=south:$Y$] at (1.3,0) {};
    \node[] (a) at (0, -0.7) {};
    \draw[line width=0.6pt, \arr](A)--(C);
    \draw[line width=0.6pt, \arr](C)--(E);
    \draw[line width=0.6pt, \arr](E)--(Y);
    \draw[line width=0.6pt, \arr](B)--(D);
    \draw[line width=0.6pt, \arr](D)--(Y);
    \draw[line width=0.6pt, \arr](C)--(D);
    \end{tikzpicture}
\end{minipage}
\begin{minipage}{0.4\textwidth}
    \centering
    \includegraphics[width=1.\textwidth]{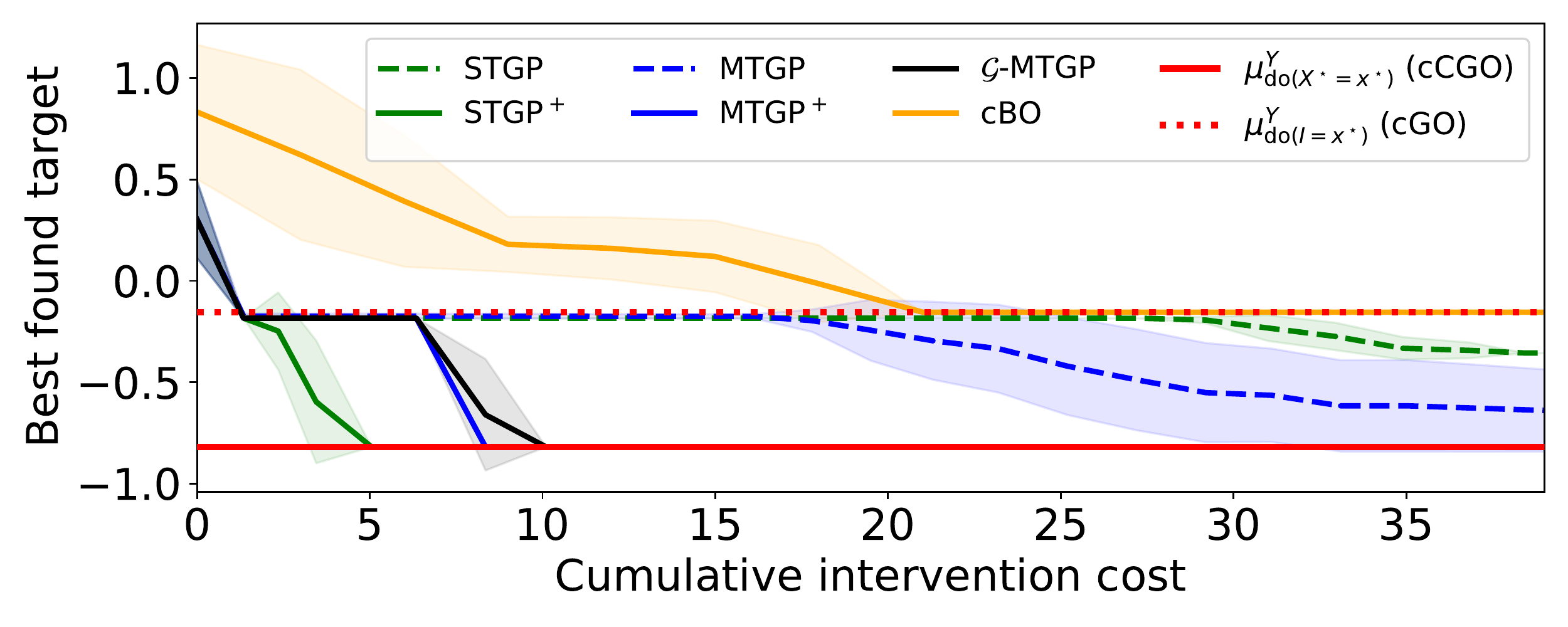}
    \includegraphics[width=1.\textwidth]{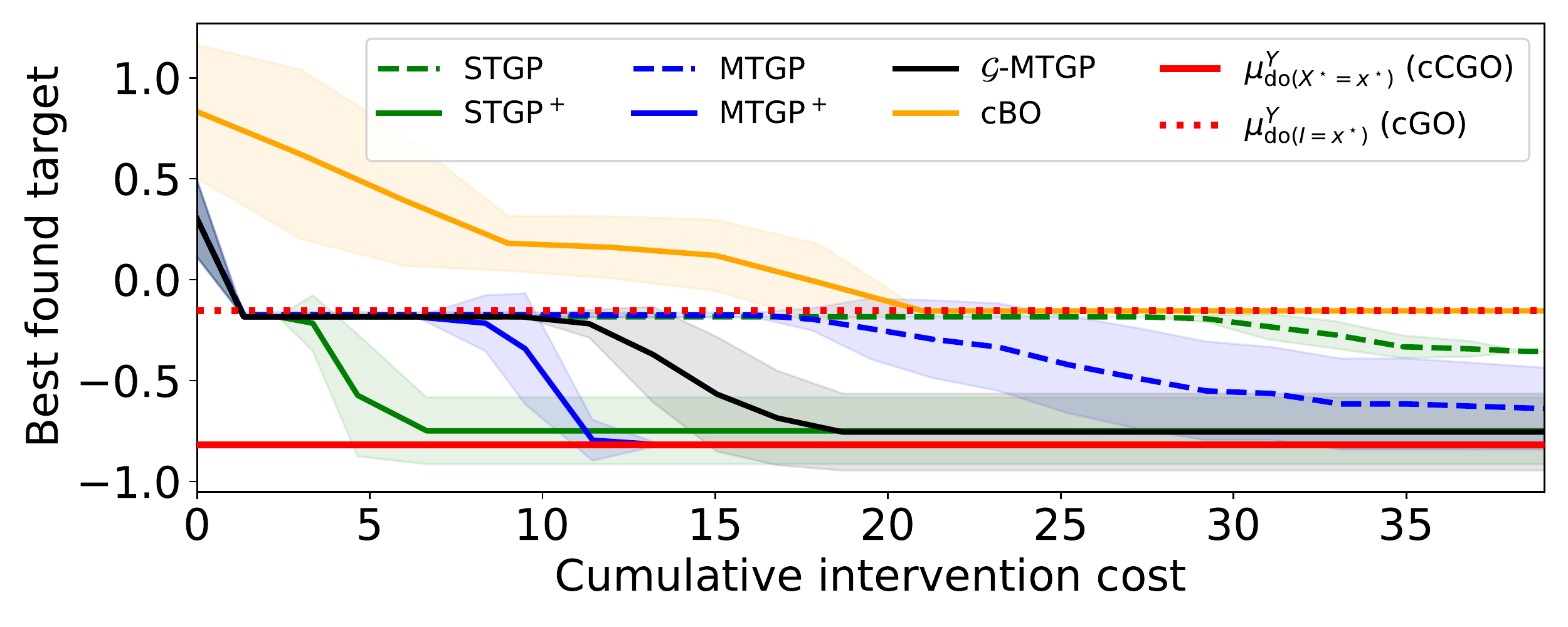}
\end{minipage}
\begin{minipage}{0.4\textwidth}
    \centering
    \includegraphics[width=1.\textwidth]{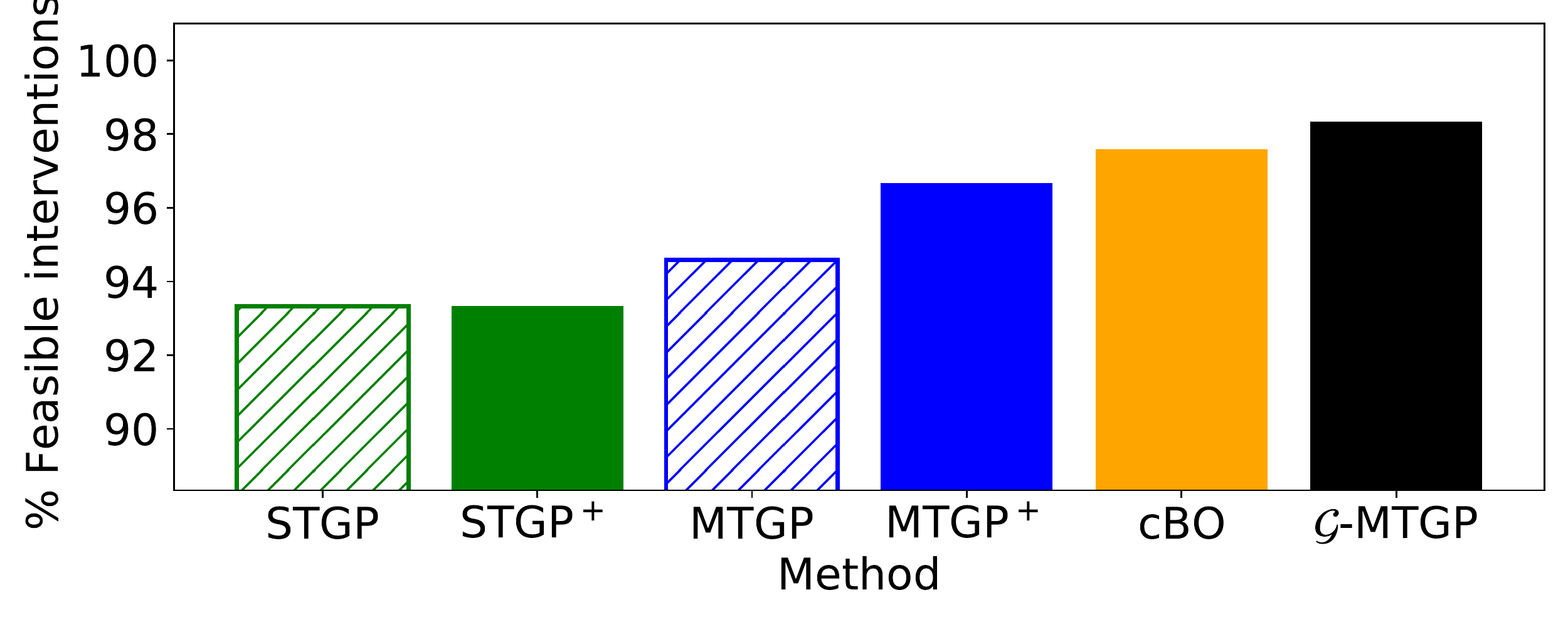}
    \includegraphics[width=1.\textwidth]{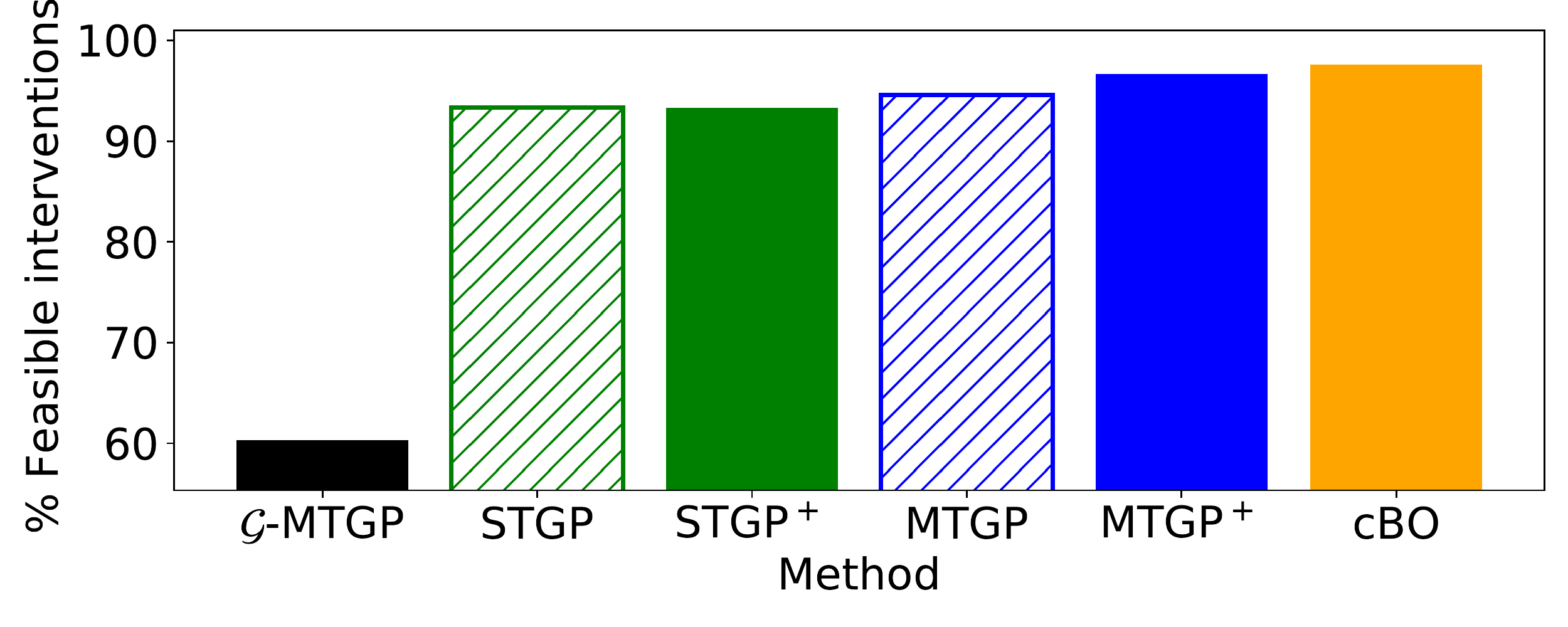}
\end{minipage}
\caption{\syntwo~with $\Nobs=500$ (top row) and $\Nobs=100$ (bottom row). \textit{Left}: Causal graphs. \textit{Center}: Convergence to the \ccgo (solid red line) and \Cgo (dotted red line) optima.
Lines give average results across different initialization of $\datai$. Shaded areas represents $\pm$ standard deviation. \textit{Right}: Average percentage of feasible interventions collected over trials.}
\label{fig:syntwo_N500_N100}
\end{figure*}

For each model, we show the average convergence to the optimal target effect ($\pm$ standard deviations given by shaded areas), and the mean percentage of feasible interventions collected over trials across 20 initialization of $\datai$ and for varying values of $\Nobs$. The combination of these two metrics allows us to understand how each method balances convergence speed and feasibility. To gain more insights into this balance, we also include in the analysis \cbo, an algorithm that randomly picks interventions (\acro{random}) and, for \synone~and \syntwo, \acro{mcbo}(see Appendix \ref{secapp:Exp}). Notice that, as \cbo and \acro{mcbo} solve the \cgo problem, a comparison with them is only relevant for cases in which the \cgo and \ccgo optima are equal. 

\subsection{Synthetic Causal Graphs}
\textbf{\synone.} 
For the \synone~graph with $\lambda^X = 1$, $\lambda^Z = 2,10$, and $\Nobs=500, 100, 10$ observational data samples from the \scm in Appendix \ref{sec:appsynthetic1}, we obtain $\allmisnull{\C \cup Y} = \{\{X\}, \{Z\}\}$. 

With $\Nobs=500$ and $\lambda^Z = 2$, using a surrogate model that exploits $\datao$ when constructing the prior \gptext parameters for functions in $g_{\X}(\x)$, as in \stgpcausal, \mtgpcausal and \Gmtgp, leads to faster average convergence (\figref{fig:synone_N500_N100_lambda2}, top row). The convergence speed is further improved when capturing the covariance structure among the target and constraint effects as done by \mtgpcausal and \Gmtgp. High convergence speed for \stgpcausal, \mtgpcausal and \Gmtgp is associated with a higher percentage of feasible interventions collected over trials compared to the other methods. Similar results are obtained with $\Nobs=100$ and $\lambda^Z = 2$ (\figref{fig:synone_N500_N100_lambda2}, bottom row) and with $\Nobs=10$ and $\lambda^Z = 2$ (\figref{fig:synone_N10_lambda2} in Appendix \ref{sec:appsynthetic1}). In these cases, the lower cardinality of $\datao$ leads to a less accurate estimation of the effects which affects the  prior mean functions for \stgpcausal, \mtgpcausal and \Gmtgp, and kernel function for \Gmtgp. In turns, this leads to a lower number of feasible interventions and convergence speed, particularly for \Gmtgp. As a consequence, \mtgpcausal outperforms all other models in this setting. By intervening on all constrained variables ($\I = \C$) with intervention domains that are in accordance with the threshold values, \Cbo only collects feasible interventions. However, it converges to the higher \Cgo optimum, as causal structure is disregarded.

Interestingly, the results obtained with $\Nobs=500$ and $\lambda^Z=10$ (\figref{fig:all_synone_N500_lambda2_lambda10} (top row) in Appendix \ref{sec:appsynthetic1}) for which the \ccgo and \cgo optima are equal, making \cbo, \acro{mcbo} and \ccbo comparable, show that both \cbo and \acro{mcbo} converge at a slower pace compared to \mtgpcausal and \Gmtgp while collecting a lower number of feasible interventions. Indeed, \cbo and \acro{mcbo} disregard the values taken by the constrained variables. Finally, when $\lambda^Z=10$, \Gmtgp and \mtgpcausal show fast convergence performance and high percentage of feasible interventions collected over trials. This is even when $\Nobs=100$ (\figref{fig:synone_N100_lambda10} in Appendix \ref{sec:appsynthetic1}).

\textbf{\syntwo.} For the \syntwo~graph with $\lambda^{C}=10, \forall C \in \C$, and $\Nobs=500, 100, 10$ observational data samples from the \scm in Appendix \ref{sec:appsynthetic2}, we obtain null-feasibility $\forall C \in \C$ and thus $\allmisnull{\C \cup Y} = \powerI \backslash \{A, D, E\}$. As in \synone, when $\Nobs=500$ using a prior \gptext construction that exploits $\datao$ leads to faster convergence (\figref{fig:syntwo_N500_N100}, top row). In particular, the accurate estimation of the target and constraint effects with $\datao$, which are used as prior mean functions, favours the prior formulation of \stgpcausal which identifies the optimal interventions after $\sim5$ trials. However, by disregarding the correlation among the effects, \stgpcausal collects a higher percentage of infeasible interventions compared to \mtgpcausal and \Gmtgp. In this setting, accurate estimation of the functions in the \scm translates to better uncertainty estimation around the effects given by the kernel function of \Gmtgp. Therefore \Gmtgp successfully trades off improvement and feasibility showing a similar convergence to \mtgpcausal but the highest percentage of feasible interventions collected.

With $\Nobs=100$ (\figref{fig:syntwo_N500_N100}, bottom row) and $\Nobs=10$ (\figref{fig:syntwo_N10} of Appendix \ref{sec:appsynthetic2}), the estimation of the target and constraint effects and the functions in the \scm from $\datao$ deteriorates leading \mtgpcausal, which learns the correlations directly from $\datai$, to outperform all other methods.

As in \synone, intervening on all variables in $\I$ simultaneously blocks the propagation of causal effects in the graph thus leading \Cbo to achieve a sub-optimal solution compared to \ccbo and a lower number of feasible interventions compared to \Gmtgp. 

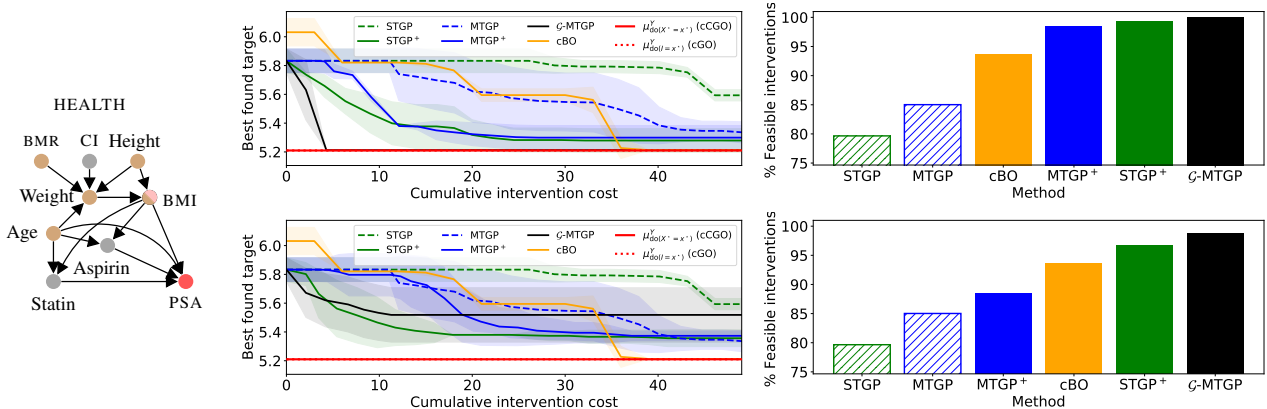
\begin{figure*}
    \centering
\begin{minipage}{.19\textwidth}
    \centering
    \scalebox{0.8}{\begin{tikzpicture}[dgraph]
    \node[] (a) [label= north:\health] at (0, 2.6) {};
    \node[dot] (cal) [fill=gray!70,label=north:\acro{ci}] at (0, 2) {};
    \node[dot] (bmr) [fill=brown!70,label=north:\small{\acro{bmr}}] at (-0.8, 2) {};
    \node[dot] (age) [fill=brown!70,label=left:\small{Age}] at (-0.6,0.8) {};
    \node[dot] (weight) [fill=brown!70,label=left:\small{Weight}] at (0, 1.4) {};
    \node[dot] (height) [fill=brown!70,label={[xshift=-0.05cm, yshift=-0.08cm]\small{Height}}] at (0.8, 2) {};
    \node[dot, double color fill={brown!70}{pink},shading angle=135] (BMI)[label={[xshift=0.5cm, yshift=-0.4cm]:$\bmi$}] at (1,1.4) {};
    \node[dot] (Aspirin) [fill=gray!70,label={[xshift=-0.1cm, yshift=-0.8cm]\small{Aspirin}}] at (0.3,0.6) {};
    \node[dot] (Statin) [fill=gray!70,label=south:\small{Statin}] at (-0.6, 0) {};
    \node[dot] (PSA) [fill={rgb:red,2;white,1},label=south:\psa] at (1.6,0) {};
    \node[] (b) at (0, -0.7) {};
    \draw[line width=0.6pt, \arr](age)--(weight);
    \draw[line width=0.6pt, \arr](bmr)--(weight);
    \draw[line width=0.6pt, \arr](cal)--(weight);
    \draw[line width=0.6pt, \arr](height)--(weight);
    \draw[line width=0.6pt, \arr](height)--(BMI);
    \draw[line width=0.6pt, \arr](weight)--(BMI);
    \draw[line width=0.6pt, \arr](BMI)--(PSA);
    \draw[line width=0.6pt, \arr](BMI)--(Aspirin);
    \draw[line width=0.6pt, \arr](age)--(Aspirin);
    \draw[line width=0.6pt, \arr](age)--(Statin);
    \draw[line width=0.6pt, \arr](Aspirin)--(PSA);
    \draw[line width=0.6pt, \arr](Statin)--(PSA);
    \draw[line width=0.6pt, \arr](BMI)to [bend right=+20](Statin);
    \draw[line width=0.6pt, \arr](age)to [bend left=+40](PSA);
    \end{tikzpicture}}
\end{minipage}
\begin{minipage}{0.4\textwidth}
    \centering
    \includegraphics[width=1.\textwidth]{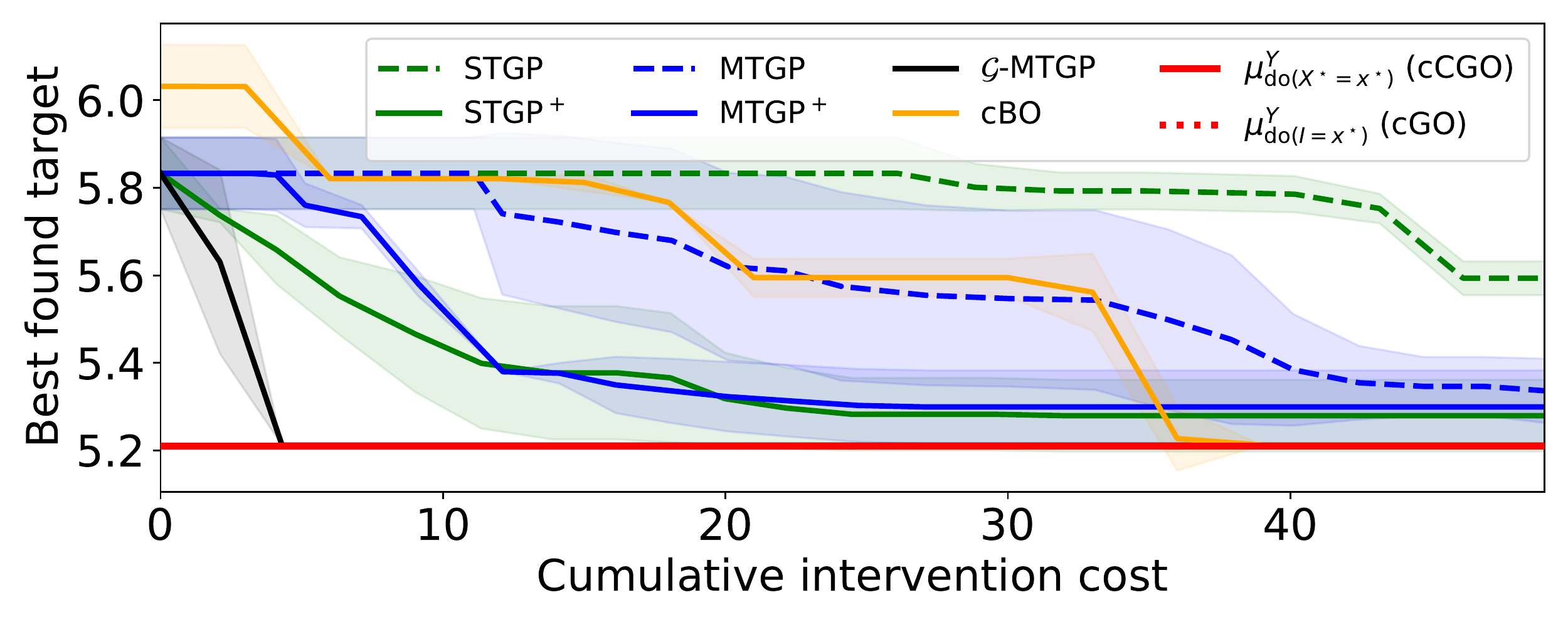}
    \includegraphics[width=1.\textwidth]{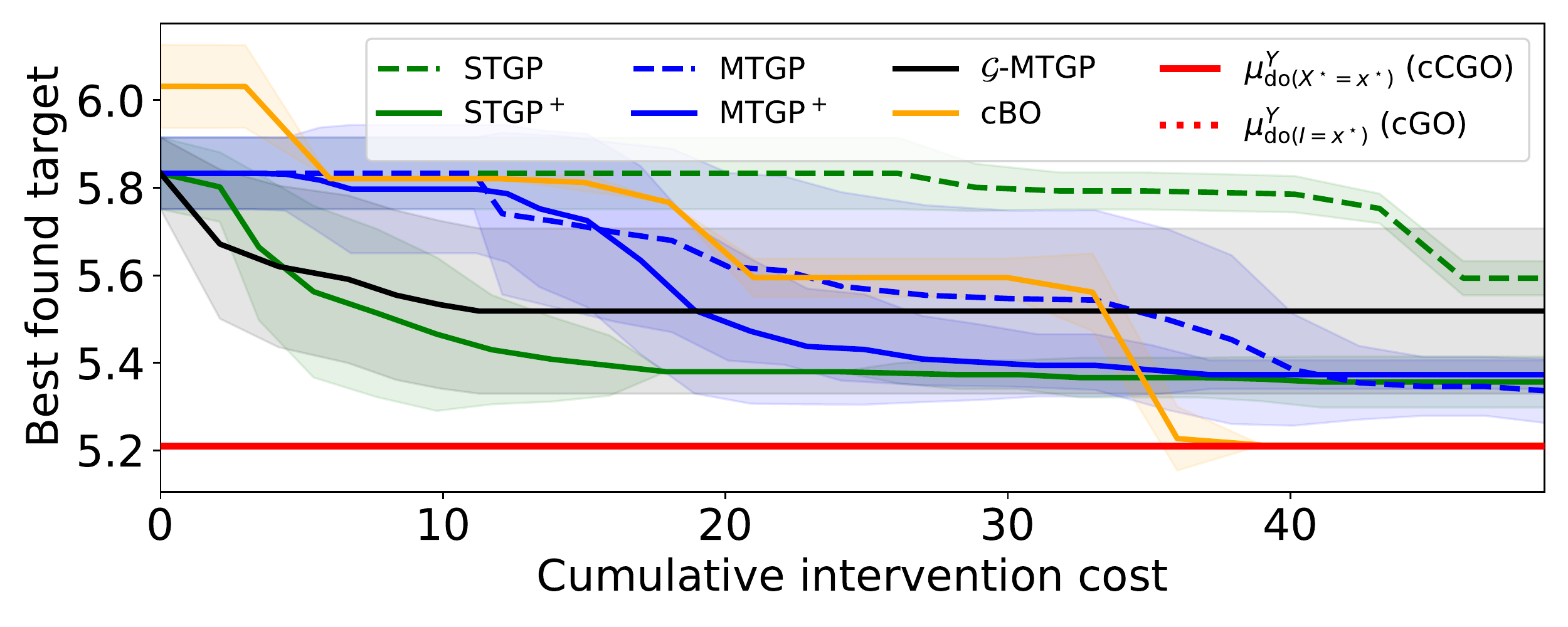}
\end{minipage}
\begin{minipage}{0.4\textwidth}
    \centering
    \includegraphics[width=1.\textwidth]{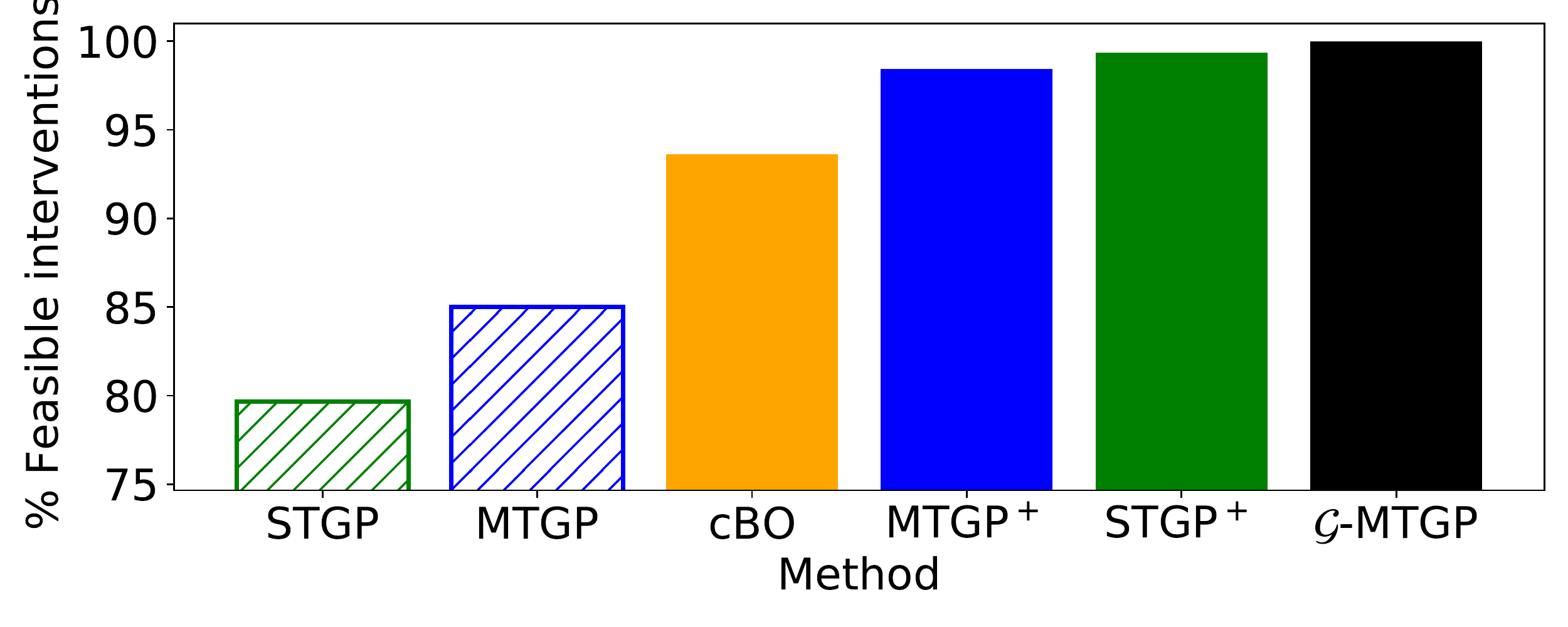}
    \includegraphics[width=1.\textwidth]{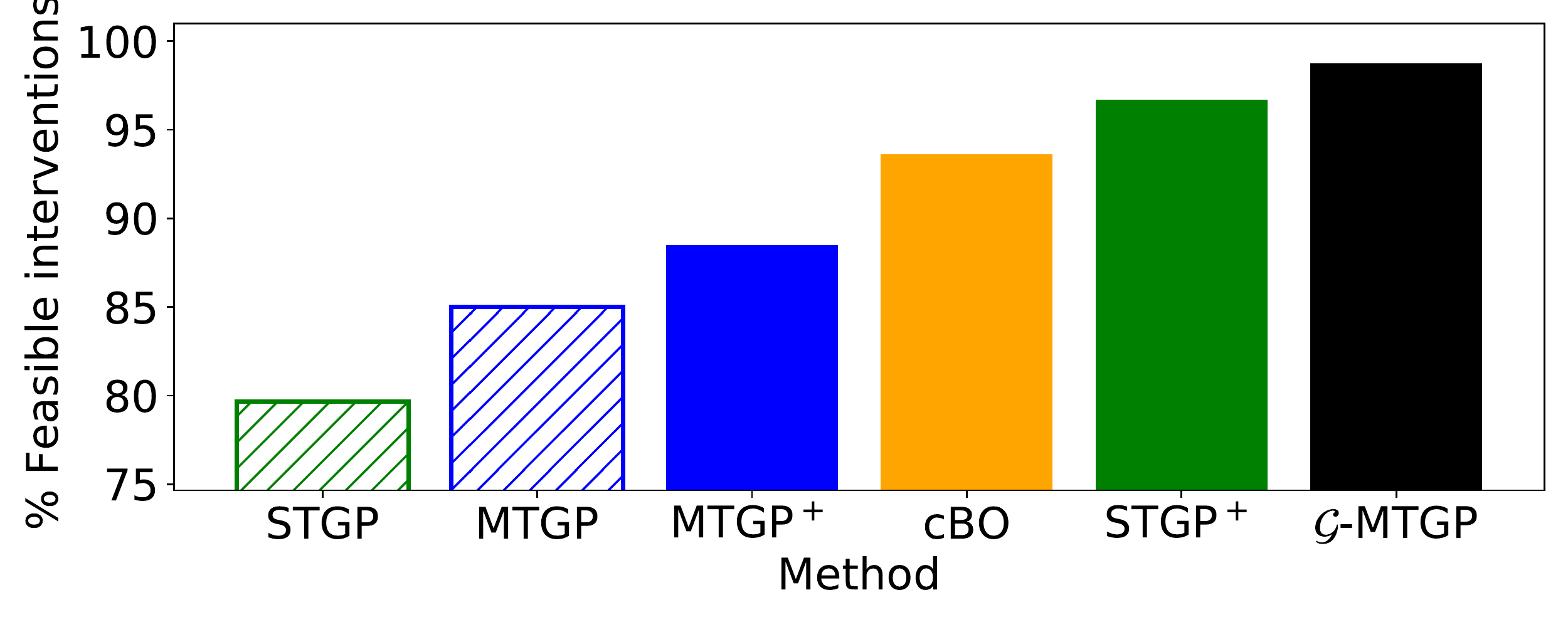}
\end{minipage}
\caption{\health~with $\Nobs=100$ (top row) and $\Nobs=10$ (bottom row). \textit{Left}: Causal graphs. \textit{Center}: Convergence to the \ccgo (solid red line) and \Cgo (dotted red line coinciding with the solid red line in this experiment) optima.
Lines give average results across different initialization of $\datai$. Shaded areas represents $\pm$ standard deviation. \textit{Right}: Average percentage of feasible interventions collected over trials.}
\label{fig:health_N100_N10}
\end{figure*}
\subsection{Real-world Causal Graphs}
\textbf{\health.} For the \health graph, we use the \scm in \citet{ferro2015use} (given in Appendix \ref{sec:appreal1}). We consider the constraint that $\bmi$ must be lower than 25, which is considered the maximum healthy level. When both $\Nobs=100$ and $\Nobs=10$, \bmi is not null-feasible thus all sets in $\allmisnull{\C \cup Y}= \{\{\text{\acro{ci}}\}, \{\text{Aspirin}, \text{\acro{ci}}\}, \{\text{Statin}, \text{\acro{ci}}\}, \{\text{Aspirin}, \text{Statin}, \text{\acro{ci}}\}\}$ include \acro{ci}. In this setting, $\I$ is the optimal intervention set, and therefore the \ccgo and \Cgo optima coincide.

When $\Nobs=100$, \Cbo is significantly slower than \Gmtgp and, by disregarding causal information, collects a lower percentage of feasible interventions over trials (\figref{fig:health_N100_N10}, top row). Even though the effects are simple linear functions, the stochasticity in $\datao$ determined by $p(\U)$ is high (\figref{fig:scatterhealth} of Appendix \ref{sec:appreal1}) compared to the previous examples and obscures their estimation. This leads to a less accurate prior mean function for \Gmtgp, \stgpcausal and \mtgpcausal which translates to a lower convergence speed for the latter two models. Despite the less accurate prior mean functions, \Gmtgp achieves the fastest convergence and the highest percentage of feasible interventions (100\%) by capturing the correlation among target and constraint effects induced by the \scm thus properly quantifying uncertainty around them. Collecting 100\% of feasible interventions is particularly important in this setting as infeasible interventions might negatively affect patients' health status. 
While \Gmtgp performs well in settings with $\Nobs=100$, it does not reach convergence when $\Nobs=10$ (\figref{fig:health_N100_N10}, bottom row). Indeed, the prior mean and kernel functions estimation from $\datao$ deteriorates when the size of the observational dataset is very small leading to an incorrect uncertainty quantification around the effects and preventing the exploration of the interventional space. This is also observed for \stgpcausal and \mtgpcausal as their prior mean function is affected by the incorrect estimation of the \scm functions and thus of the target and constraint effects. This leads \Cbo to outperform all other methods in this setting.

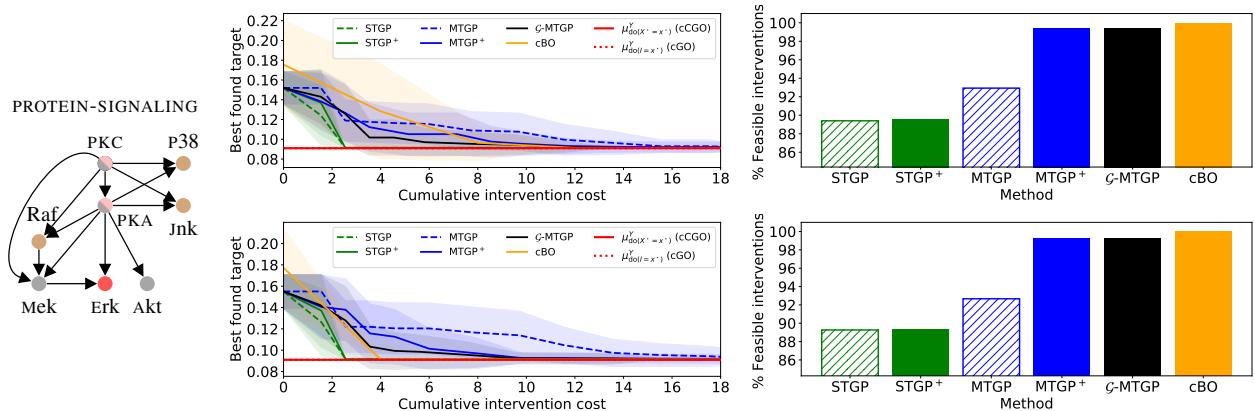
\begin{figure*}
    \centering
\begin{minipage}{.19\textwidth}
    \centering
    \scalebox{0.8}{\begin{tikzpicture}[dgraph]
    \node[] (b) [label= north:\protein] at (0, 2.6) {};
    \node[dot,double color fill={gray!70}{pink},shading angle=135] (PKC) [label=north:\acro{pkc}] at (0,2) {};
    \node[dot] (Raf)
    [fill=brown!70,label={[xshift=0.05cm, yshift=0.09cm]:\acronospace{R}af}] at (-1.1,0.7) {};
    \node[dot] (Jnk) [fill=brown!70,label=south:\acronospace{j}nk] at (1.3, 1.3) {};
    \node[dot] (P38) [fill=brown!70,label=north:\acronospace{p}38] at (1.3, 2.) {};
    \node[dot,double color fill={gray!70}{pink},shading angle=135] (PKA) [fill=pink,label={[xshift=0.5cm, yshift=-0.6cm]:\acro{pka}}] at (0,1.3) {};
    \node[dot] (Mek) [fill=gray!70,label=south:\acronospace{m}ek] at (-1.1,0) {};
    \node[dot] (Erk) [fill={rgb:red,2;white,1},label=south:\acronospace{e}rk] at (0,0) {};
    \node[dot] (Akt) [fill=gray!70, label=south:\acronospace{a}kt] at (0.7, 0.) {};
    \node[] (b) at (0, -0.7) {};
    \draw[line width=0.6pt, \arr](PKC)--(PKA);
    \draw[line width=0.6pt, \arr](PKC)--(Raf);
    \draw[line width=0.6pt, \arr](PKC)--(Jnk);
    \draw[line width=0.6pt, \arr](PKC)--(P38);
    \draw[line width=0.6pt, \arr](PKC)to [bend right=+100] (Mek);
    \draw[line width=0.6pt, \arr](PKA)--(Raf);
    \draw[line width=0.6pt, \arr](PKA)--(Mek);
    \draw[line width=0.6pt, \arr](PKA)--(Erk);
    \draw[line width=0.6pt, \arr](PKA)--(Akt);
    \draw[line width=0.6pt, \arr](PKA)--(Jnk);
    \draw[line width=0.6pt, \arr](PKA)--(P38);
    \draw[line width=0.6pt, \arr](Raf)--(Mek);
    \draw[line width=0.6pt, \arr](Mek)--(Erk);
    \end{tikzpicture}}
\end{minipage}
\begin{minipage}{0.4\textwidth}
    \centering
    \includegraphics[width=1.\textwidth]{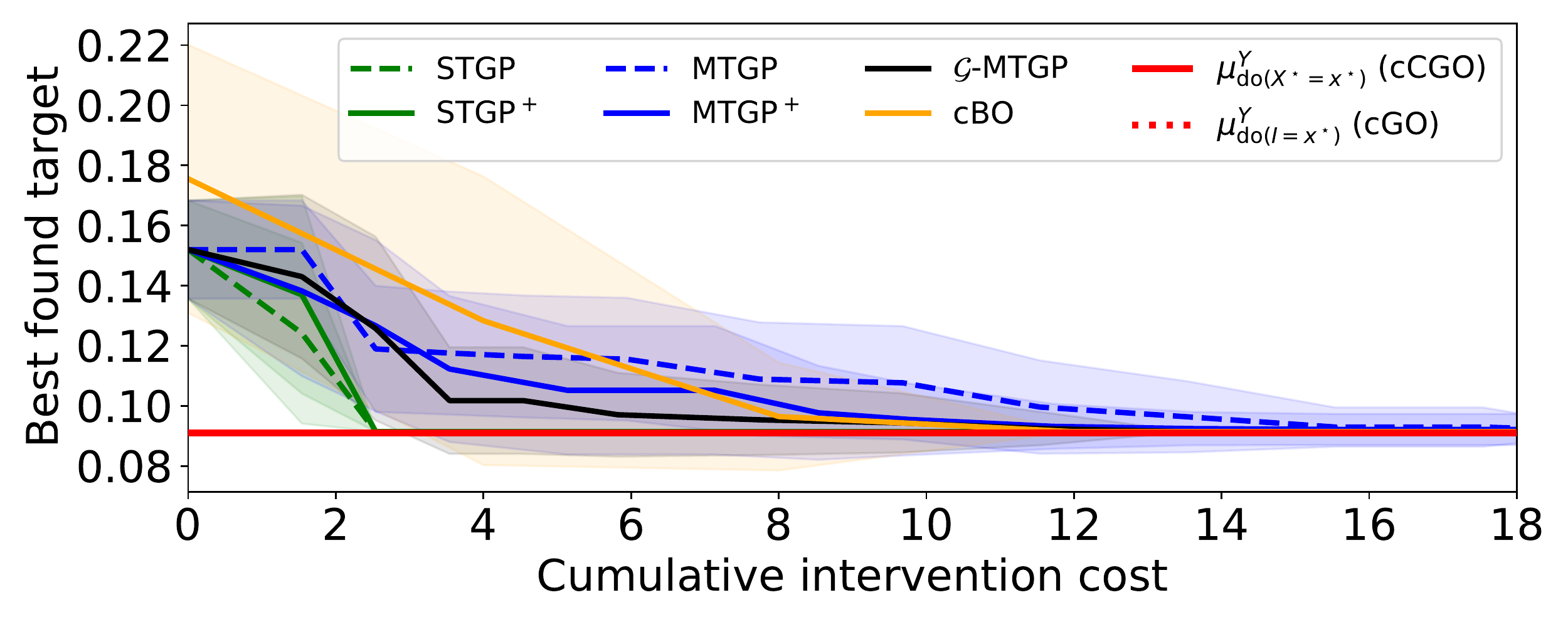}
    \includegraphics[width=1.\textwidth]{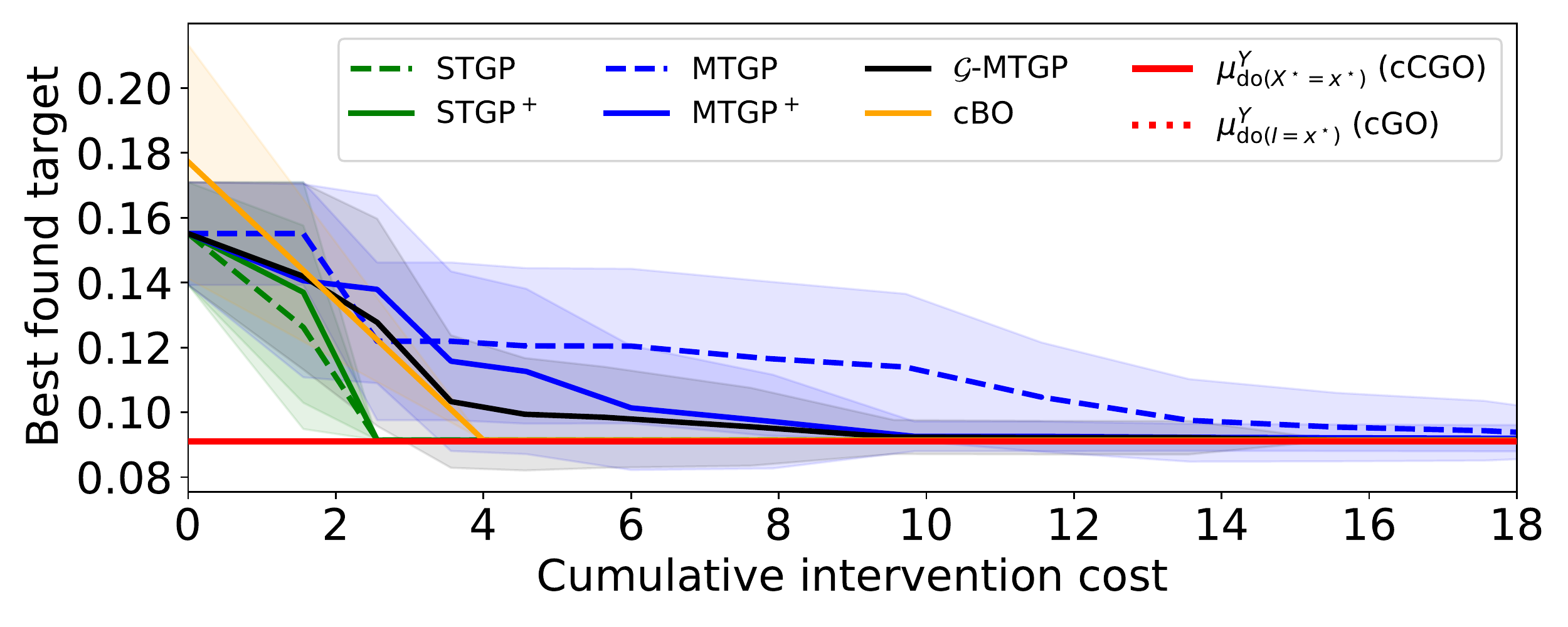}
\end{minipage}
\begin{minipage}{0.4\textwidth}
    \centering
    \includegraphics[width=1.\textwidth]{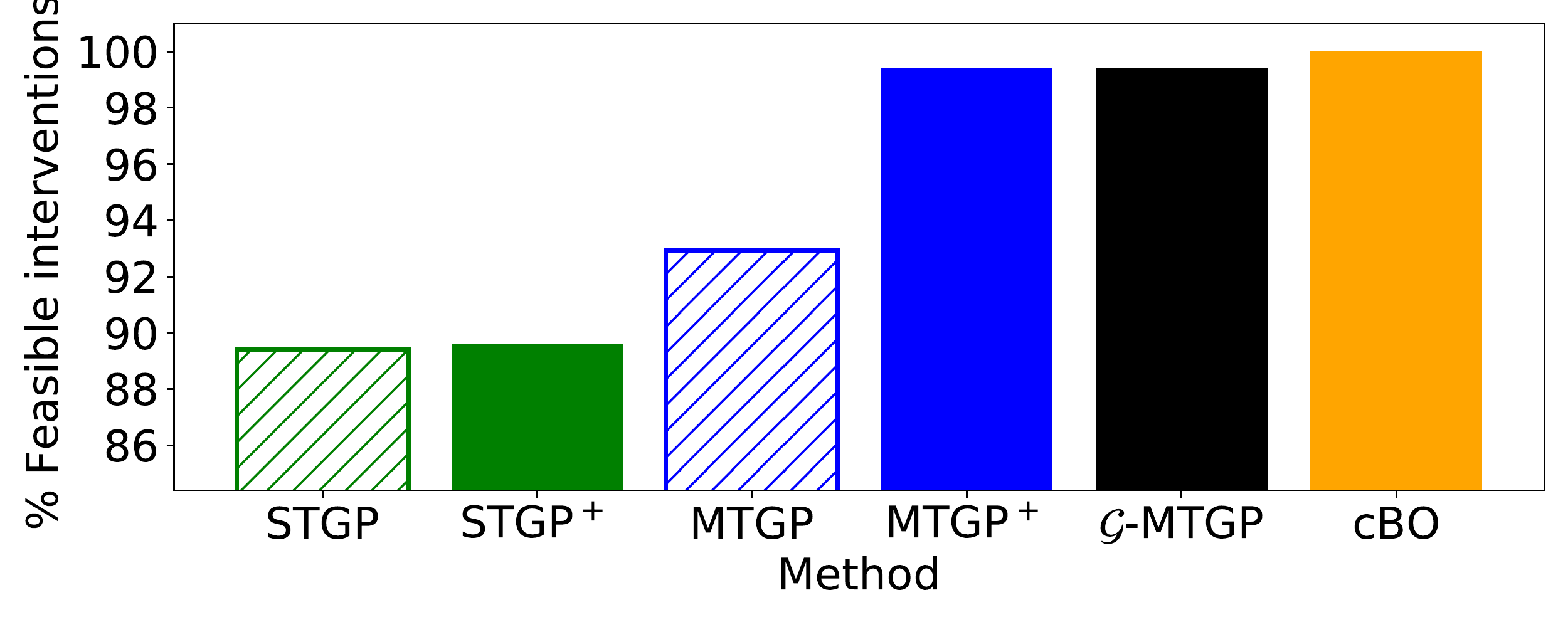}
    \includegraphics[width=1.\textwidth]{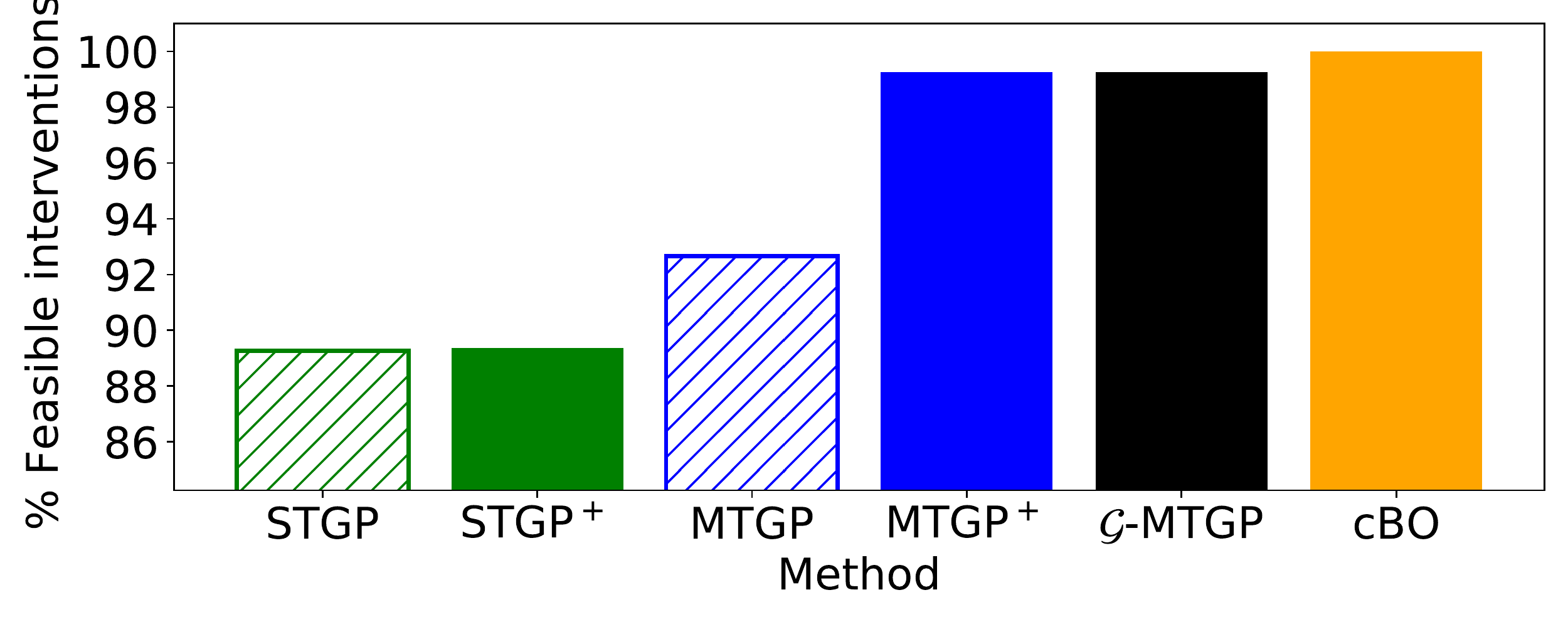}
\end{minipage}
\caption{\protein~with $\Nobs=100$ and $\Nobs=10$ (bottom row). \textit{Left}: Causal graphs. \textit{Center}: Convergence to the \ccgo (solid red line) and \Cgo (dotted red line coinciding with the solid red line in this experiment) optima.
Lines give average results across different initialization of $\datai$. Shaded areas represents $\pm$ standard deviation. \textit{Right}: Average percentage of feasible interventions collected over trials.}
\label{fig:protein_N100_N10}
\end{figure*}

\textbf{\protein.}
For the \protein graph, we use the observational  dataset from \citet{sachs2005causal}, to construct an \scm (see Appendix \ref{sec:appreal2}). Given $\Nobs=100, 10$ observational data samples from the \scm, all constrained variables are null-feasible, giving $\allmisnull{\C \cup Y} = \{\{\acro{pkc}\}, \{\acro{pka}\}, \{\acro{m}\text{ek}\}, \{\acro{pkc}, \acro{pka}\}, \{\acro{pkc}, \acro{m}\text{ek}\}, \{\acro{pka}, \\ \acro{m}\text{ek}\}\}$. All methods converge to the same optimum (\figref{fig:protein_N100_N10}). Indeed, the target effect achieved by intervening on $\I$ is equal to the one achieved by intervening on the optimal intervention set $\{\text{\pka}, \text{\mek}\}$ (this is the result of $\text{\erk}$ being independent on $\text{\pkc}$ and $\text{\akt}$ given $\text{\mek}$ and \pka). In addition, in this example $\C_{\I} = \emptyset$ thus by intervening on $\I$ with interventional domains that are in accordance with the threshold values \Cbo achieves a 100\% average feasibility. 

Overall \Cbo performs comparably to \Gmtgp and \mtgpcausal (\figref{fig:protein_N100_N10}, top row) when $\Nobs=100$, while \cbo is faster but selects more infeasible interventions (\figref{fig:all_protein} in Appendix \ref{sec:appreal2}). Despite achieving slower convergence than single task models, \mtgpcausal and \Gmtgp select a higher percentage of feasible interventions (99.13\%) with \Gmtgp converging slightly faster then \mtgpcausal. The feasibility aspect is again particularly important as every non feasible intervention results in the inhibition of $\pka$, which can impede healthy functions of the cell. Hence, a method that yields greater than 99\% feasible interventions at the cost of ($\sim 10$) additional interventions (\mtgpcausal and \Gmtgp) is deemed preferable compared to methods that explore the intervention space more aggressively but yield a lower number of feasible interventions (\cbo, \stgp and \stgpcausal). 

As in the other experiments, the convergence performance of \Gmtgp and \mtgpcausal slightly deteriorates when $\Nobs=10$ (\figref{fig:protein_N100_N10}, bottom row) due to an incorrect uncertainty quantification around the effects which prevents the exploration of the interventional space. As in the results for $\Nobs = 100$, notice how \Cbo achieves a 100\% average feasibility as all constrained variables are intervened ($\C \subset \I$, $\C_{\I} = \emptyset$) with interventional ranges that are in accordance with the threshold values. The algorithm quickly identify the \Cgo optimum, which is in this case equal to the \ccgo optimum.

\section{Discussion}
\label{sec:discussion}
In this paper, we introduced the \ccbo approach for identifying interventions optimizing a target effect under constraints.
We proposed different \gptext surrogate models for the target and constraint effects that leverage observational data $\datao$ and the structure of the \scm. Our results show that incorporating $\datao$ in the \gptext prior construction leads to faster identification of optimal interventions and higher percentage of feasible interventions selected. They also show that further performance improvement can be obtained using multi-task \gptext models which capture the correlation among target and constraint effects. Accounting for the \scm structure in the covariance matrix, as done by \Gmtgp, is especially beneficial with more complex correlation structures or with a high number of effects. We found \Gmtgp to successfully trade off improvement and feasibility, achieving a fast convergence while collecting the highest percentage of feasible interventions. However, as \Gmtgp exploits the fitted \scm functions in the computation of the prior parameters, when these functions cannot be learned accurately \mtgpcausal is preferable as it can learn the correlation directly from $\datai$.

Our search space reduction procedure can be thought as placing an intervention cost for each $\X \in \powerI$ that is equal to its cardinality. This procedure could be modified using different cost structures or augmented with budget constraints. For example, one could exclude from $\powerI$ the intervention sets non satisfying a budget and then proceed by excluding sets according to the proposed procedure. 

\ccbo requires knowledge of the true causal graph underlying the system of interest, an assumption that might not be satisfied in practice. Using \ccbo with an incorrect graph $\graph^\prime$ could lead to inaccurate \gptext prior parameters and invalid search space reduction. Indeed, there is no guarantee that $n\mathbb{M}_{\C \cup Y,\graph^\prime}$ would contain the optimal intervention set when the independence and the ancestor-type relationships encoded in $\mathcal{G^\prime}$ differ from those in $\graph$. Extending the methodology to deal with settings characterized by misspecified or unknown $\graph$, similarly \eg to \citet{branchini2023causal}, represents an important future direction.

From a computational perspective, \ccbo might suffer from poor scalability when $|\C|$ or the number of collected interventional data samples are high. In the latter case, sparse \gptext methods \eg inducing inputs \cite{titsias2009variational}, could be used to significantly reduce the computational complexity of both single-task and multi-task \gptext{s}. Alternatively, one could consider sharing information across the surrogate models of 
different sets in $n\mathbb{M}_{\mathbf{C} \cup Y, \mathcal{G}}$ in order to reduce the total number of interventions. Evaluating \ccbo on real-world settings where $|\C|$ is high would require simulators characterized by a high number of variables. In terms of convergence properties, while the constrained \ei does not currently provide theoretical guarantees, one could extend \ccbo to use a constrained acquisition function achieving asymptotic convergence, \eg \gptext-\acro{ucb},  \cite{srinivas2009gaussian, lu2022no} or a non-myopic acquisition function to avoid exploration issues \cite{lam2017lookahead}. 

Another future direction for safe-critical applications would be to extend the current framework to deal with safety constraints in order to guarantee that the number of feasible interventions collected never falls below a critical value. Finally, while \ccbo focuses on hard interventions in which variables are set to specific values, an interesting extension would be to consider more general soft-interventions. 

\section*{Acknowledgements}
The authors would like to thank Alexis Bellot for his valuable feedback.

\bibliography{bib}

\begin{thebibliography}{48}
\providecommand{\natexlab}[1]{#1}
\providecommand{\url}[1]{\texttt{#1}}
\expandafter\ifx\csname urlstyle\endcsname\relax
  \providecommand{\doi}[1]{doi: #1}\else
  \providecommand{\doi}{doi: \begingroup \urlstyle{rm}\Url}\fi

\bibitem[Aglietti et~al.(2020)Aglietti, Lu, Paleyes, and
  Gonz{\'a}lez]{aglietti2020causal}
Aglietti, V., Lu, X., Paleyes, A., and Gonz{\'a}lez, J.
\newblock Causal {B}ayesian optimization.
\newblock In \emph{International Conference on Artificial Intelligence and
  Statistics}, pp.\  3155--3164, 2020.

\bibitem[Alvarez et~al.(2011)Alvarez, Rosasco, and
  Lawrence]{alvarez2011kernels}
Alvarez, M.~A., Rosasco, L., and Lawrence, N.~D.
\newblock Kernels for vector-valued functions: A review.
\newblock \emph{arXiv preprint arXiv:1106.6251}, 2011.

\bibitem[Ariafar et~al.(2019)Ariafar, Coll-Font, and Brooks]{ariafar2019admmbo}
Ariafar, S., Coll-Font, J., and Brooks, Dana Hand~Dy, J.~G.
\newblock A{DMMBO}: {B}ayesian optimization with unknown constraints using
  {ADMM}.
\newblock \emph{Journal of Machine Learning Research}, 20\penalty0
  (123):\penalty0 1--26, 2019.

\bibitem[Atan et~al.(2018)Atan, Jordon, and Van~der Schaar]{atan2018deep}
Atan, O., Jordon, J., and Van~der Schaar, M.
\newblock Deep-treat: Learning optimal personalized treatments from
  observational data using neural networks.
\newblock In \emph{AAAI Conference on Artificial Intelligence}, pp.\
  2071--2078, 2018.

\bibitem[Bergmann \& Graichen(2020)Bergmann and Graichen]{bergmann2020safe}
Bergmann, D. and Graichen, K.
\newblock Safe {B}ayesian optimization under unknown constraints.
\newblock In \emph{IEEE Conference on Decision and Control}, pp.\  3592--3597,
  2020.

\bibitem[Berkenkamp et~al.(2016)Berkenkamp, Schoellig, and
  Krause]{berkenkamp2016safe}
Berkenkamp, F., Schoellig, A.~P., and Krause, A.
\newblock Safe controller optimization for quadrotors with {G}aussian
  processes.
\newblock In \emph{IEEE International Conference on Robotics and Automation},
  pp.\  491--496, 2016.

\bibitem[Berkenkamp et~al.(2021)Berkenkamp, Krause, and
  Schoellig]{berkenkamp2021bayesian}
Berkenkamp, F., Krause, A., and Schoellig, A.~P.
\newblock Bayesian optimization with safety constraints: {S}afe and automatic
  parameter tuning in robotics.
\newblock \emph{Machine Learning}, pp.\  1--35, 2021.

\bibitem[Branchini et~al.(2023)Branchini, Aglietti, Dhir, and
  Damoulas]{branchini2023causal}
Branchini, N., Aglietti, V., Dhir, N., and Damoulas, T.
\newblock Causal entropy optimization.
\newblock In \emph{International Conference on Artificial Intelligence and
  Statistics}, pp.\  8586--8605, 2023.

\bibitem[Bull(2011)]{bull2011convergence}
Bull, A.~D.
\newblock Convergence rates of efficient global optimization algorithms.
\newblock \emph{Journal of Machine Learning Research}, 12\penalty0 (10), 2011.

\bibitem[Curi et~al.(2020)Curi, Berkenkamp, and Krause]{curi2020efficient}
Curi, S., Berkenkamp, F., and Krause, A.
\newblock Efficient model-based reinforcement learning through optimistic
  policy search and planning.
\newblock \emph{Advances in Neural Information Processing Systems},
  33:\penalty0 14156--14170, 2020.

\bibitem[Dai et~al.(2020)Dai, Song, and Yue]{dai2020multi}
Dai, S., Song, J., and Yue, Y.
\newblock Multi-task {B}ayesian optimization via {G}aussian process upper
  confidence bound.
\newblock In \emph{ICML 2020 Workshop on Real World Experiment Design and
  Active Learning}, 2020.

\bibitem[Eriksson \& Poloczek(2021)Eriksson and Poloczek]{eriksson2021scalable}
Eriksson, D. and Poloczek, M.
\newblock Scalable constrained {B}ayesian optimization.
\newblock In \emph{International Conference on Artificial Intelligence and
  Statistics}, pp.\  730--738, 2021.

\bibitem[Feliot et~al.(2017)Feliot, Bect, and Vazquez]{feliot2017bayesian}
Feliot, P., Bect, J., and Vazquez, E.
\newblock A {B}ayesian approach to constrained single-and multi-objective
  optimization.
\newblock \emph{Journal of Global Optimization}, 67\penalty0 (1-2):\penalty0
  97--133, 2017.

\bibitem[Ferro et~al.(2015)Ferro, Pina, Severo, Dias, Botelho, and
  Lunet]{ferro2015use}
Ferro, A., Pina, F., Severo, M., Dias, P., Botelho, F., and Lunet, N.
\newblock Use of statins and serum levels of prostate specific antigen.
\newblock \emph{Acta Urol{\'o}gica Portuguesa}, 32\penalty0 (2):\penalty0
  71--77, 2015.

\bibitem[Fr{\'e}min \& Meloche(2010)Fr{\'e}min and Meloche]{fremin2010basic}
Fr{\'e}min, C. and Meloche, S.
\newblock From basic research to clinical development of {MEK1/2} inhibitors
  for cancer therapy.
\newblock \emph{Journal of Hematology \& Oncology}, 3\penalty0 (1):\penalty0
  1--11, 2010.

\bibitem[Gardner et~al.(2014)Gardner, Kusner, Xu, Weinberger, and
  Cunningham]{gardner2014bayesian}
Gardner, J.~R., Kusner, M.~J., Xu, Z.~E., Weinberger, K.~Q., and Cunningham,
  J.~P.
\newblock Bayesian optimization with inequality constraints.
\newblock In \emph{International Conference on Machine Learning}, pp.\
  937--945, 2014.

\bibitem[Gelbart et~al.(2014)Gelbart, Snoek, and Adams]{gelbart2014bayesian}
Gelbart, M.~A., Snoek, J., and Adams, R.~P.
\newblock Bayesian optimization with unknown constraints.
\newblock \emph{arXiv preprint arXiv:1403.5607}, 2014.

\bibitem[Griffiths \& Hern{\'a}ndez-Lobato(2017)Griffiths and
  Hern{\'a}ndez-Lobato]{griffiths2017constrained}
Griffiths, R.-R. and Hern{\'a}ndez-Lobato, J.~M.
\newblock Constrained {B}ayesian optimization for automatic chemical design.
\newblock \emph{arXiv preprint arXiv:1709.05501}, 2017.

\bibitem[H{\aa}kansson et~al.(2020)H{\aa}kansson, Lindblom, Gottesman, and
  Johansson]{haakansson2020learning}
H{\aa}kansson, S., Lindblom, V., Gottesman, O., and Johansson, F.~D.
\newblock Learning to search efficiently for causally near-optimal treatments.
\newblock In \emph{Advances in Neural Information Processing Systems},
  volume~33, pp.\  1333--1344, 2020.

\bibitem[Hakhamaneshi et~al.(2021)Hakhamaneshi, Abbeel, Stojanovic, and
  Grover]{hakhamaneshi2021jumbo}
Hakhamaneshi, K., Abbeel, P., Stojanovic, V., and Grover, A.
\newblock J{UMBO}: {S}calable multi-task {B}ayesian optimization using offline
  data.
\newblock \emph{arXiv preprint arXiv:2106.00942}, 2021.

\bibitem[Hern{\'a}ndez-Lobato et~al.(2014)Hern{\'a}ndez-Lobato, Hoffman, and
  Ghahramani]{hernandez2014predictive}
Hern{\'a}ndez-Lobato, J.~M., Hoffman, M.~W., and Ghahramani, Z.
\newblock Predictive entropy search for efficient global optimization of
  black-box functions.
\newblock \emph{arXiv preprint arXiv:1406.2541}, 2014.

\bibitem[Keane et~al.(2008)Keane, Forrester, and
  Sobester]{keane2008engineering}
Keane, A., Forrester, A., and Sobester, A.
\newblock \emph{Engineering design via surrogate modelling: a practical guide}.
\newblock American Institute of Aeronautics and Astronautics, Inc., 2008.

\bibitem[Kirschner et~al.(2019)Kirschner, Mutny, Hiller, Ischebeck, and
  Krause]{kirschner2019adaptive}
Kirschner, J., Mutny, M., Hiller, N., Ischebeck, R., and Krause, A.
\newblock Adaptive and safe {B}ayesian optimization in high dimensions via
  one-dimensional subspaces.
\newblock In \emph{International Conference on Machine Learning}, pp.\
  3429--3438, 2019.

\bibitem[Lam \& Willcox(2017)Lam and Willcox]{lam2017lookahead}
Lam, R. and Willcox, K.
\newblock Lookahead bayesian optimization with inequality constraints.
\newblock In \emph{Advances in Neural Information Processing Systems}, 2017.

\bibitem[Lattimore et~al.(2016)Lattimore, Lattimore, and
  Reid]{lattimore2016causal}
Lattimore, F., Lattimore, T., and Reid, M.~D.
\newblock Causal bandits: Learning good interventions via causal inference.
\newblock In \emph{Advances in Neural Information Processing Systems}, 2016.

\bibitem[Lee \& Bareinboim(2018)Lee and Bareinboim]{lee2018structural}
Lee, S. and Bareinboim, E.
\newblock Structural causal bandits: where to intervene?
\newblock In \emph{Advances in Neural Information Processing Systems}, pp.\
  2568--2578, 2018.

\bibitem[Letham et~al.(2019)Letham, Karrer, Ottoni, and
  Bakshy]{letham2019constrained}
Letham, B., Karrer, B., Ottoni, G., and Bakshy, E.
\newblock Constrained {B}ayesian optimization with noisy experiments.
\newblock \emph{Bayesian Analysis}, 14\penalty0 (2):\penalty0 495--519, 2019.

\bibitem[Lu \& Paulson(2022)Lu and Paulson]{lu2022no}
Lu, C. and Paulson, J.~A.
\newblock No-regret bayesian optimization with unknown equality and inequality
  constraints using exact penalty functions.
\newblock \emph{IFAC-PapersOnLine}, 55\penalty0 (7):\penalty0 895--902, 2022.

\bibitem[Mathern et~al.(2021)Mathern, Steinholtz, Sj{\"o}berg, {\"O}nnheim, Ek,
  Rempling, Gustavsson, and Jirstrand]{mathern2021multi}
Mathern, A., Steinholtz, O.~S., Sj{\"o}berg, A., {\"O}nnheim, M., Ek, K.,
  Rempling, R., Gustavsson, E., and Jirstrand, M.
\newblock Multi-objective constrained {B}ayesian optimization for structural
  design.
\newblock \emph{Structural and Multidisciplinary Optimization}, 63\penalty0
  (2):\penalty0 689--701, 2021.

\bibitem[Pearl(2000)]{pearl2000causality}
Pearl, J.
\newblock \emph{Causality: Models, Reasoning and Inference}, volume~29.
\newblock Springer, 2000.

\bibitem[Pearl et~al.(2016)Pearl, Glymour, and Jewell]{pearl2016causal}
Pearl, J., Glymour, M., and Jewell, N.~P.
\newblock \emph{Causal Inference in Statistics: A Primer}.
\newblock John Wiley \& Sons, 2016.

\bibitem[Picheny et~al.(2016)Picheny, Gramacy, Wild, and
  Digabel]{picheny2016bayesian}
Picheny, V., Gramacy, R.~B., Wild, S., and Digabel, S.~L.
\newblock Bayesian optimization under mixed constraints with a slack-variable
  augmented {L}agrangian.
\newblock In \emph{Advances in Neural Information Processing Systems}, pp.\
  1443--1451, 2016.

\bibitem[Rasmussen \& Williams(2006)Rasmussen and
  Williams]{rasmussen2006gaussian}
Rasmussen, C.~E. and Williams, C. K.~I.
\newblock \emph{Gaussian Processes for Machine Learning}.
\newblock The MIT Press, 2006.

\bibitem[Sachs et~al.(2005)Sachs, Perez, Pe{\textquoteright}er, Lauffenburger,
  and Nolan]{sachs2005causal}
Sachs, K., Perez, O., Pe{\textquoteright}er, D., Lauffenburger, D.~A., and
  Nolan, G.~P.
\newblock Causal protein-signaling networks derived from multiparameter
  single-cell data.
\newblock \emph{Science}, 308\penalty0 (5721):\penalty0 523--529, 2005.

\bibitem[Schonlau et~al.(1998)Schonlau, Welch, and Jones]{schonlau1998global}
Schonlau, M., Welch, W.~J., and Jones, D.~R.
\newblock Global versus local search in constrained optimization of computer
  models.
\newblock \emph{Lecture Notes-Monograph Series}, pp.\  11--25, 1998.

\bibitem[Snoek et~al.(2012)Snoek, Larochelle, and Adams]{snoek2012practical}
Snoek, J., Larochelle, H., and Adams, R.~P.
\newblock Practical {B}ayesian optimization of machine learning algorithms.
\newblock In \emph{Advances in Neural Information Processing Systems},
  volume~25, 2012.

\bibitem[S{\'o}bester et~al.(2014)S{\'o}bester, Forrester, Toal, Tresidder, and
  Tucker]{sobester2014engineering}
S{\'o}bester, A., Forrester, A.~I., Toal, D.~J., Tresidder, E., and Tucker, S.
\newblock Engineering design applications of surrogate-assisted optimization
  techniques.
\newblock \emph{Optimization and Engineering}, 15\penalty0 (1):\penalty0
  243--265, 2014.

\bibitem[Srinivas et~al.(2009)Srinivas, Krause, Kakade, and
  Seeger]{srinivas2009gaussian}
Srinivas, N., Krause, A., Kakade, S.~M., and Seeger, M.
\newblock Gaussian process optimization in the bandit setting: No regret and
  experimental design.
\newblock \emph{arXiv preprint arXiv:0912.3995}, 2009.

\bibitem[Sui et~al.(2015)Sui, Gotovos, Burdick, and Krause]{sui2015safe}
Sui, Y., Gotovos, A., Burdick, J., and Krause, A.
\newblock Safe exploration for optimization with {G}aussian processes.
\newblock In \emph{International Conference on Machine Learning}, pp.\
  997--1005, 2015.

\bibitem[Sui et~al.(2018)Sui, Zhuang, Burdick, and Yue]{sui2018stagewise}
Sui, Y., Zhuang, V., Burdick, J., and Yue, Y.
\newblock Stagewise safe {B}ayesian optimization with {G}aussian processes.
\newblock In \emph{International Conference on Machine Learning}, pp.\
  4781--4789, 2018.

\bibitem[Sussex et~al.(2023)Sussex, Makarova, and Krause]{sussex2023model}
Sussex, S., Makarova, A., and Krause, A.
\newblock Model-based causal {B}ayesian optimization.
\newblock In \emph{International Conference on Learning Representations}, 2023.

\bibitem[Swersky et~al.(2013)Swersky, Snoek, and Adams]{swersky2013multi}
Swersky, K., Snoek, J., and Adams, R.~P.
\newblock Multi-task {B}ayesian optimization.
\newblock In \emph{Advances in Neural Information Processing Systems}, 2013.

\bibitem[Tigas et~al.(2022)Tigas, Annadani, Jesson, Sch{\"o}lkopf, Gal, and
  Bauer]{tigas2022interventions}
Tigas, P., Annadani, Y., Jesson, A., Sch{\"o}lkopf, B., Gal, Y., and Bauer, S.
\newblock Interventions, where and how? {E}xperimental design for causal models
  at scale.
\newblock In \emph{Advances in Neural Information Processing Systems}, 2022.

\bibitem[Titsias(2009)]{titsias2009variational}
Titsias, M.
\newblock Variational learning of inducing variables in sparse gaussian
  processes.
\newblock In \emph{Artificial Intelligence and Statistics}, pp.\  567--574,
  2009.

\bibitem[Tran et~al.(2019)Tran, Sun, Furlan, Pagalthivarthi, Visintainer, and
  Wang]{tran2019pbo}
Tran, A., Sun, J., Furlan, J.~M., Pagalthivarthi, K.~V., Visintainer, R.~J.,
  and Wang, Y.
\newblock p{BO-2GP-3B}: {A} batch parallel known/unknown constrained {B}ayesian
  optimization with feasibility classification and its applications in
  computational fluid dynamics.
\newblock \emph{Computer Methods in Applied Mechanics and Engineering},
  347:\penalty0 827--852, 2019.

\bibitem[Vazquez \& Bect(2010)Vazquez and Bect]{vazquez2010convergence}
Vazquez, E. and Bect, J.
\newblock Convergence properties of the expected improvement algorithm with
  fixed mean and covariance functions.
\newblock \emph{Journal of Statistical Planning and Inference}, 140\penalty0
  (11):\penalty0 3088--3095, 2010.

\bibitem[Zhang et~al.(2012)Zhang, Tsiatis, Laber, and
  Davidian]{zhang2012robust}
Zhang, B., Tsiatis, A.~A., Laber, E.~B., and Davidian, M.
\newblock A robust method for estimating optimal treatment regimes.
\newblock \emph{Biometrics}, 68\penalty0 (4):\penalty0 1010--1018, 2012.

\bibitem[Zhang(2020)]{zhang2020designing}
Zhang, J.
\newblock Designing optimal dynamic treatment regimes: A causal reinforcement
  learning approach.
\newblock In \emph{International Conference on Machine Learning}, pp.\
  11012--11022, 2020.

\end{thebibliography}
\bibliographystyle{icml2023}

\newpage
\appendix
\onecolumn
\twocolumn[{}]

\section{Reducing the Search Space}
\label{secapp:rss}
Recall the definition of \cmis relative to $(\C \cup Y,\graph)$.
\begin{nameddef*}{Definition \ref{def:mis_set}}[\cmis]
A set $\X \subseteq \I$ is said to be a \cmis relative to $(\C \cup Y,\graph)$ if there is no $\X^\prime \subset \X$ with $\C_{\X}=\C_{\X^\prime}$ such that $\mu_{\doi(\X=\x)}^W = \muXprime^W$, where $\x'$ indicates the subset of $\x$ corresponding to variables $\X^{\prime}$, $\forall \x\in D(\X)$, $\forall W \in \C_{\X}\cup Y$, and $\forall$ \scm  with causal graph $\graph$. 
\end{nameddef*}
\noindent Also recall that $\allmis{\C \cup Y}$ is the subset of $\powerI$ whose elements are \cmis{s} relative to $({\C \cup Y},\graph)$, that $\an(W,\calG)$ denotes the set of ancestors of $W$ in $\calG$, that $\an(\W,\graph):=\cup_{W \in \W} \an(W,\graph)$, and that $\graph_{\overline{\X}}$ is the graph with all incoming edges onto all elements of $\X$ removed.

\begin{namedprop*}{Proposition~\ref{prop:compute_mis}}[Characterization of \cmis]
\textit{$\X\subseteq\I$ is a \cmis relative to $({\C \cup Y},\graph)$  $\iff$ $\X \subseteq \an(\C \cup Y,\graph_{\overline{\X}}) \cup (\C \cap \X)$.}
\end{namedprop*}
\noindent \begin{proof}\\
($\Longrightarrow$) We first prove that if $\X \subseteq \an(\C \cup Y, \graph_{\overline{\X}}) \cup (\C \cap \X)$ then $\X$ is a \cmis relative to $(\C \cup Y,\graph)$.\\

$\X \subseteq \an(\C \cup Y,\graph_{\overline{\X}}) \cup (\C \cap \X)$ implies that, for any $\X' \subset \X$, the set $\Q = \X \backslash \X^\prime$ is also a subset of $\an(\C \cup Y, \graph_{\overline{\X}}) \cup (\C \cap \X)$. Therefore (i) $\Q \subset \an(\C \cup Y, \graph_{\overline{\X}})$, or (ii) $\Q \subset \C \cap \X$, or (iii) $\Q$ includes both variables in $\an(\C \cup Y, \graph_{\overline{\X}})$ and $\C \cap \X$. In case (ii), $\Q \subseteq \C$ and, as $\X = \Q \cup \X'$, we obtain $\C \cap \X = \C \cap (\Q \cup \X') = (\C \cap \Q) \cup (\C \cap \X') = \Q \cup (\C \cap \X') \supset \C \cap \X'$ implying $\C_{\X} \subset \C_{\X'}$ which violates the requirement $\C_{\X} = \C_{\X'}$ of Definition \ref{def:mis_set}. 
In cases (i) and (iii), $\Q$ includes at least one variable, say $Q$, 
with at least one directed path, say from $Q$ to $W \in \C \cup Y$, that is not passing through $\X^\prime$ (as the incoming edges into $\X^\prime$ are removed in $\graph_{\overline{\X}}$). 
Consider a \scm with $V=\sum_{i=1}^{|\pa(V)|} \pa(V)_i + U_V$, where $\pa(V)_i$ is the $i$-th parent, and $U_V \sim \mathcal{N}(0,1)$. In this \scm, $\mu_{\doi(\X^\prime=\x^\prime)}^W = a\x^\prime + b \mu^{Q}_{\doi(\X^\prime=\x^\prime)}$ where $a$ and $b$ are two positive constants, while $\mu_{\doi(Q=q,\X^\prime=\x^\prime)}^W =  a\x^\prime + b q$, so that taking $q=\mu_{\doi(\X^\prime=\x^\prime)}^{Q}+1$ gives $\mu_{\doi(Q=q,\X^\prime=\x^\prime)}^W >\mu_{\doi(\X^\prime=\x^\prime)}^W$. Therefore, for any $\X^\prime \subset \X$ we can construct a \scm such that $\mu_{\doi(\X=\x)}^W > \muXprime^W$ for at least one $W \in \C \cup Y$. As there is no $\X^\prime \subset \X$ such that $\mu_{\doi(\X=\x)}^W = \muXprime^W$ $\forall W \in \C \cup Y$ and $\forall$ \scm with causal graph $\graph$, $\X$ satisfies the requirements of Definition \ref{def:mis_set} for being a \cmis relative to $(\C \cup Y, \graph)$.\\

\noindent ($\Longleftarrow$)
We now prove that if $\X$ is a \cmis relative to $(
\C \cup Y, \graph)$ then $\X \subseteq \an(\C \cup Y,\graph_{\overline{\X}}) \cup (\C \cap \X)$. \\[5pt]
If $\X$ were not a subset of $\an(\C \cup Y,\graph_{\overline{\X}}) \cup (\C \cap \X)$, then we could define the non empty set $\Q = \X \backslash \big(\X \cap (\an(\C \cup Y,\graph_{\overline{\X}}) \cup (\C \cap \X))\big)$ and the set $\X^\prime = \X \backslash \Q \subset \X$ such that (1) $\C_{\X} = \C_{\X'}$ and (2) all effects on variables in $\C \cup Y$ for $\X'$ and $\X$ are equal, contradicting $\X$ being a \cmis relative to $(\C \cup Y,\graph)$.
Condition (1) would hold as $\X \cap (\an(\C \cup Y,\graph_{\overline{\X}}) \cup (\C \cap \X)) = (\X \cap \an(\C \cup Y,\graph_{\overline{\X}})) \cup (\X \cap \C)$ implies that we can express $\Q$ as $\Q=\X\backslash (\B \cup (\X \cap \C))$ for $\B=\X \cap \an(\C \cup Y,\graph_{\overline{\X}})$, showing that $\Q$ does not include variables in $\X$ that are in $\C$, and therefore that $\C_{\X} = \C_{\X'}$.
Condition (2) would follow from the fact that the non-overlapping sets $\Q$ and $\C \cup Y$ satisfy $(\C \cup Y) \indep_{\graph_{\overline{\X}}}\Q \,|\, \X^\prime$ as: (i) there could not be causal paths from any variable in $\Q$ to any variable in $(\C \cup Y)$ in $\graph_{\overline{\X}}$ (as $\Q$ does not contain the ancestors of $(\C \cup Y)$ by definition), (ii) non-causal frontdoor paths would be closed as the incoming edges into the colliders that might be included in $\X'$ would be removed in $\graph_{\overline{\X}}$, and (iii) all backdoor paths would be removed in $\graph_{\overline{\X}}$. Therefore, by the Rule 3 of do-calculus, we would have $\mu^W_{\doi(\Q=\q,\X^\prime=\x^\prime)} =\mu^W_{\doi(\X^\prime=\x^\prime)}$  $\forall W \in \C \cup Y$. 
\end{proof} \\

\begin{namedprop*}{Theorem~\ref{theorem:sufficiency_cmis}}[Sufficiency of $\allmis{\C \cup Y}$]
\textit{
$\allmis{\C \cup Y}$ contains a solution of the \ccgo problem (if a solution exists), $\forall$ \scm with causal graph $\graph$.}  
\end{namedprop*}
\noindent \begin{proof}
Consider a set $\X$ that satisfies \eqref{eq:ccgo}. By Definition \ref{def:mis_set}, if $\X\notin\allmis{\C \cup Y}$ there exists a set $\X^\prime\subset \X$ with $\C_{\X} = \C_{\X^\prime}$ such that $\mu_{\doi(\X=\x)}^Y = \muXprime^Y$ and $\mu_{\doi(\X=\x)}^C = \muXprime^C, \; \forall C \in \C_{\X^\prime}$  and therefore $\X^\prime$ also satisfies \eqref{eq:ccgo}. If $\X^\prime\notin\allmis{\C \cup Y}$, we can apply a similar reasoning, and proceed until we find a set for which there is no subset that gives the same effects, which therefore is in $\allmis{\C \cup Y}$.
\end{proof}

\begin{table*}
\centering
\caption{Summary of the prior mean $m_{\X}^V(\x)$ and kernel $S_{\X}^V(\x, \x')$ parameters associated to the different surrogate models. $\{g_{\X}^{V, (s)}\}_{s=1}^{S'}$ denote a set of $S'$ realisations of $g_{\X}^V$ obtained by sampling from $p(\U)$ and the posterior distributions of $f_V$ given $\datao$, $\hat{\sigma}^V_{\doi(\X=\x)} = \sqrt{\frac{1}{S'}\sum_{s=1}^{S'} \left(g_{\X}^{V, (s)}(\x) - m_{\X}^V(\x\right)^2}$, $\sigma^2_f$ and $l$ are the kernel hyper-parameters, and $S_{\X, q}(\x, \x')$ is the kernel function for $u_{\X, q}$ with associated scalar coefficient $a_{\X, q}^{V_k}$.}
\begin{tabular}{lcc}
\toprule
&  \multicolumn{2}{c}{Prior parameters for $g_{\X}^V$}\\
\cmidrule(lr){2-3} 
& $m^V_{\X}(\x)$ & $S^V_{\X}(\x, \x')$ \\
\cmidrule(lr){2-3}
\Cbo &  0 & $\sigma^2_f \exp(-\frac{||\x - \x'||^2}{2l^2})$ \\
\cbo &  0 & $\sigma^2_f \exp(-\frac{||\x - \x'||^2}{2l^2})$ \\
\stgp & 0 & $\sigma^2_f \exp(-\frac{||\x - \x'||^2}{2l^2})$\\
\stgpcausal & $\frac{1}{S'} \sum_{s=1}^{S'} g^{V,(s)}_{\X}(\x)$ & $\sigma^2_f \exp(-\frac{||\x - \x'||^2}{2l^2})$ + $\hat{\sigma}^V_{\doi(\X=\x)}\times \hat{\sigma}^V_{\doi(\X=\x')}$\\
\mtgp & 0 & $\sum_{q=1}^Q \left(a^V_{\X,q}\right)^2S_{\X,q}(\x, \x')$\\
\mtgpcausal & $\frac{1}{S'} \sum_{s=1}^{S'} g^{V,(s)}_{\X}(\x)$ & $\sum_{q=1}^Q \left(a^V_{\X,q}\right)^2S_{\X,q}(\x, \x')$\\
\Gmtgp & $\frac{1}{S'} \sum_{s=1}^{S'} g^{V,(s)}_{\X}(\x)$ & $
    \frac{1}{S'} \sum_{s=1}^{S'} g^{V,(s)}_{\X}(\x)g^{V,(s)}_{\X}(\x')-\Big(\frac{1}{S'}\sum_{s=1}^{S'} g^{V,(s)}_{\X}(\x)\Big)
    \Big(\frac{1}{S'}\sum_{s=1}^{S'} g^{V,(s)}_{\X}(\x')\Big)
$\\
\bottomrule
\end{tabular}
\label{tab:tablemethods}
\end{table*}

\begin{lemma}
\label{lemma:prop_an}
For any disjoint set of variables $\W_1$, $\W_2$, $\W_3$, 
if $\W_1 \cap\an(\W_2,\graph_{\overline{\W_1,\W_3}})=\emptyset$ then $\W_2\indep_{\graph_{\overline{\W_1,\W_3}}} \W_1 \,|\, \W_3 $.
\end{lemma}

\noindent \begin{proof}
$\W_1 \cap\an(\W_2,\graph_{\overline{\W_1,\W_3}})=\emptyset$ implies that there are no directed paths from (any element of) $\W_1$ to (any element of) $\W_2$ in $\graph_{\overline{\W_1,\W_3}}$. In addition, all backdoor paths from $\W_1$ to $\W_2$ in $\graph$ are removed in $\graph_{\overline{\W_1,\W_3}}$. 
Therefore, all paths from $\W_1$ to $\W_2$ in $\graph_{\overline{\W_1,\W_3}}$ are non-directed frontdoor paths. As $\W_3$ cannot be a collider on paths from $\W_1$ to $\W_2$ in $\graph_{\overline{\W_1,\W_3}}$ ($\W_3$ does not have incoming edges in $\graph_{\overline{\W_1,\W_3}}$), these paths cannot be opened by conditioning on $\W_3$. In conclusion, all paths from $\W_1$ to $\W_2$ are closed in ${\graph_{\overline{\W_1,\W_3}}}$ given $\W_3$, therefore $\W_2\indep_{\graph_{\overline{\W_1,\W_3}}} \W_1 \,|\, \W_3$.
\end{proof}\\

\noindent Let $\allmisnull{\C \cup Y}(\X, C):= \{\X^\prime \in \allmis{\C \cup Y}: \X^\prime \supset \X, \; \X^{\prime}=\X\cup\X_1$ with $\X_1 \cap \an(\CX \backslash C \cup Y,\graph_{\overline{\X^{\prime}}})=\emptyset \; \text{and} \; \C_{\X^\prime}=\CX \; \text{or} \; \X \cap \text{an}(C', \graph_{\overline{\X}}) = \emptyset \; \text{and} \; \mu^{C'} \geq \lambda^{C'} \; \text{for all} \; C' \in \CX \backslash \C_{\X^\prime}\}$; let $\allmisnull{\C \cup Y}(\X) := \Big\{\bigcup_{C \in \CX, \X \cap \an(C,\graph_{\overline{\X}})=\emptyset} [\X, \allmisnull{\C \cup Y}(\X,C)]_{\mu^C \geq \lC} \Big\}$ with $[\X, \allmisnull{\C \cup Y}(\X,C)]_{\mu^C \geq \lC} = \allmisnull{\C \cup Y}(\X,C)$ if $\mu^C \geq \lC$ and $\X$ otherwise. 

\begin{theorem}[Sufficiency of $\allmisnull{\C \cup Y}$]
\label{theorem:sufficiency_cmisnull}
$\allmisnull{\C \cup Y}:= \allmis{\C \cup Y} \backslash \bigcup_{\X \in \allmis{\C \cup Y}} \allmisnull{\C \cup Y}(\X)$ contains a solution of the \ccgo problem (if a solution exists), $\forall$ \scm  with causal graph $\graph$.
\end{theorem}

\noindent \begin{proof}
Consider a set $\X'\in \allmis{\C \cup Y}$ that satisfies \eqref{eq:ccgo}. If $\X' \not \in \allmisnull{\C \cup Y}$, then $\X'\subseteq\bigcup_{\X \in \allmis{\C \cup Y}} \allmisnull{\C \cup Y}(\X)$, and therefore there exist at least one set $\X\in \allmis{\C \cup Y}$ and at least one $C\in \CX$ with $\X \cap \an(C,\graph_{\overline{\X}})=\emptyset$ (implying $\muX^C=\mu^C$). If $\mu^C<\lambda^C$ then $[\X, \allmisnull{\C \cup Y}(\X,C)]_{\mu^C \geq \lC}=\X$, and thus $\X'\subseteq \X$ implying that $\C_{\X'}\supseteq \CX$ and therefore that $\X'$ does not satisfies \eqref{eq:ccgo} as $\mu^C<\lambda^C$. If instead $\mu^C\geq\lambda^C$, 
$\X'\in\allmisnull{\C \cup Y}(\X,C)$, and thus $\X' \supset \X$, $\X' = \X \cup \X_1$ with $\X_1 \cap \an(\CX \backslash C \cup Y,\graph_{\overline{\X'}})= \emptyset$. In addition, $\X$ has either the same constraint set of $\X'$ (\ie $\C_{\X'} = \CX$) or is such that its additional constraints are satisfied ($\X \cap \an(C', \graph_{\overline{\X}})$ and $\mu^{C'} \geq \lambda^{C'}$ for all $C' \in \CX \backslash \C_{\X'}$). Therefore $\X$ also satisfies \eqref{eq:ccgo}. If $\X \not \in \allmisnull{\C \cup Y}$ we can apply a similar reasoning and proceed until we find a set in $\allmisnull{\C \cup Y}$.
\end{proof}\\[6pt]
\noindent The \cmisreduce~procedure is given below.
\paragraph{\cmisreduce.}\label{para:searchspace} 
First construct $\allmis{\C \cup Y}$ as: $\allmis{\C \cup Y}=\emptyset$, $\forall\X \in \powerI$, if $\X \subseteq \an(\C\cup Y,\graph_{\overline{\X}}) \cup (\C \cap\X)$ add $\X$ to
$\allmis{\C \cup Y}$. Then, set $\allmisnull{\C \cup Y}$ to $\allmis{\C \cup Y}$. If $\datao\neq\emptyset$, 
$\forall \X \in \allmis{\C \cup Y}$, $\forall C \in \CX$ with $\X \cap \an(C,\graph_{\overline{\X}})=\emptyset$, if $C$ is not null-feasible (according to an estimate $\hat\mu^C$ of $\mu^C$) remove $\X$ from $\allmisnull{\C \cup Y}$. If instead $C$ is null-feasible, remove from $\allmisnull{\C \cup Y}$ all $\X^{\prime}\supset \X$ with $\X^{\prime}=\X\cup\X_1$ where $\X_1 \cap \an(\CX \backslash C \cup Y,\graph_{\overline{\X^{\prime}}})=\emptyset$ and such that $\C_{\X^\prime}=\CX$ or $\X \cap \an(C', \graph_{\overline{\X}}) = \emptyset$ and $\mu^{C'} \geq \lambda^{C'}$ for all $C' \in \CX \backslash \C_{\X'}$. 

\section{Baselines and Surrogate Models}
\label{secapp:surrogatemodels}
In the experimental section we compare the performance of \ccbo using different surrogate models against \cbo, \Cbo and, for \synone~and \syntwo, also against \acro{mcbo}. 

For the surrogate models of \cbo, \Cbo, and \stgp we assume a zero prior mean function $m_{\X}^V(\x) = 0$ and an \acro{rbf} kernel $S_{\X}^V(\x, \x') = \sigma^2_f \exp(\frac{||\x - \x'||^2}{2l^2})$ for all $\X \subseteq \powerI$ and $V \in \C_{\X} \cup Y$. \acro{rbf} kernels are also considered for the kernel functions $S_{\X,q}(\x, \x')$, $q=1,\ldots,Q$, associated to the latent \gptext{s} of \mtgp and \mtgpcausal. Notice that the surrogate model parameters for \Cbo, \cbo and \stgp are equal. Indeed, these methods only differ in terms of search space and acquisition functions used. \Cbo solves an non-causal constrained global optimization problem therefore considers only one intervention set, \ie $\X = \I$, models the associated target $\mu^Y_{\doi(\I = \x)}$ and constraint effects $\mu^{\C_{\I}}_{\doi(\I = \x)}$ independently and selects interventions via the constrained expected improvement acquisition function \citep{gardner2014bayesian}. \cbo uses the same surrogate models construction but explores the intervention sets included in $\allmisnull{\C \cup Y}$ via a standard expected improvement acquisition function thus solving an unconstrained optimization problem. Finally, \stgp considers the surrogate model constructions of \cbo and \Cbo but explores the sets included in $\allmisnull{\C \cup Y}$ selecting interventions via a constrained expected improvement acquisition function. 

For all surrogate models exploiting $\datao$ in the \gptext prior mean functions (\stgpcausal, \mtgpcausal and \Gmtgp), we compute $m_{\X}^V(\x)$ for all $\X \in \allmisnull{\C \cup Y}$ and $V \in \C_{\X} \cup Y$ by averaging the values of $V$ obtained by sampling from the fitted \scm where the functions for $\X$ are fixed to $\x$. Specifically, we model each function $f_V$ with a \gptext $f_V(\pa(v)) \sim \gp (0, S^V(\pa(v),\pa(v)'))$, where $\pa(v)$ denotes a value taken by $\pa(V)$. We use an \acro{rbf} kernel defined as $S^V(\pa(v),\pa(v)') = \sigma^2_f \exp(-\frac{||\pa(v) - \pa(v)'||^2}{2l^2})$.
We assume a Gaussian likelihood for $\datao$ and compute the posterior distribution for all $f_V$ in closed form via standard \gptext updates. We then sample from $p(\U)$ and these posteriors to obtain a set of samples $\{g_{\X}^{V, (s)}\}_{s=1}^{S'}$ for all $\X \in \allmisnull{\C \cup Y}$. The mean functions for the target and constraint effects are then obtained as $m_{\X}^V(\x) = \frac{1}{S'}\sum_{s=1}^{S'}g_{\X}^{V, (s)}$ with $V = Y$ and $V = C, \forall C \in \C$ respectively. A similar procedure is used to compute the kernel function $S_{\X}^{V, W}(\x, \x')$ for \Gmtgp. In particular, given the samples $\{g_{\X}^{V, (s)}\}_{s=1}^{S'}$ and $\{g_{\X}^{W, (s)}\}_{s=1}^{S'}$ for each couple of variables $(V, W)$ in $\C \cup Y$, we compute the correlation across the associated effects as $\frac{1}{S'} \sum_{s=1}^{S'} g^{V,(s)}_{\X}(\x)g^{W,(s)}_{\X}(\x')-\Big(\frac{1}{S'}\sum_{s=1}^{S} g^{V,(s)}_{\X}(\x)\Big) \Big(\frac{1}{S'}\sum_{s=1}^{S'} g^{W,(s)}_{\X}(\x')\Big)$. See Table \ref{tab:tablemethods} for a summary of the prior parameters used for each surrogate model. 

Similarly to \cbo, \acro{mcbo} solves an unconstrained optimization problem. Instead of explicitly modelling the target effect, \acro{mcbo} assumes each $V \in \V$ to be of the form $V = f_{V}(\pa_{\graph}(V), \mathbf{A}_V) + U_V$, where $\mathbf{A}_V$ is a set of action variables whose values can be set by the investigator. It then places a vector-valued \gptext prior on the functions $\{f_V\}_{V\in\V}$ and exploits the posterior distribution of this \gptext together with a reparameterization trick introduced by \citet{curi2020efficient} to derive confidence bounds for the target effect. These bounds are then used within an upper confidence bound acquisition function. Notice that, by explicitly modelling the functions in the \scm, \acro{mcbo} cannot deal with settings in which there are unobserved confounders. 

\begin{figure}[t]
\centering
\includegraphics[width=0.46\textwidth]{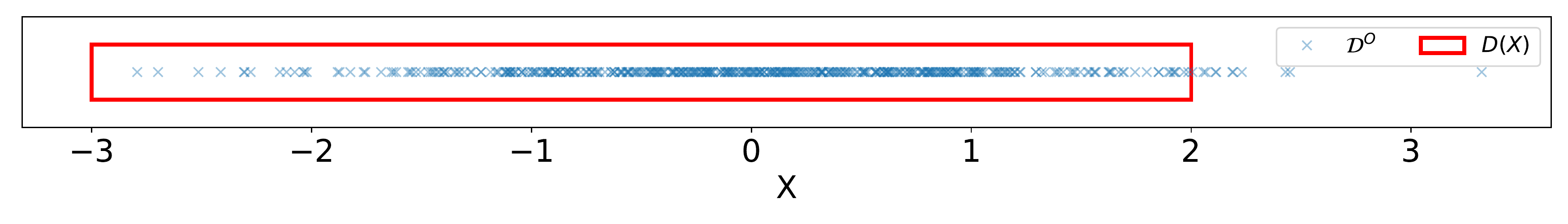}
\includegraphics[width=0.46\textwidth]{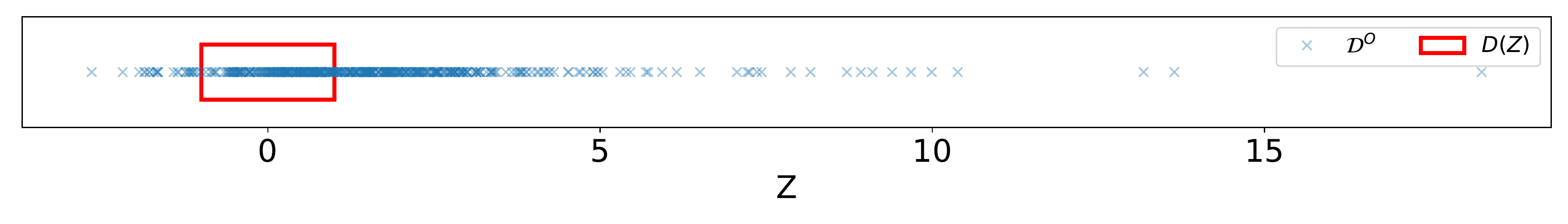}
\caption{\synone~with $\Nobs=500$. Scatter plots for the observational data together with interventional ranges $D(X)$ (top) and $D(Z)$ (bottom).}
\label{fig:scatter_synone}
\end{figure}

\section{Experimental Details and Additional Results}
\label{secapp:Exp}
\begin{figure}[t]
\centering
\includegraphics[width=0.46\textwidth]{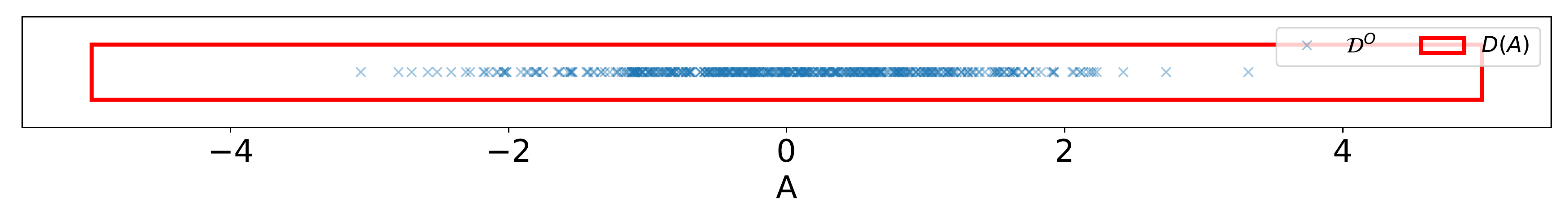}
\includegraphics[width=0.46\textwidth]{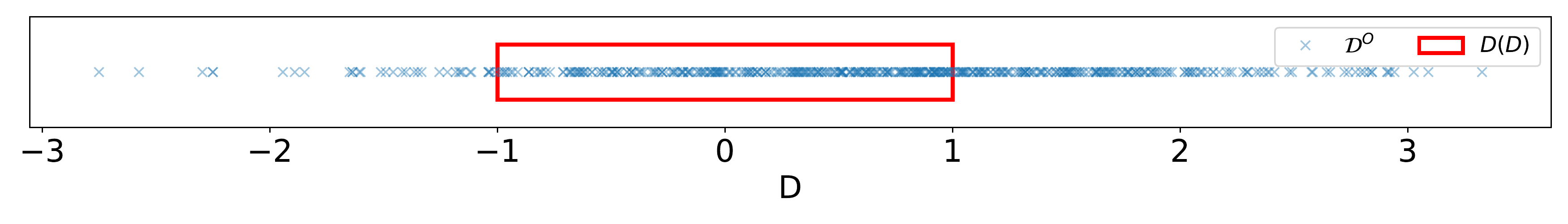}
\includegraphics[width=0.46\textwidth]{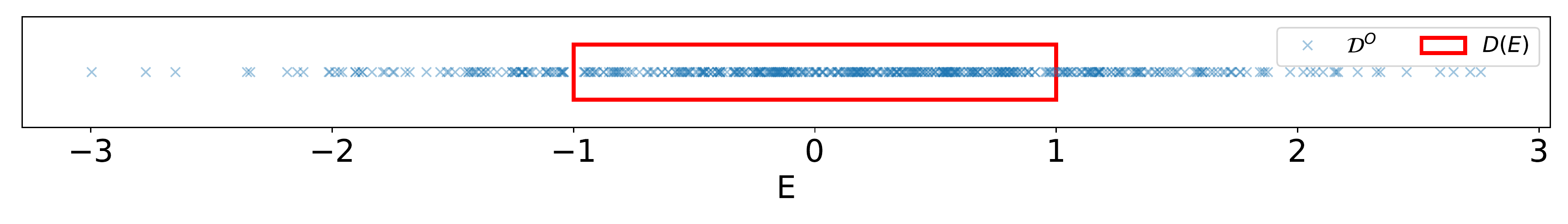}
\caption{\syntwo~with $\Nobs=500$. Scatter plots for the observational data together with interventional ranges $D(A)$ (top), $D(D)$ (center) and $D(E)$ (bottom).}
\label{fig:scatter_syntwo}
\end{figure}
For all experiments we initialize the kernel hyper-parameters $(\sigma^2_f, l)$ of the surrogate models to 1 and optimize them with a standard type-2 \acro{ml} approach. When sampling from the modified \scm is required in order to compute the prior mean or kernel functions we use $S' = 10$ samples. The same number of samples is used when the computation of the acquisition function requires a Monte Carlo approximation. 

\begin{figure}[t]
\centering
\includegraphics[width=0.45\textwidth]{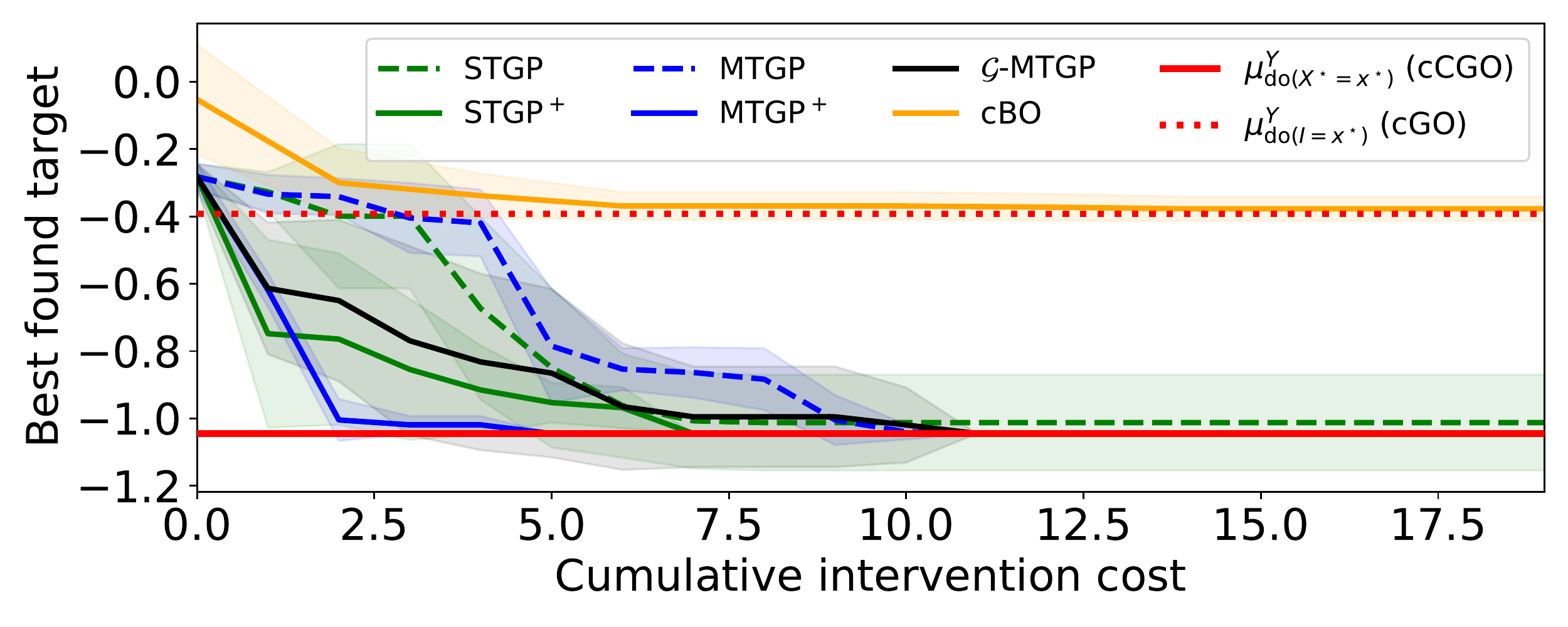}
\includegraphics[width=0.45\textwidth]{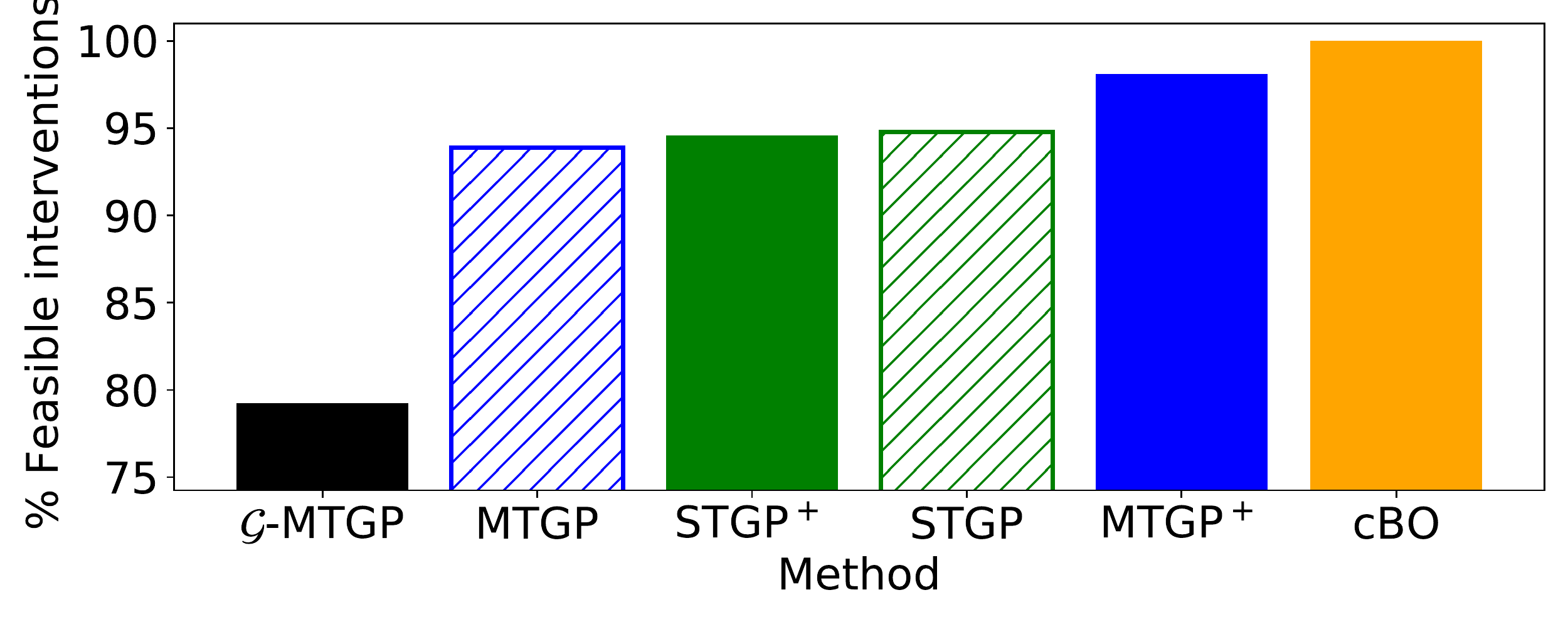}
\caption{$\synone$ with $\Nobs=10$ and $\lambda^Z=2$. \textit{Top}: Convergence to \ccgo (solid red line) and \Cgo (dotted red line) optima. Lines give average results across different initialization of $\datai$. Shaded areas represents $\pm$ standard deviation. \textit{Bottom}: Average percentage of feasible interventions collected over trials.}
\label{fig:synone_N10_lambda2}
\end{figure}

\subsection{\synone}
\label{sec:appsynthetic1}
\begin{figure*}[t]
\centering
\includegraphics[width=0.44\textwidth]{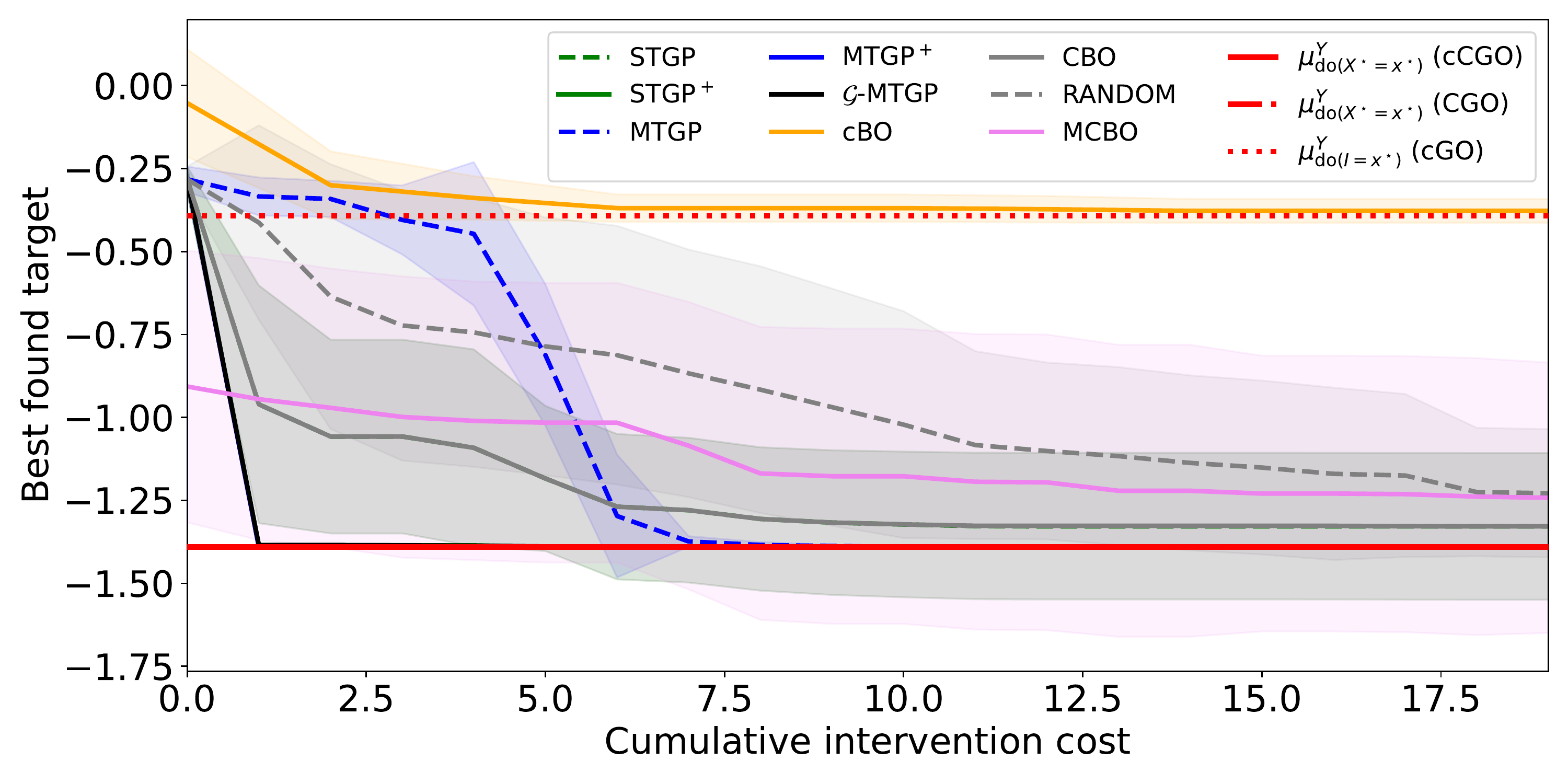}
\includegraphics[width=0.44\textwidth]{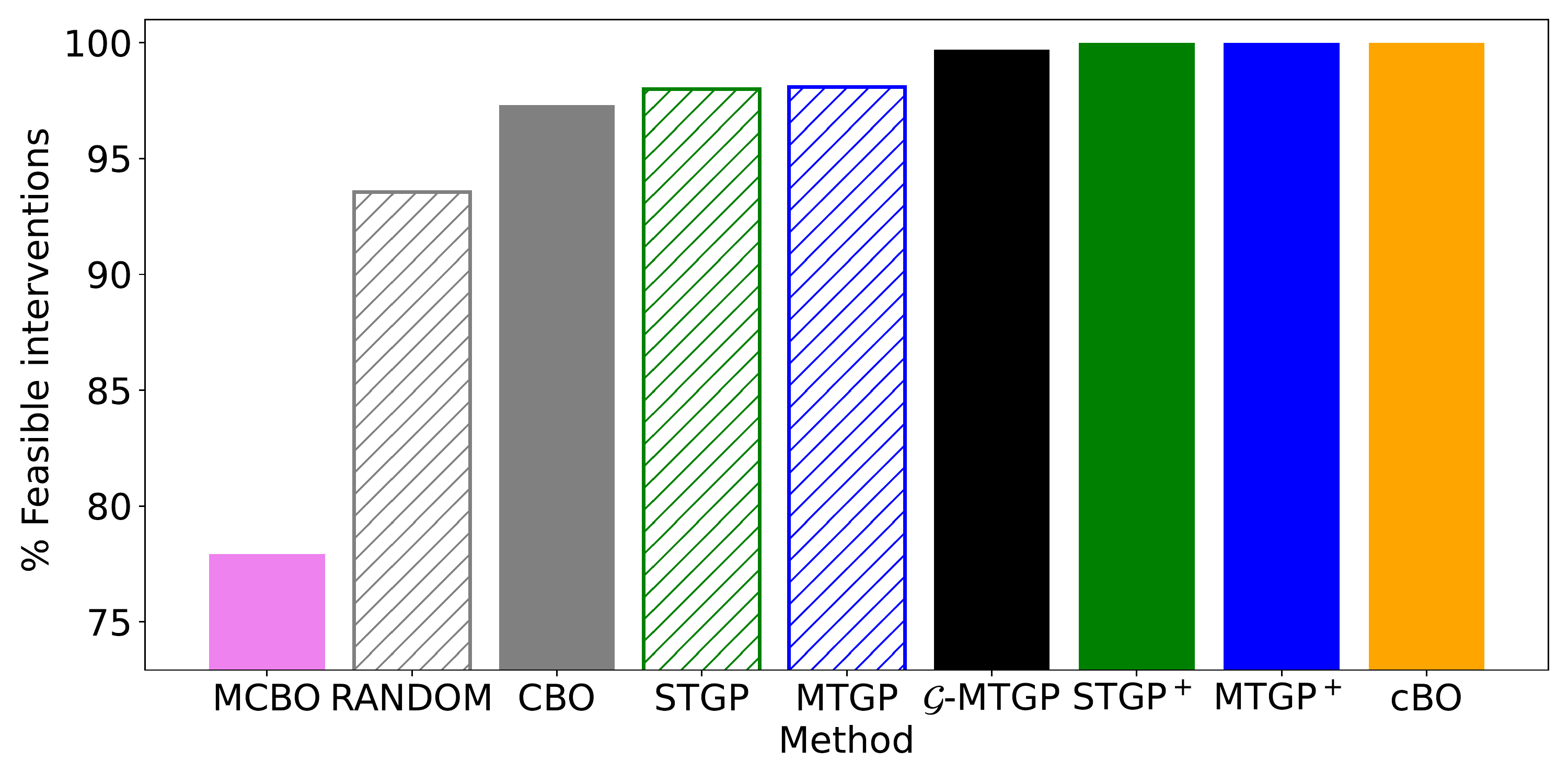}
\includegraphics[width=0.44\textwidth]{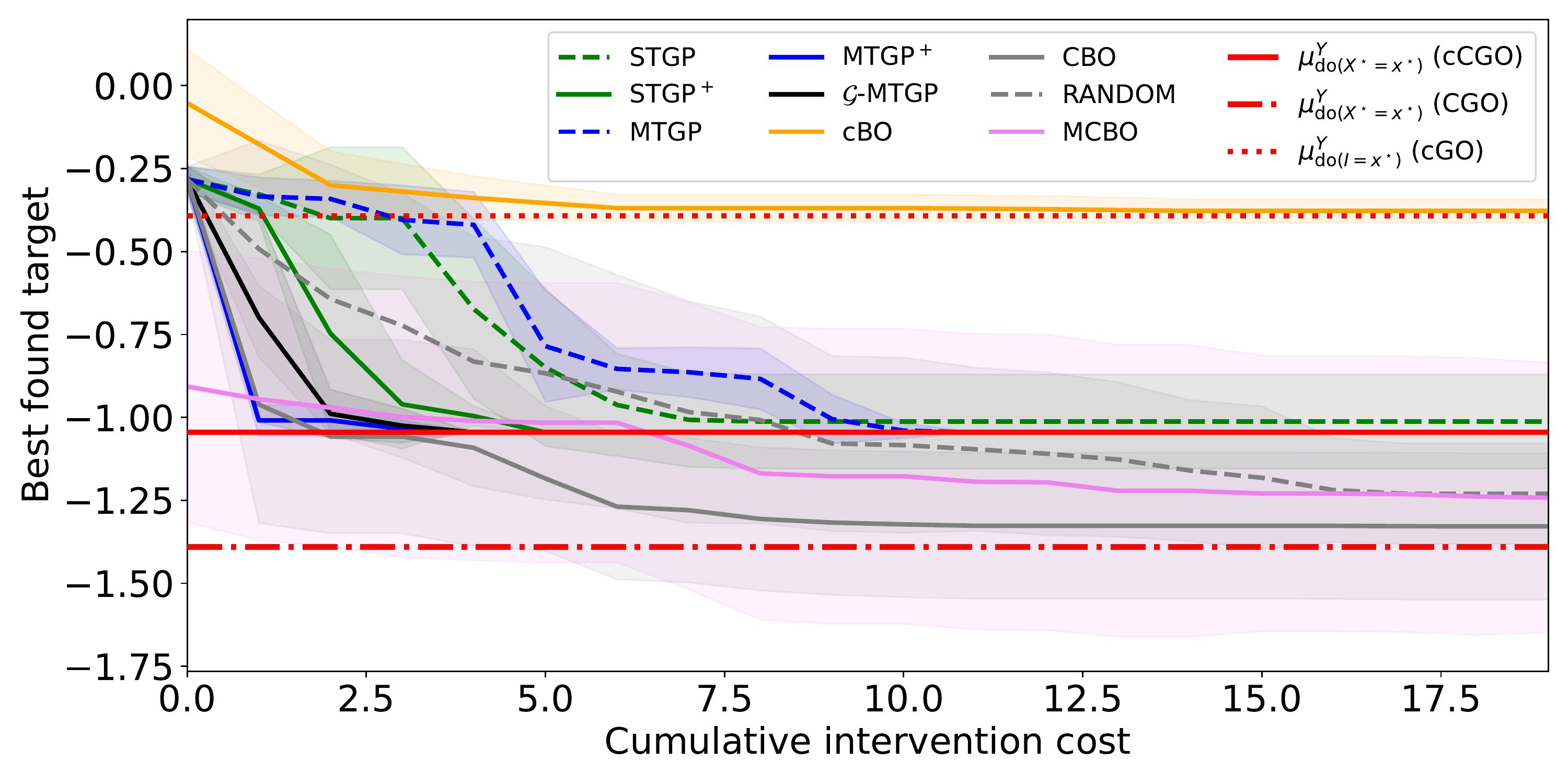}
\includegraphics[width=0.44\textwidth]{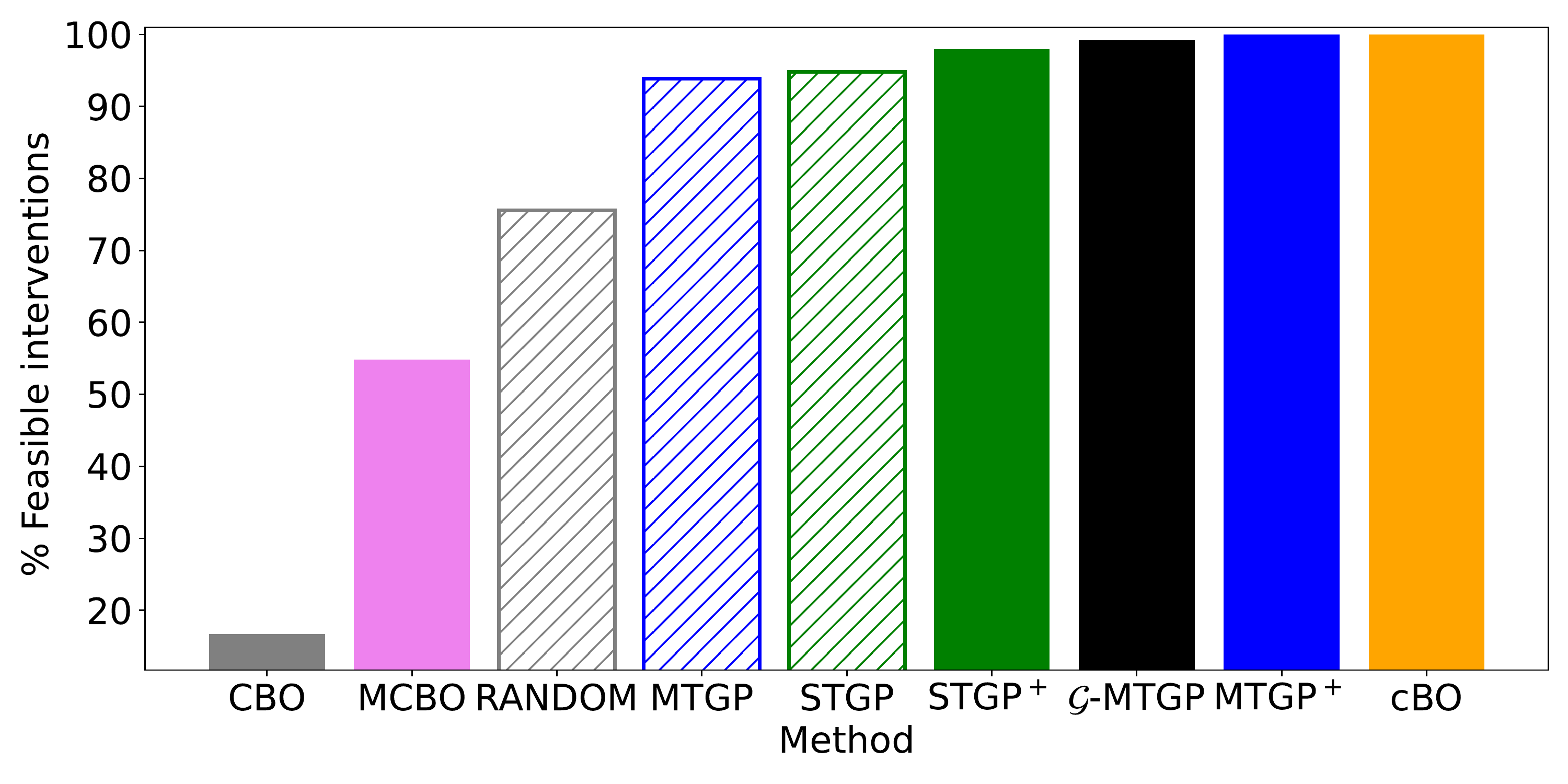}
\caption{\synone~with $\Nobs=500$ and $\lambda^Z = 10$ (top row), and with $\Nobs=500$ and $\lambda^Z = 2$ (bottom row). \textit{Left}: Convergence to the \ccgo (solid red line), \cgo (dash-dotted red line) and \Cgo (dotted red line) optima.
\textit{Right}: Average percentage of feasible interventions collected over trials.}
\label{fig:all_synone_N500_lambda2_lambda10}
\end{figure*}
\begin{figure}[t]
\centering
\includegraphics[width=0.45\textwidth]{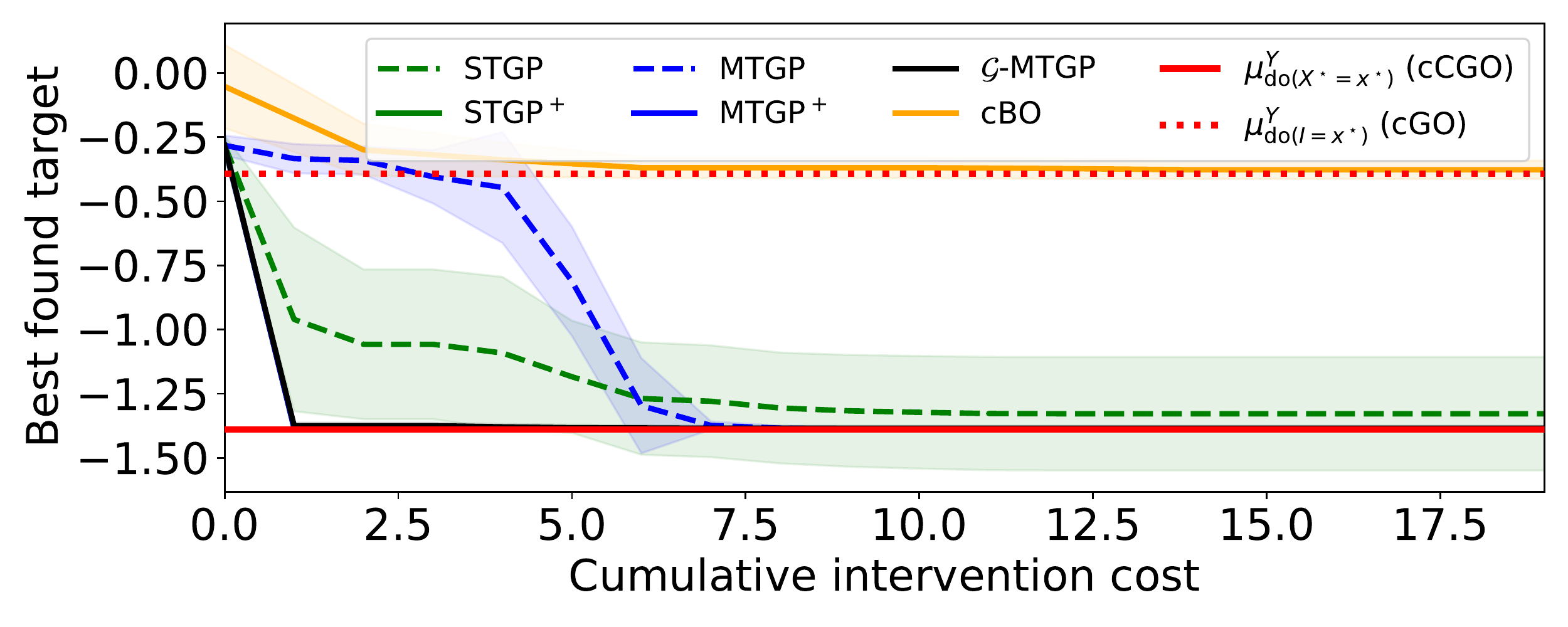}
\includegraphics[width=0.45\textwidth]{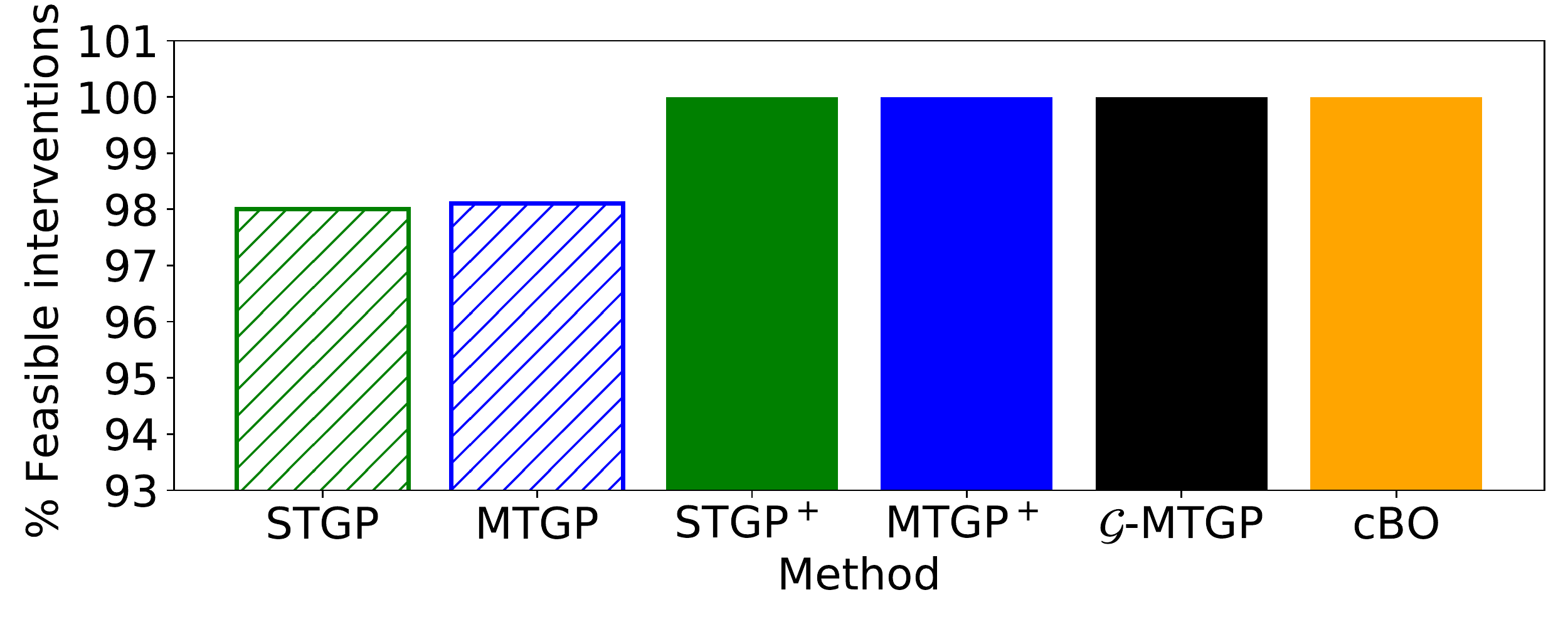}
\caption{\synone~with $\Nobs=100$ and $\lambda^Z = 10$. \textit{Top}: Convergence to the \ccgo (solid red line) and \Cgo (dotted red line) optima.
\textit{Bottom}: Average percentage of feasible interventions collected over trials.}
\label{fig:synone_N100_lambda10}
\end{figure}
For the causal graph in \figref{fig:causalgraphs2-3}(b) with $\I=\C=\{X,Z\}$ we consider the \scm
\begin{align*}
X &= U_X, \,\, Z = \exp(-X) + U_Z,\\
Y &= \cos(Z) - \exp(-Z/20) + U_Y,
\end{align*}
with $U_X,U_Z,U_Y \sim \mathcal{N}(0,1)$. We set the interventional ranges to $D(X) = [-3,2]$ and $D(Z) = [-1,1]$, the constraint thresholds to $(\lambda^{X}, \lambda^{Z}) = (1, 2)$ for \figref{fig:synone_N500_N100_lambda2}, \figref{fig:synone_N10_lambda2} and \figref{fig:all_synone_N500_lambda2_lambda10} (bottom row) and to $(\lambda^{X}, \lambda^{Z}) = (1, 10)$ for \figref{fig:all_synone_N500_lambda2_lambda10} (top row) and \figref{fig:synone_N100_lambda10}, and require the constraint effects to be lower that the thresholds. Notice that, even in cases when $\Nobs$ is high, there exists a mismatch between the observational and interventional ranges. \figref{fig:scatter_synone} shows how the interventional ranges are only partially covered by the observational data, especially for $X$. Therefore, estimating the target and constraint effects with $\datao$ for interventions values outside of the observational ranges, say for $X = -2$, might lead to inaccurate results which translate to unreliable prior parameters for the surrogate models exploiting $\datao$. A combination of interventional and observational data is needed in order to quantify uncertainty around the effects and to choose the intervention to perform next while ensuring satisfaction of the constraints, and thus to efficiently identify an optimal feasible intervention.

In this graph, $\powerI = \{\{X\}, \{Z\}, \{X,Z\}\}$, and $\allmis{\C \cup Y} = \powerI$ as  $\X \subseteq \an(\{X,Z,Y\}, \graph_{\overline{\X}}) \cup (\{X,Z\} \cap \X)$, $\forall \X \in \powerI$. 
The set $\X =\{Z\}$ has only one constrained variable, $X$, that is reducible (as $\X \cap \an(X, \graph_{\overline{\X}})=\emptyset$) and null-feasible for $\Nobs=500, 100, 10$. We can thus exclude $\X^\prime=\{X,Z\} \supset \X$ from $\allmis{\C \cup Y}$. Indeed, $\{X,Z\}$ is of the form $\X^\prime=\X \cup X$ with $X \notin \an(\emptyset \cap Y, \graph_{\overline{\X^\prime}})$. We therefore obtain $\allmisnull{\C \cup Y} =\{\{X\}, \{Z\}\}$.

\figref{fig:all_synone_N500_lambda2_lambda10} (top row) shows the convergence results for  $\Nobs=500$ and $\lambda^Z = 10$, for which the \ccgo and \cgo problems have the same optimum. Notice how both \cbo and \acro{mcbo} converge to the optimum at a slower pace compared to \stgpcausal, \mtgpcausal and \Gmtgp. Indeed, \cbo and \acro{mcbo} only take into account the target value observed after performing each intervention. \cbo also models each function individually and, by disregarding the values of the constraints, needs to perform more interventions before identifying an optimal one. More importantly, \mtgpcausal, \stgpcausal and \Gmtgp collect more than 99\% of feasible interventions over trials. This is a critical aspect of the method as in real-world applications the investigator might not know whether an optimal solution is in a feasible region or not. Using \ccbo in such settings does not slow down the identification of an optimal solution, but rather improves it, while allowing to efficiently restrict the exploration regions. 
For completeness, in \figref{fig:all_synone_N500_lambda2_lambda10} (bottom row) we also report the comparison with \cbo, \acro{mcbo} and \acro{random} for $\Nobs=500$ and $\lambda^Z=2$.
\begin{figure}
\centering
\includegraphics[width=0.45\textwidth]{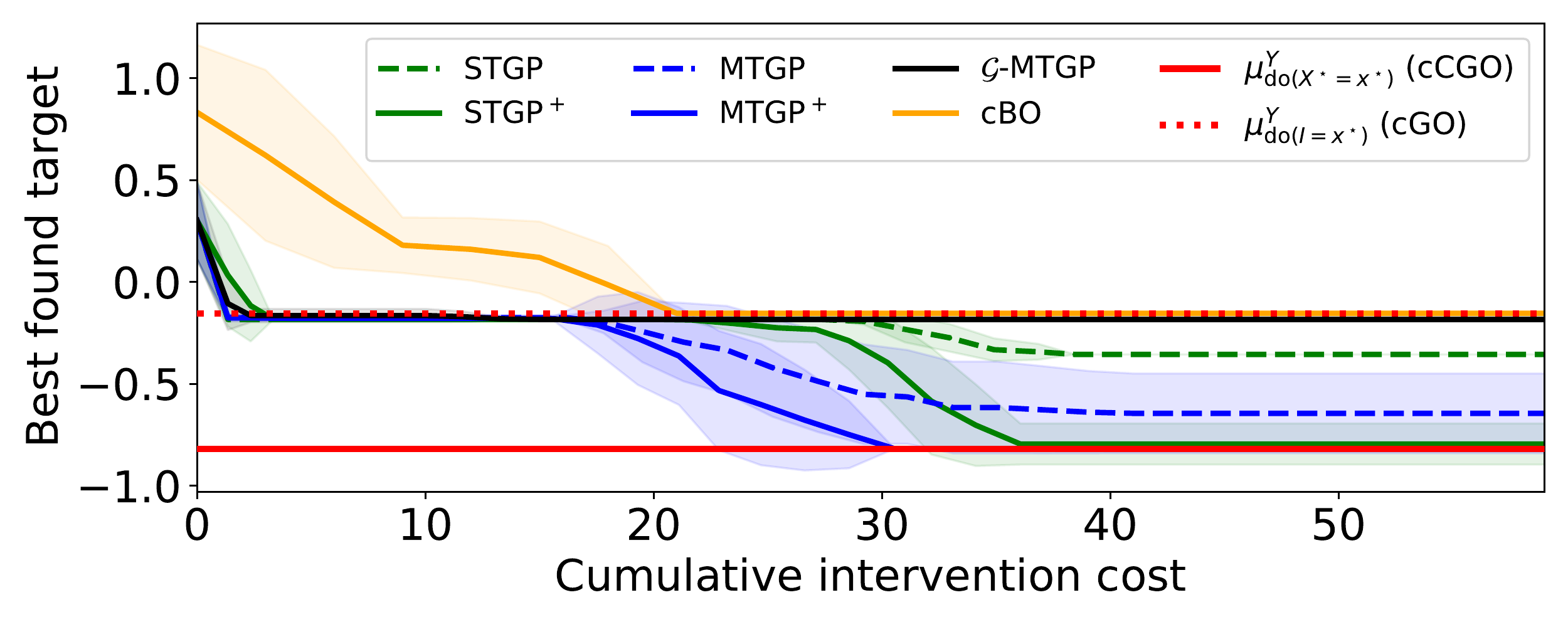}
\includegraphics[width=0.45\textwidth]{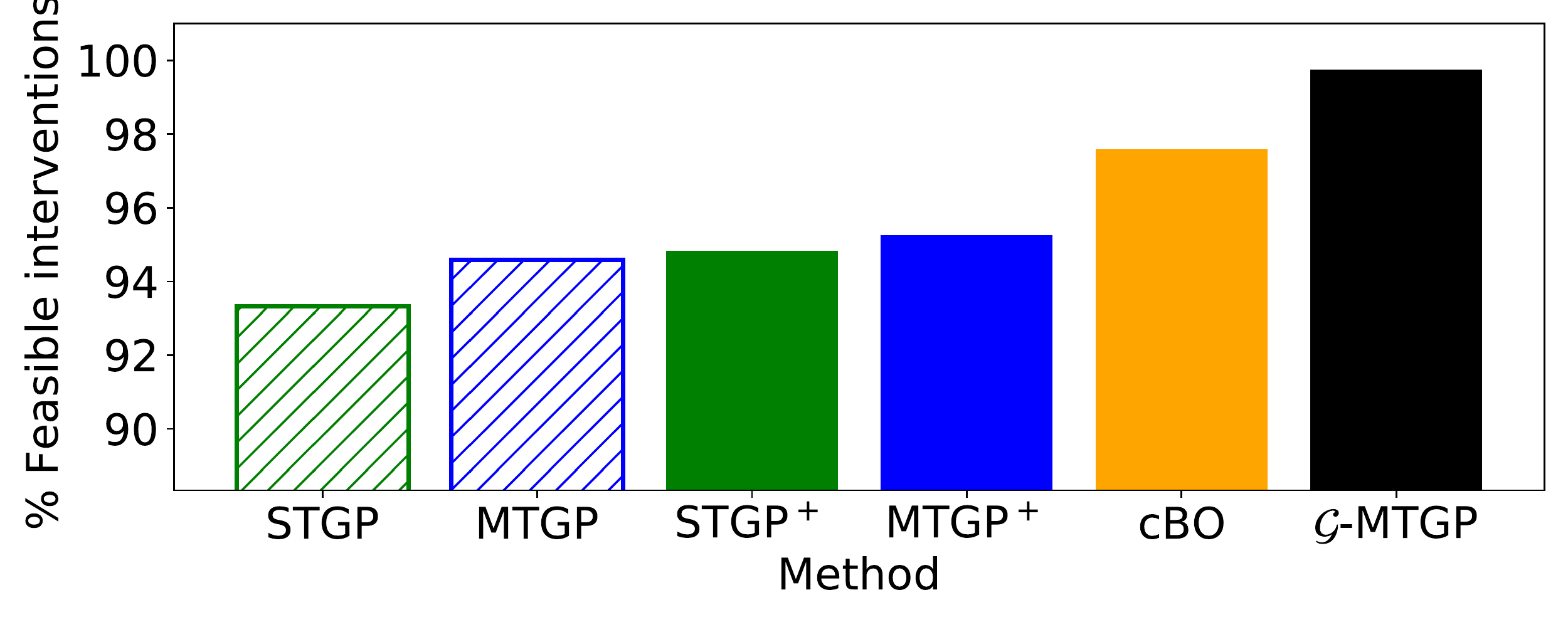}
\caption{\syntwo~with $\Nobs=10$. \textit{Top}: Convergence to the \ccgo (solid red line) and \Cgo (dotted red line) optima. \textit{Bottom}: Average percentage of feasible interventions collected over trials.}
\label{fig:syntwo_N10}
\end{figure}
\subsection{\syntwo}
\label{sec:appsynthetic2}
For the causal graph in \figref{fig:causalgraphs2-3}(a) with $\I = \{A,D,E\}$ and $\C =\{C,D,E\}$ we consider the \scm
\begin{align*}
A &= U_A,\,\, B = U_B,\\
C &= \exp(-A)/5 + U_C,\\
D &= \text{cos}(B) + C/10 + U_D,\\
E &= \exp(-C)/10 + U_E,\\
Y &= \text{cos}(D) -D/5 + \text{sin}(E) - E/4 + U_Y,
\end{align*}
with $U_A,U_B,U_C,U_D,U_E,U_Y \sim \mathcal{N}(0,1)$.
We set the interventional ranges to $D(A) = [-5,5]$, $D(D) = [-1,1]$ and $D(E) = [-1,1]$, the constraint thresholds to $\lambda^C = 10$, $\lambda^D = 10$, $\lambda^E = 10$, and require all constraint effects to be smaller than the thresholds. 
\begin{figure*}
\centering
\includegraphics[width=0.45\textwidth]{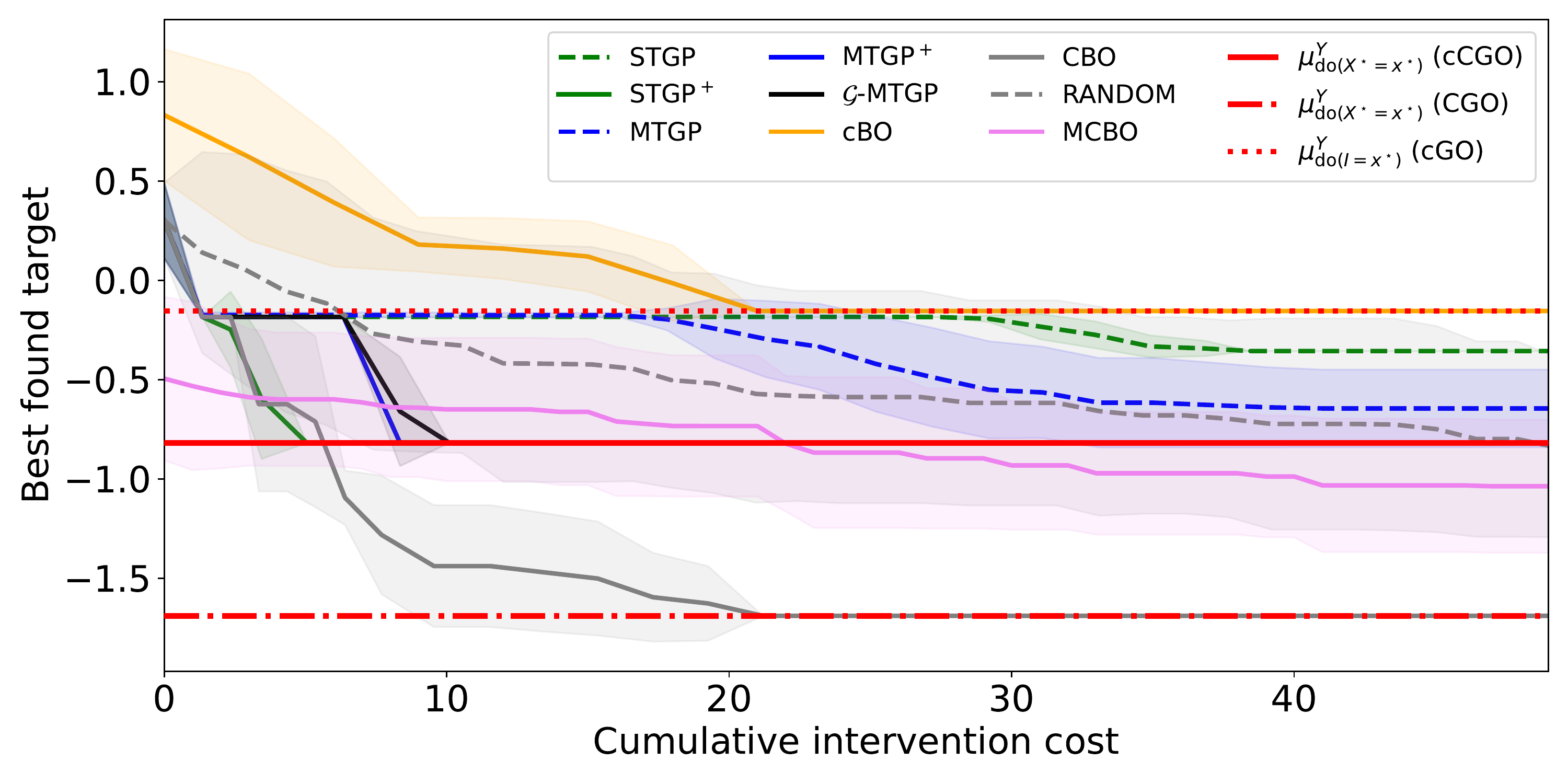}
\includegraphics[width=0.45\textwidth]{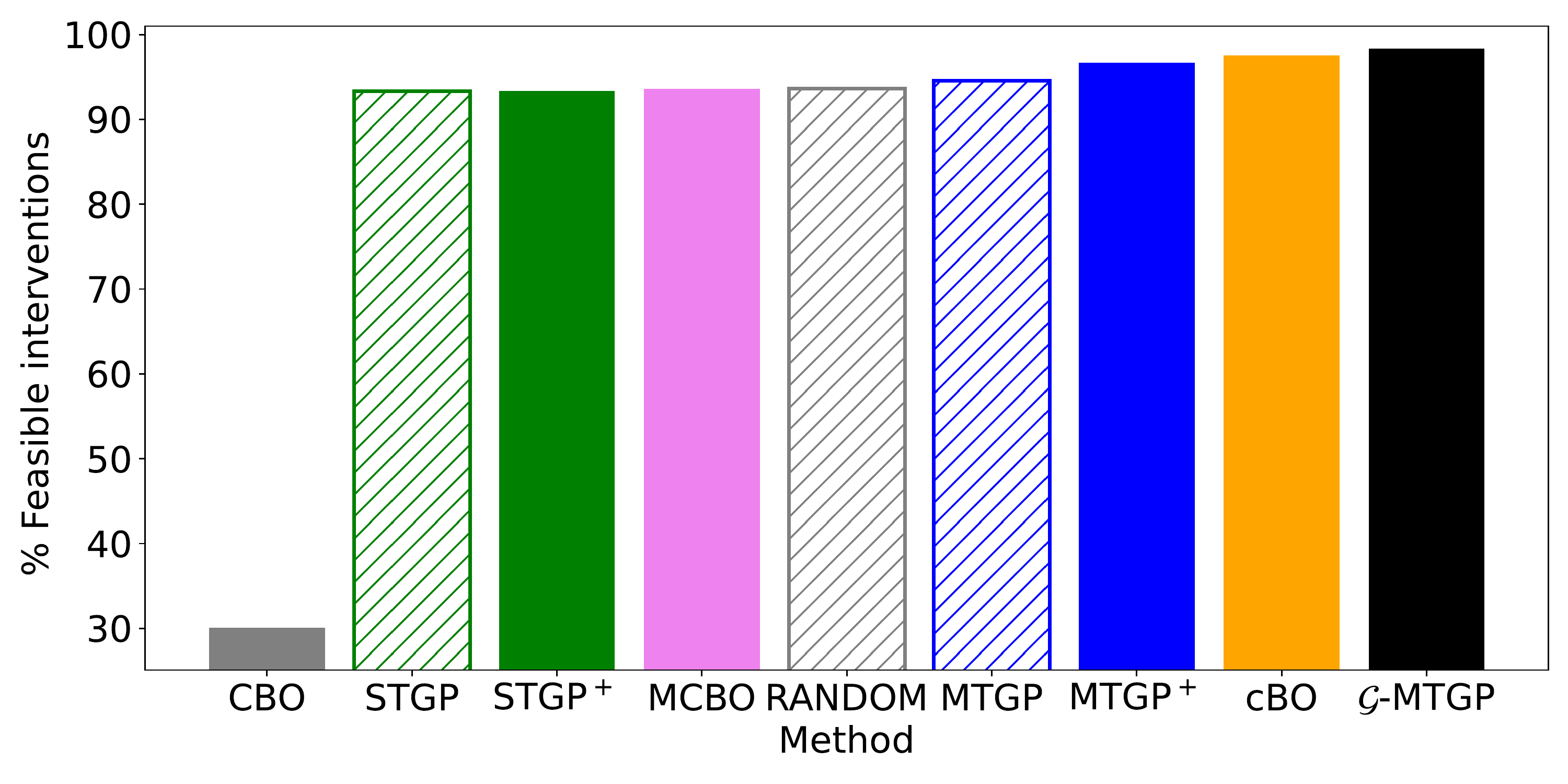}
\caption{\syntwo~with $\Nobs=500$. \textit{Top}: Convergence to the \ccgo (solid red line), \cgo (dash-dotted red line) and \Cgo (dotted red line) optima.
\textit{Bottom}: Average percentage of feasible interventions collected over trials.}
\label{fig:all_syntwo_N500}
\end{figure*}
For $\Nobs=500, 100, 10$ all variables in $\C$ are null-feasible, giving $\allmisnull{\C \cup Y} = \powerI \backslash \{A,D,E\}=\{\{A\},\{D\},\{E\},\{A,D\},\{A,E\},\{D,E\}\}$ (see the example in \secref{sec:rss}). As in \synone, even in cases when $\Nobs$ is high, there exists a mismatch between the observational and interventional ranges (\figref{fig:scatter_syntwo}) for all interventions sets in $\allmisnull{\C \cup Y}$. 

\figref{fig:syntwo_N10} shows the convergence plots and associated percentage of feasible interventions when $\Nobs=10$. As discussed in \secref{sec:discussion}, when the functions in the \scm cannot be learned accurately due to limited or noisy observational data, \mtgpcausal should be preferred. Indeed, a lower value for $\Nobs$ leads to a less accurate estimation of the target and constraint effects from $\datao$ which negatively affects the prior mean functions of \stgpcausal, \mtgpcausal and \Gmtgp. \Gmtgp is further penalized in this case as also the kernel functions are computed exploiting the fitted \scm functions. As a consequence, we observe slower convergence and a lower percentage of feasible interventions collected when using \Gmtgp. In particular, notice how the inaccurate estimation of the constraint effects translates into a significantly lower convergence speed for \Gmtgp compared to the settings in which $\Nobs$ is higher. On the contrary, \mtgpcausal successfully trades off feasibility and improvement in these experiments reaching convergence while collecting a high percentage of feasible interventions over trials. Finally, by breaking the existing causal relationships and intervening on all variables in $\I$, \Cbo blocks the propagation of causal effects in the graph and converges to a solution (\Cgo) that is sub-optimal with respect to the one achieved by all \ccbo instances. 

Finally, \figref{fig:all_syntwo_N500} shows how, in settings where the \ccgo and \cgo optima differ, \cbo, \acro{mcbo}, and \acro{random} converge to a solution which is lower than the \ccgo one. In addition, by disregarding the constraints, \cbo, \acro{mcbo}, and \acro{random} collect a high number of infeasible interventions over trials.

\begin{figure}[t]
\centering
\includegraphics[width=0.46\textwidth]{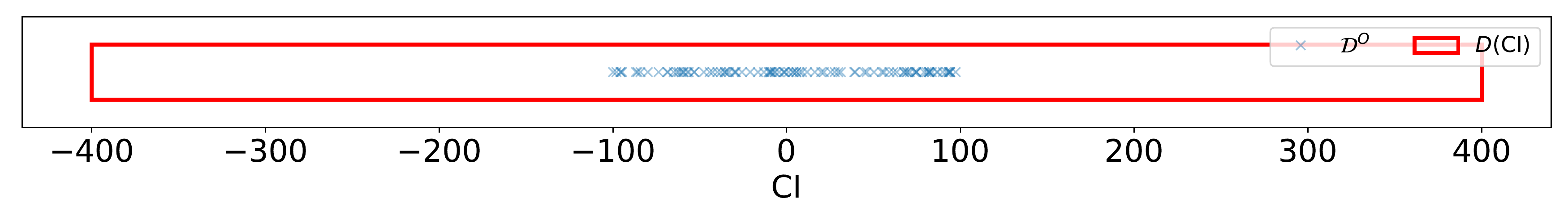}
\includegraphics[width=0.46\textwidth]{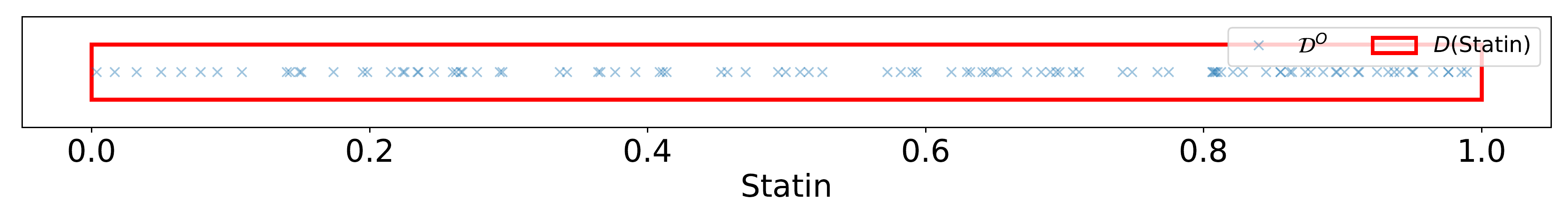}
\includegraphics[width=0.46\textwidth]{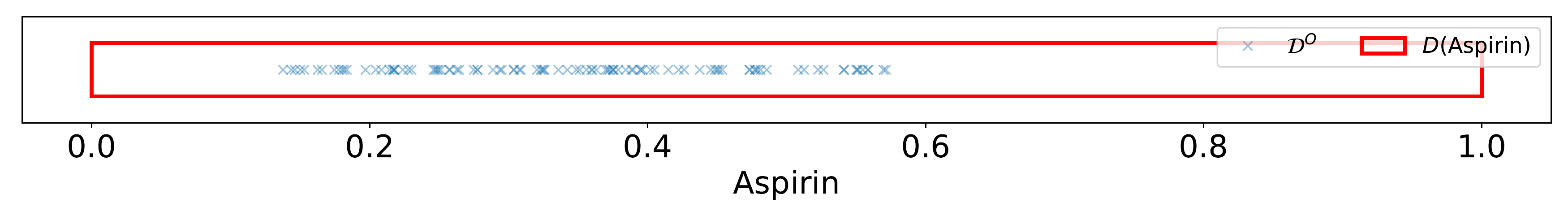}
\caption{\health with $\Nobs=100$. Scatter plots for the observational data together with interventional ranges $D(\acro{ci})$ (top), $D(\text{Statin})$ (center) and $D(\text{Aspirin})$ (bottom).}
\label{fig:scatter_health}
\end{figure}
\subsection{\health}
\label{sec:appreal1}
For the causal graph in \figref{fig:causalgraphs1}(b) with $\I = \{\text{Statin}, \text{Aspirin}, \acro{ci}\}$ and $\C=\{\bmi\}$, the \scm 
is taken from \citet{ferro2015use} and can be written as
\begin{align*}
&\text{Age} = U_{\text{Age}}, \,\, \text{\acro{ci}} = U_{\text{\acro{ci}}}, \,\, \text{\acro{bmr}} = 1500 + 10 \times U_{\text{\acro{bmr}}},\\
&\text{Height} = 175 + 10\times U_{\text{Height}},\\
&\text{Weight} = \frac{\acro{bmr} + 6.8 \times \text{Age} - 5 \times \text{Height}}{(13.7 + \text{\acro{ci}}\times 150/7716)},\\
&\text{\bmi} = \text{Weight} / (\text{Height} / 100)^2,\\
&\text{Aspirin} = \sigma(-8.0 + 0.10\times \text{Age}  + 0.03 \times \text{\bmi}),\\
&\text{Statin} = \sigma(-13.0 + 0.10\times \text{Age}  + 0.20\times \text{\bmi}),\\
&\acro{psa} = 6.8 + 0.04 \times \text{Age}  - 0.15 \times \text{\bmi}  - 0.60 \times \text{Statin} \\ 
&\hspace{0.9cm} + 0.55 \times \text{Aspirin} + \sigma(2.2 - 0.05\times \text{Age} \\ 
&\hspace{0.9cm}  + 0.01 \times \text{\bmi} - 0.04 \times \text{Statin}  \\ 
&\hspace{0.9cm}  + 0.02\times \text{Aspirin}) + U_{\acro{psa}},
\end{align*}
with $U_{\text{Age}} \sim \mathcal{U}(55, 75)$,  $U_{\text{\acro{ci}}} \sim \mathcal{U}(-100, 100)$, $U_{\text{\acro{bmr}}} \sim  t\mathcal{N}(-1, 2)$, $U_{\text{Height}}\sim t\mathcal{N}(-0.5, 0.5)$, $U_{\acro{psa}} \sim \mathcal{N}(0, 0.4)$, where $\mathcal{U}(\cdot, \cdot)$ denotes a uniform random variable, $t\mathcal{N}(a, b)$ a standard Gaussian random variable truncated between $a$ and $b$, and $\sigma(\cdot)$ the sigmoidal transformation defined as $\sigma(\x) = \frac{1}{1 + \exp(-x)}$. We set the interventional ranges to $D(\text{Statin}) = [0,1]$, $D(\text{Aspirin}) = [0,1]$ and $D(\text{\acro{ci}}) = [-400, 400]$, and require \bmi to be lower than 25.

\begin{figure}
\centering
\includegraphics[width=0.20\textwidth]{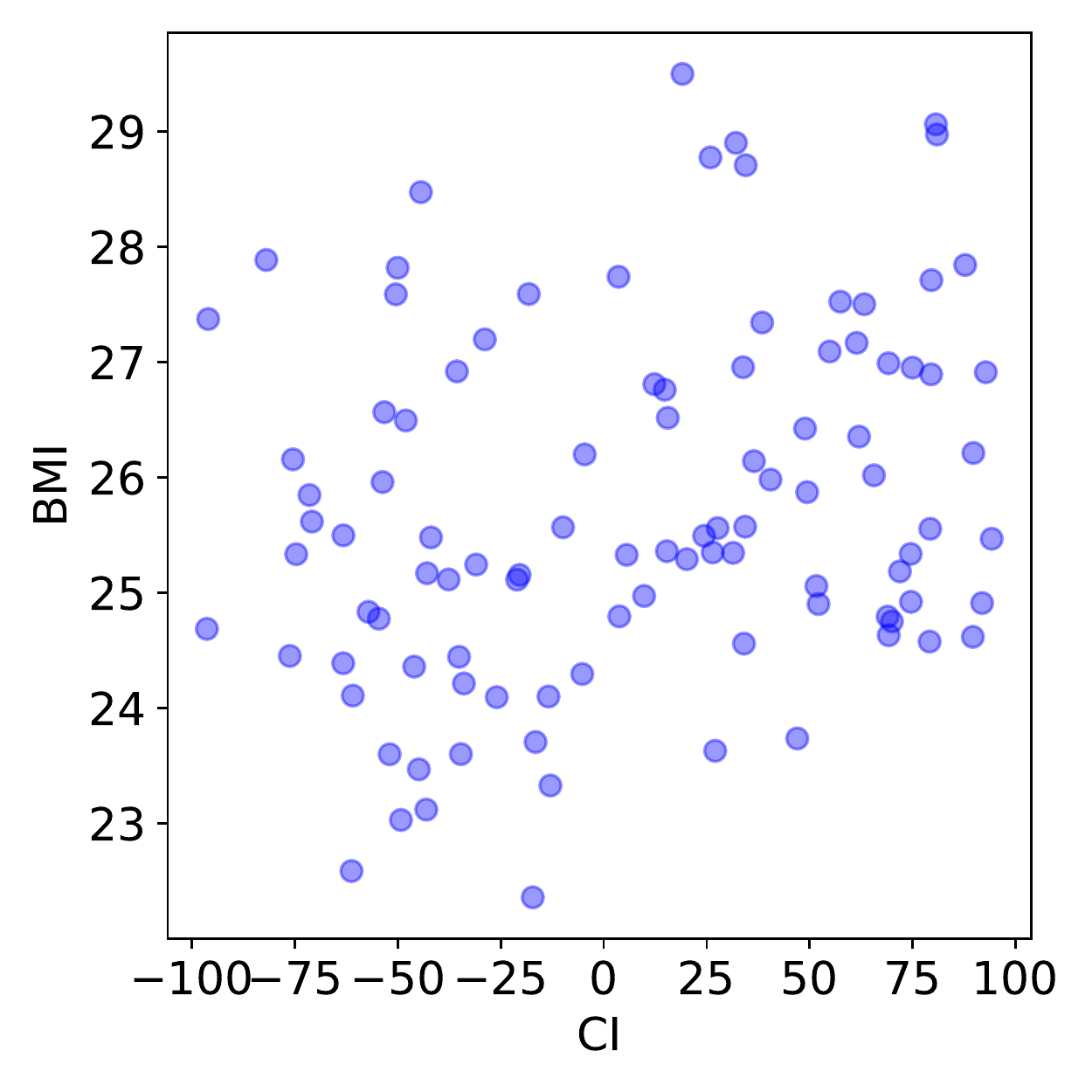}
\includegraphics[width=0.20\textwidth]{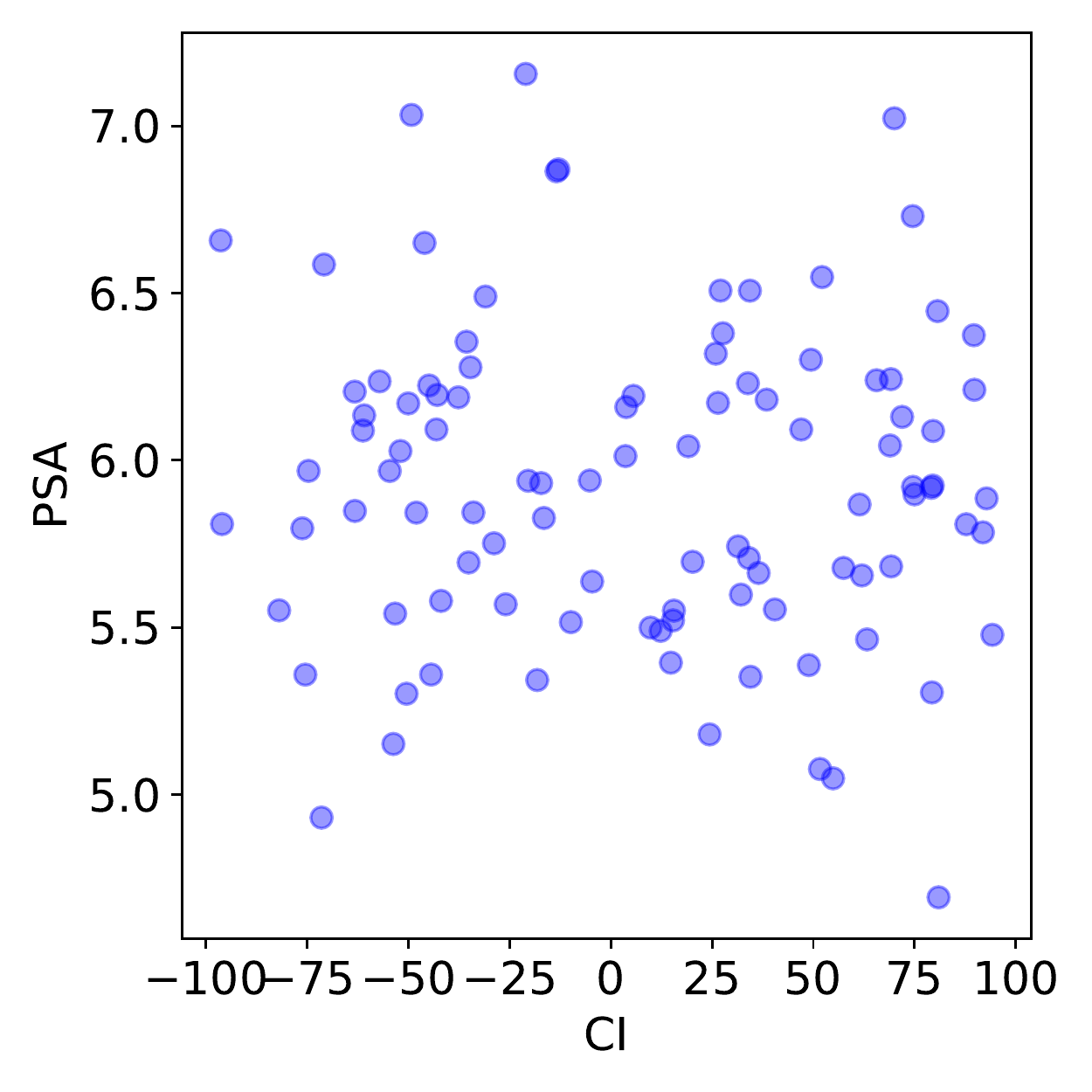}
\includegraphics[width=0.20\textwidth]{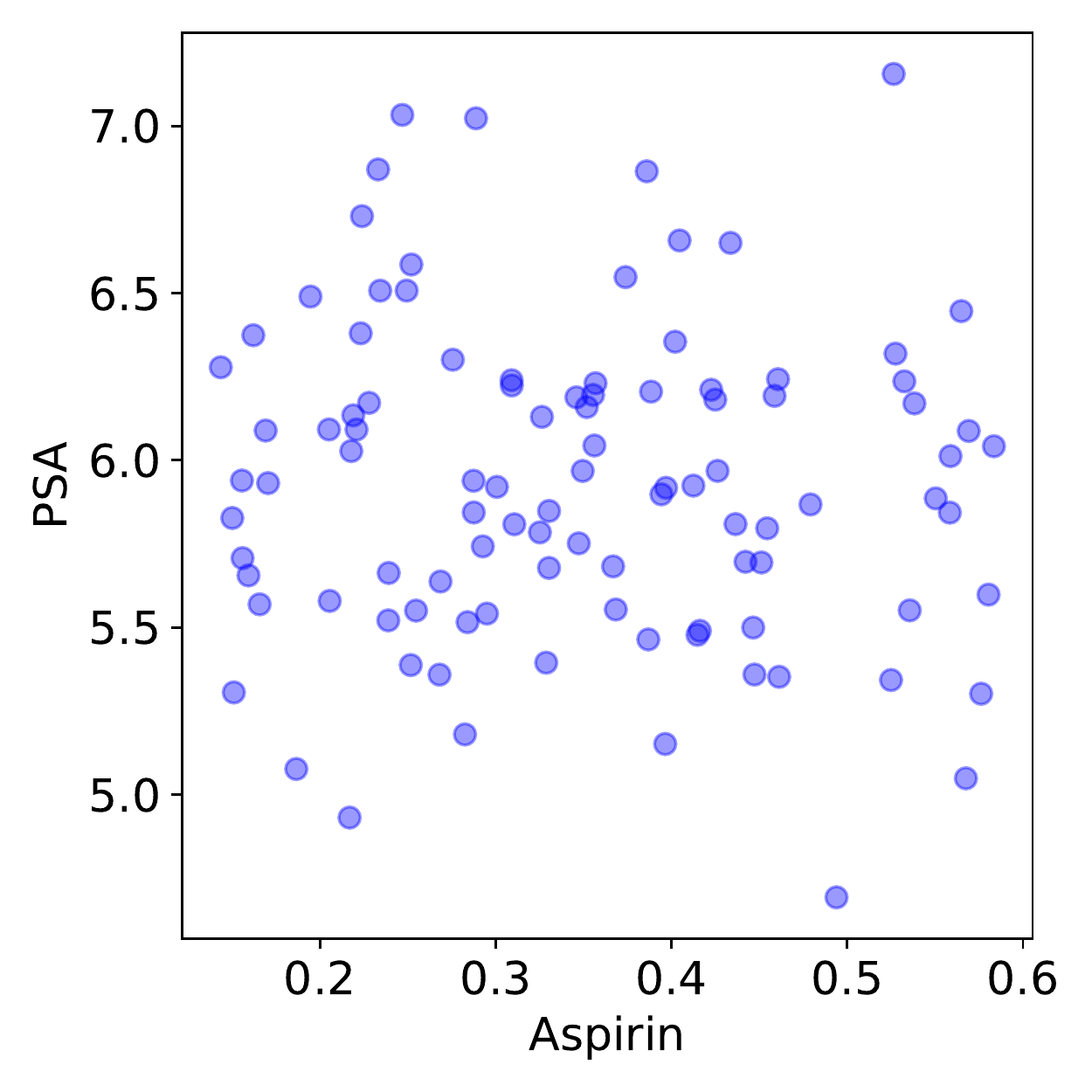}
\includegraphics[width=0.20\textwidth]{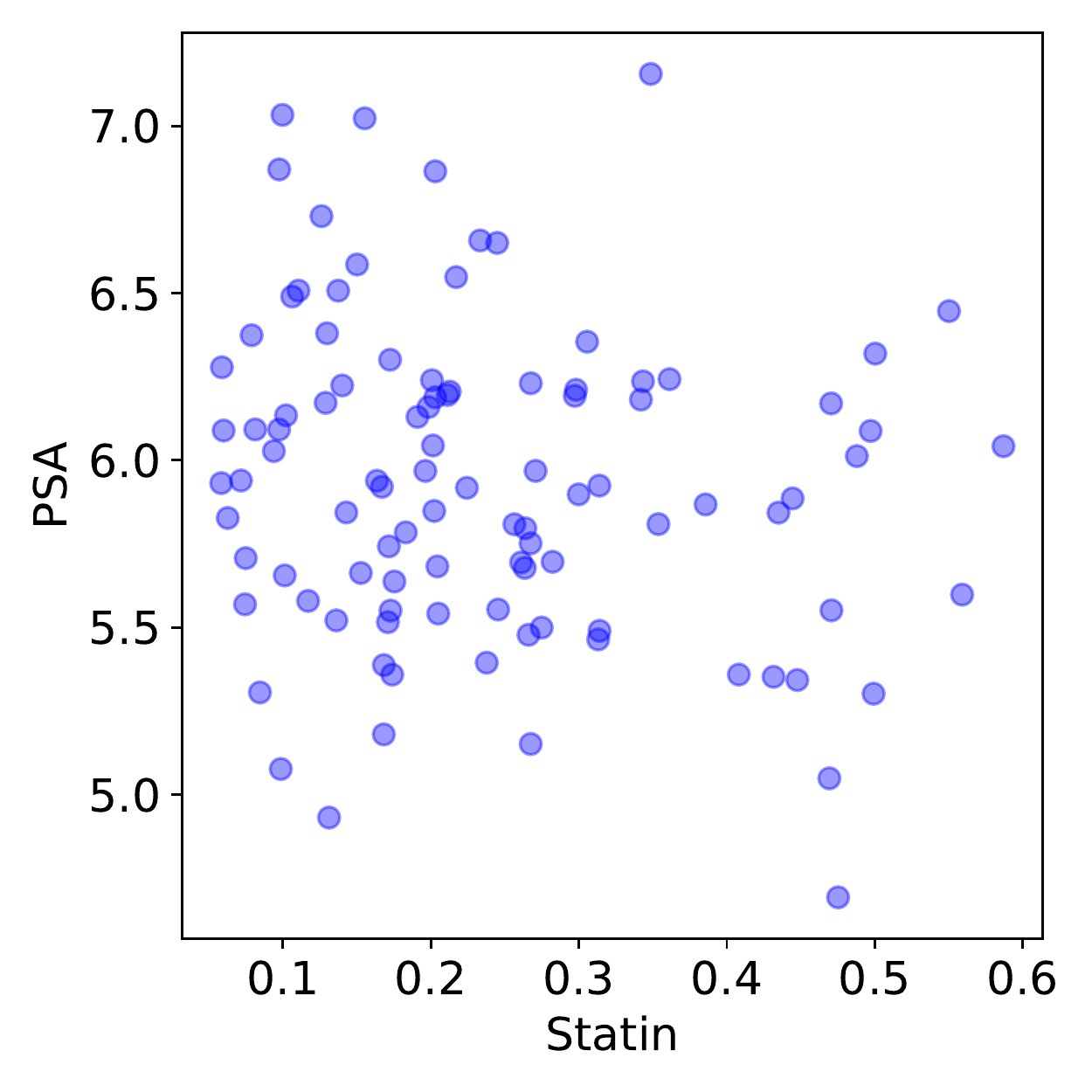}
\caption{\health. $\Nobs=100$ observational data samples from the \scm in \secref{sec:appreal1}.}
\label{fig:scatterhealth}
\end{figure}

For $\Nobs = 100$, these interventional ranges are only partially covered by the observational data (\figref{fig:scatter_health}) thus significantly complicating the estimation of the effects with $\datao$, especially when the stochasticity determined by $p(\U)$ is high (see scatter plots for \acro{ci}, Statin, Aspirin and \psa in \figref{fig:scatterhealth}).

We have $\powerI =\{\{\text{Aspirin}\}, \{\text{Statin}\}, \{\text{\acro{ci}}\},  \{\text{Aspirin}, \text{Statin}\}, \\ \{\text{Aspirin}, \text{\acro{ci}}\}, \{\text{Statin}, \text{\acro{ci}}\}, \{\text{Aspirin}, \text{Statin}, \text{\acro{ci}}\}\}$ which is equal to 
$\allmis{\C \cup Y}$.
\bmi is reducible for $\X=\{\text{Aspirin}\}$, $\X=\{\text{Statin}\}$ and $\X=\{\text{Aspirin}, \text{Statin}\}$ as $\X \cap \an(\bmi,\graph_{\overline{\X}})=\emptyset$. Given an observational dataset of size $\Nobs=100, 10$, $\hat{\mu}^{\bmi} \approx 26$ making \bmi not null-feasible. We therefore obtain $\allmisnull{\C \cup Y} = \{\{\text{\acro{ci}}\}, \{\text{Aspirin}, \text{\acro{ci}}\}, \{\text{Statin}, \text{\acro{ci}}\}, \{\text{Aspirin}, \text{Statin}, \text{\acro{ci}}\}\}$. 

\begin{figure}
\centering
\includegraphics[width=0.47\textwidth]{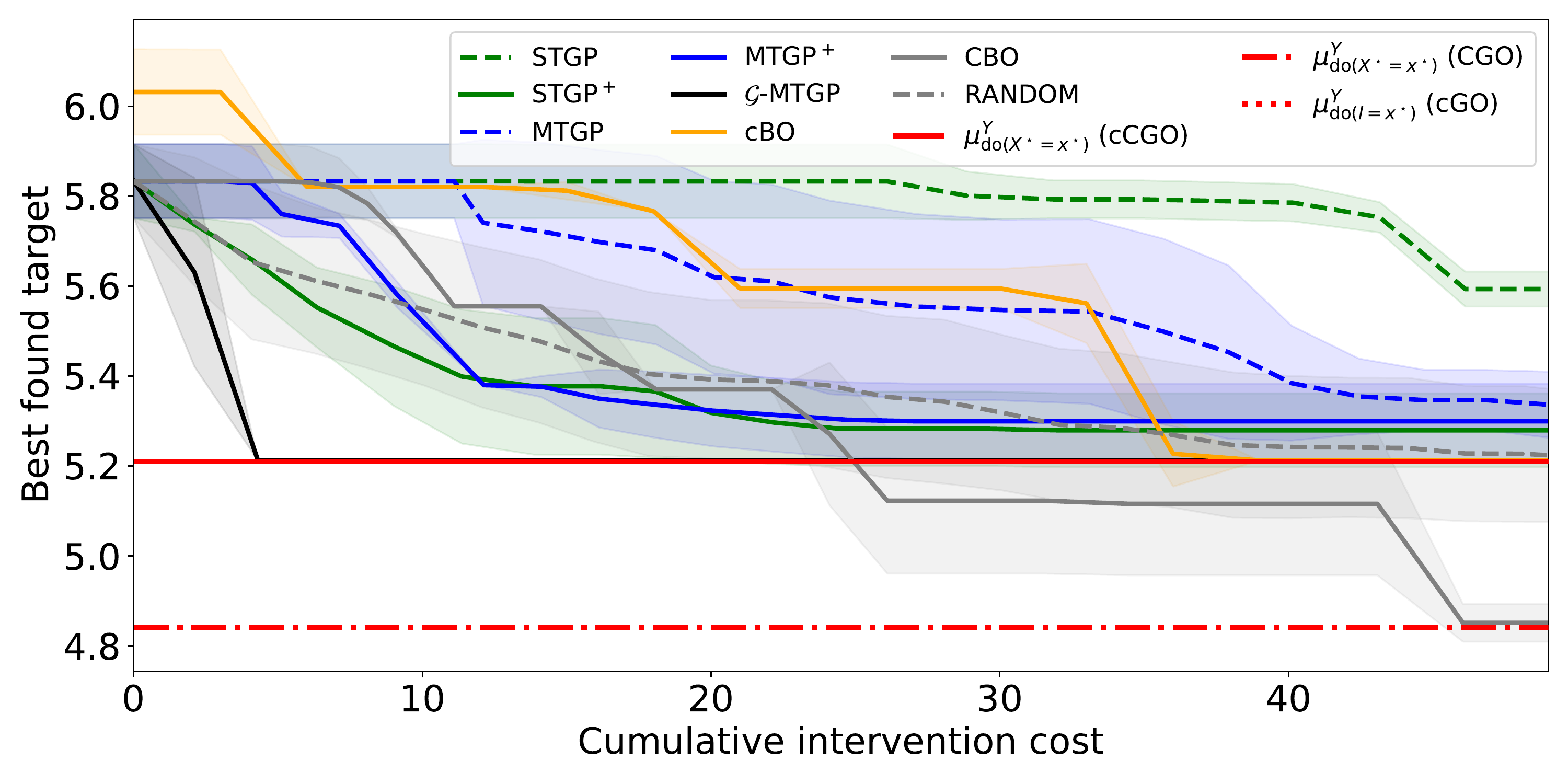}
\includegraphics[width=0.48\textwidth]{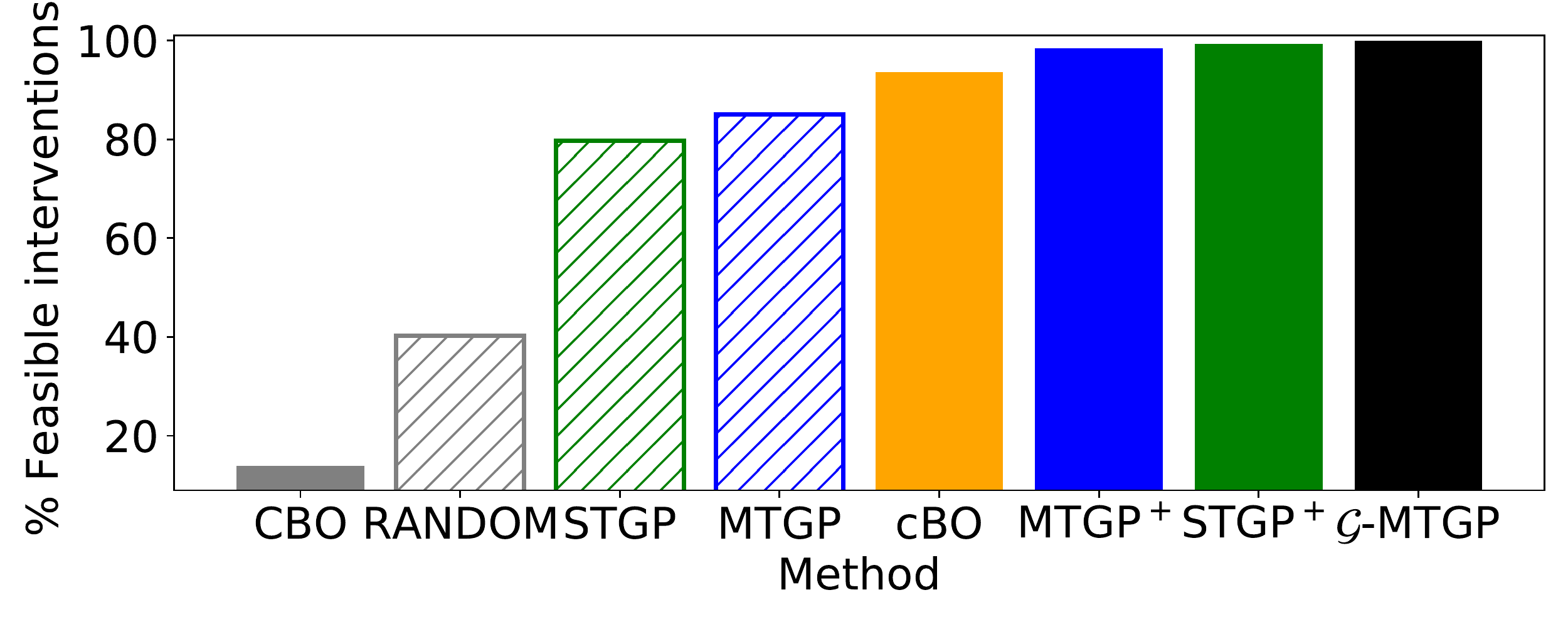}
\caption{$\health$ with $\Nobs=100$. \textit{Top}: Convergence to the \ccgo (solid red line), \cgo (dash-dotted red line) and \Cgo (dotted red line) optima.
\textit{Bottom}: Average percentage of feasible interventions collected over trials.
}
\label{fig:all_health}
\end{figure}

For completeness we report the full comparison with \cbo and \acro{random} in \figref{fig:all_health}. As in the previous experiments, the \ccgo and \cgo optima differ thus \cbo and \acro{random} converge, at a much lower speed, to different values while collecting a large number of infeasible interventions.

\subsection{\protein}
\label{sec:appreal2}
For the causal graph\footnote{We exclude the \acronospace{p}lc$\gamma$ phospho-protein, the \acro{pip2} and the \acro{pip3} phospho-lipids, as these form a separate graph with no connections to \acronospace{e}rk.} of \figref{fig:causalgraphs1}(a) with $\I = \{\text{\acro{pkc}}, \text{\acro{pka}}, \text{\acronospace{m}ek}, \text{\akt} \}$ and $\C = \{\text{\acro{pkc}}, \text{\acro{pka}}\}$, 
\citet{sachs2005causal} give a dataset but no \scm. Thus we first fit a \scm by placing a \gptext prior with zero mean and \acro{rbf} kernel on $f_V$, $\forall V \in \V$, and using the observational data. We assume a Gaussian distribution with zero mean and unit variance for $\acro{pkc}$. We use the learned \scm to generate observational and interventional data. 

\begin{figure}[t]
\centering
\includegraphics[width=0.46\textwidth]{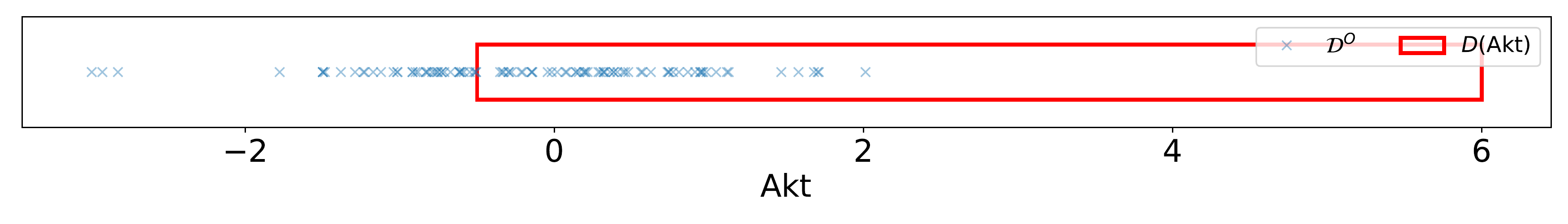}
\includegraphics[width=0.46\textwidth]{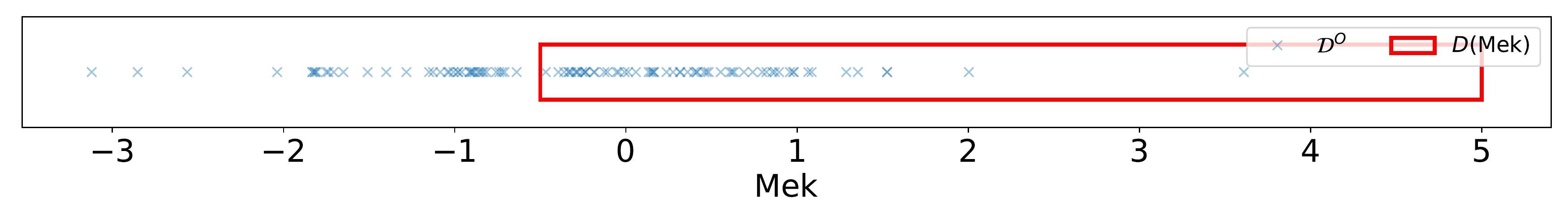}
\includegraphics[width=0.46\textwidth]{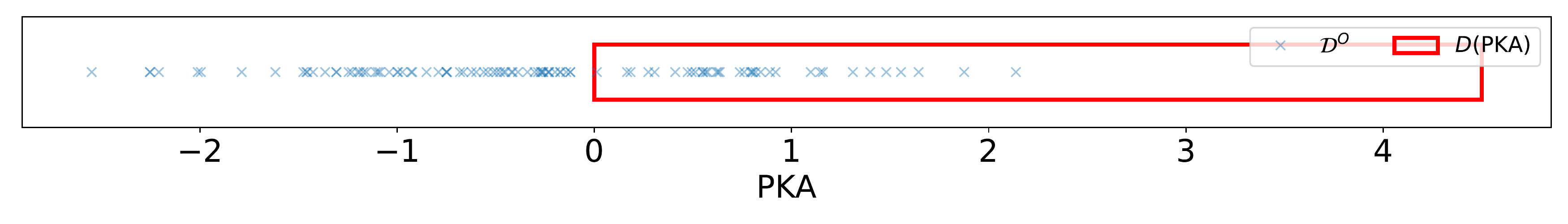}
\includegraphics[width=0.46\textwidth]{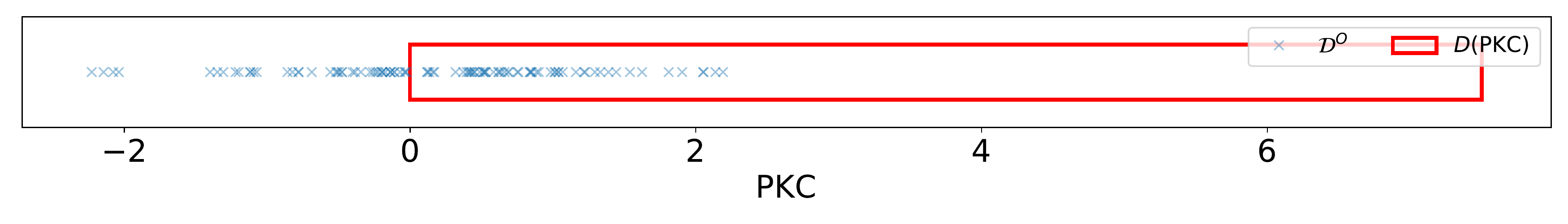}
\caption{\protein with $\Nobs=100$. Scatter plots for the observational data together with interventional ranges $D(\acro{\akt})$ (top row), $D(\text{\mek})$ (second row), $D(\text{\pka})$ (third row) and $D(\text{\pkc})$ (bottom row).}
\label{fig:scatter_protein}
\end{figure}

As in the previous experiments, the interventional ranges are only partially covered by $\Nobs=100$ observational samples (\figref{fig:scatter_protein}) which complicates the estimation of the effects with $\datao$. The power set of $\I$ is given by $\powerI = \{\{\acro{pkc}\}, \{\acro{pka}\}, \{\acro{m}\text{ek}\}, \{\text{\akt}\}, \{\acro{pkc}, \acro{pka}\}, \{\acro{pkc}, \acro{m}\text{ek}\}, \\ \{\acro{pkc}, \text{\akt}\}, \{\acro{pka}, \acro{m}\text{ek}\}, \{\acro{pka}, \text{\akt}\}, \{\acro{m}\text{ek}, \text{\akt}\}, \{\acro{pkc}, \\ \pka, \acro{m}\text{ek}\}, \{\acro{pkc}, \pka, \text{\akt}\}, \{\pka, \acro{m}\text{ek}, \text{\akt}\}\}$. Note that, for $\X \in \{\{\text{\akt}\}, \{\acro{pkc}, \text{\akt}\}, \{\acro{pka}, \text{\akt}\}, \{\acro{m}\text{ek}, \\ \text{\akt}\}, \{\acro{pkc}, \pka, \text{\akt}\}, \{\pka, \acro{m}\text{ek}, \text{\akt}\}\}$, we have  $\X \not\subseteq \an(\{\text{\pkc}, \text{\pka},\text{\erk}\}, \graph_{\overline{\X}}) \cup (\{\text{\pkc}, \text{\pka}\} \cap \X)$ thus we can exclude these sets from $\powerI$ and obtain $\allmis{\C \cup Y} = \powerI \backslash \{\{\text{\akt}\}, \{\acro{pkc}, \text{\akt}\}, \{\acro{pka}, \text{\akt}\}, \{\acro{m}\text{ek}, \text{\akt}\}, \{\acro{pkc}, \\ \pka, \text{\akt}\}, \{\pka, \acro{m}\text{ek}, \text{\akt}\}\}$. Given $\datao$ with size $\Nobs=100, 10$, all constrained variables are null-feasible. 

The set $\X=\{\acro{pkc}, \acro{pka}\}$ in $\allmis{\C \cup Y}$ has only one constrained variable $\text{\pkc}$ that is reducible (as $\X \cap \an(\text{\pkc}, \graph_{\overline{\X}}) =\emptyset$) and null-feasible for $\Nobs=100, 10$. We can thus exclude $\X^\prime = \{\acro{pkc}, \acro{pka}, \acro{m}\text{ek}\} \supset \X$ from $\allmis{\C \cup Y}$ as it is of the form $\X^\prime = \X \cup \acro{pkc}$ with $\acro{pkc} \not \in \an(\emptyset \cup Y, \graph_{\overline{\X^\prime}}) = \emptyset$. We therefore obtain $\allmisnull{\C \cup Y} = \allmis{\C \cup Y} \backslash \{\acro{pkc}, \acro{pka}, \acro{m}\text{ek}\}\}$.

\begin{figure}[t]
\centering
\includegraphics[width=0.45\textwidth]{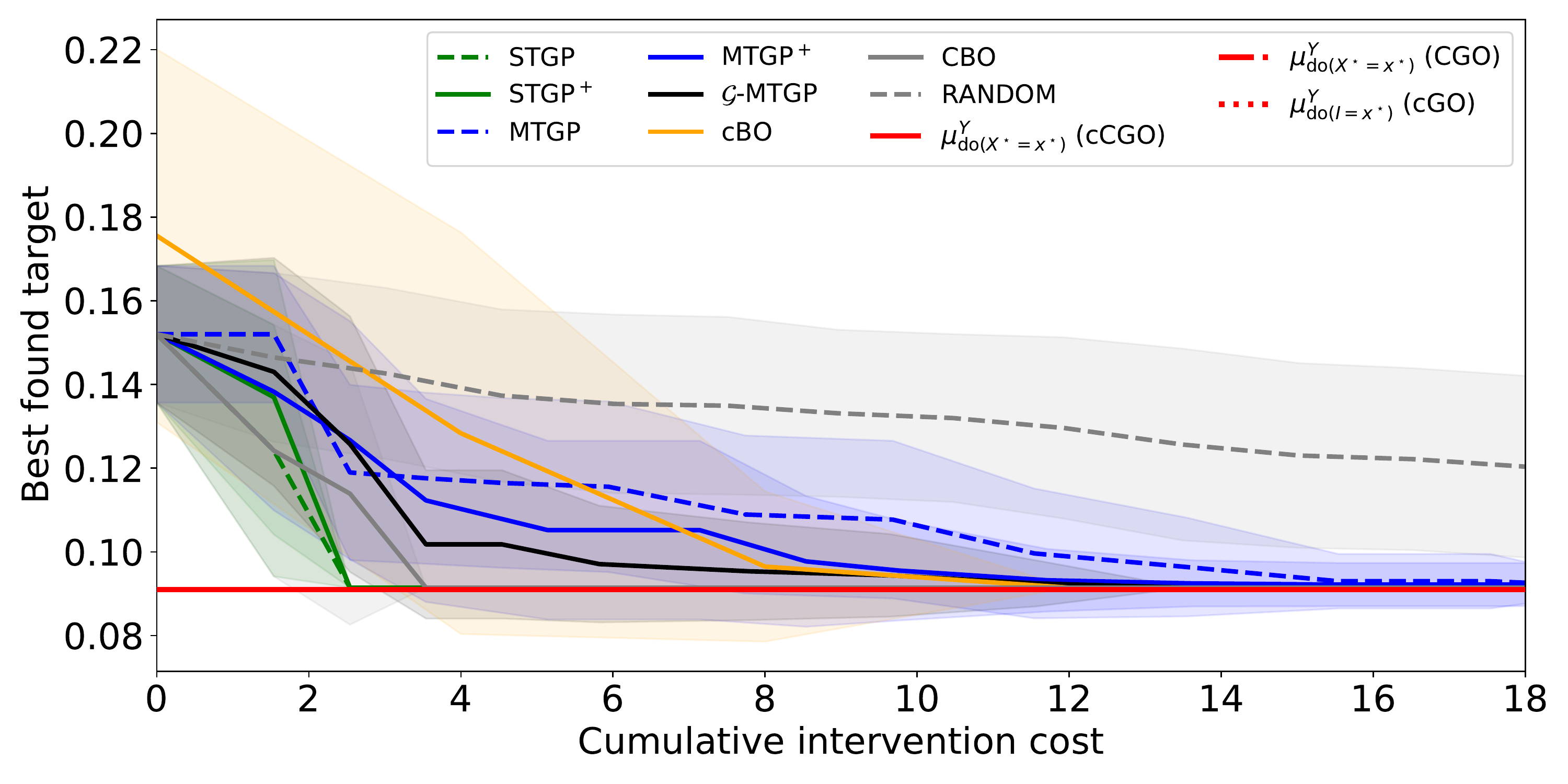}
\includegraphics[width=0.45\textwidth]{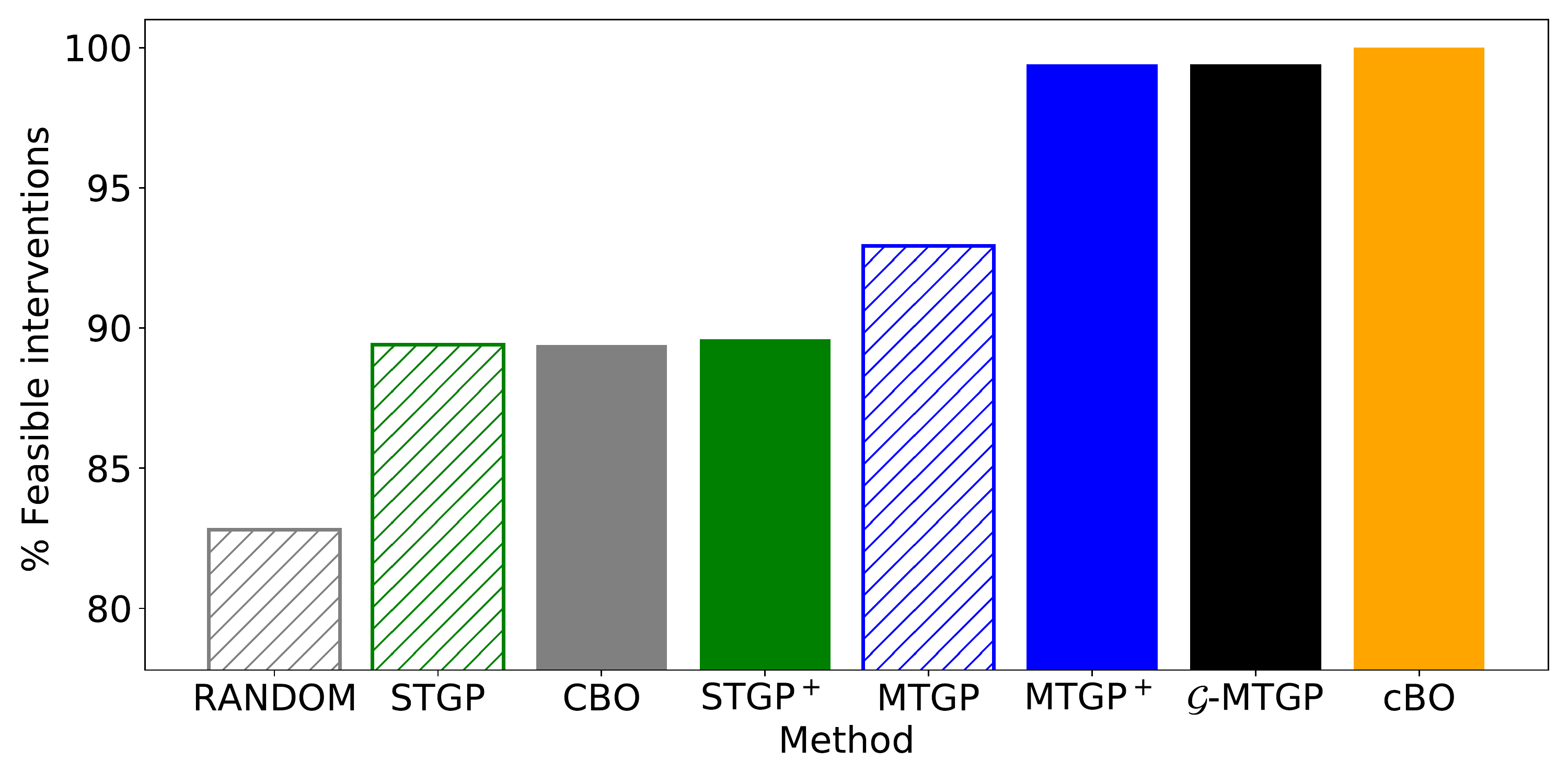}
\caption{$\protein$ with $\Nobs=100$. \textit{Top}: Convergence to the \ccgo (solid red line), \cgo (dash-dotted red line) and \Cgo (dotted red line) optima. \textit{Bottom}: Percentage of feasible interventions collected over trials.}
\label{fig:all_protein}
\end{figure}

In addition, as the \cgo optimum equals the \ccgo optimum, \cbo and \acro{random} converge to the same value achieved by \ccbo(\figref{fig:all_protein}, top row) but collect a significantly higher number of infeasible interventions (\figref{fig:all_protein}, bottom row).
\end{document}